%% file: main.tex
\documentclass[lettersize,journal]{IEEEtran}
\IEEEoverridecommandlockouts
\input{preamble/flags}
\input{preamble/packages_IEEE}
\input{preamble/custom_commands}

\input{preamble/ieee_settings}

\doubleblindfalse

\begin{document}
\title{When AI Bends Metal: AI-Assisted Optimization of Design Parameters in Sheet Metal Forming}

\input{preamble/authors_IEEE}
\markboth{}{When AI Bends Metal: AI-Assisted Optimization of Design Parameters in Sheet Metal Forming}

\maketitle
\input{content/00_abstract_IEEE}


\input{content/01_introduction}

\input{content/02_sheet_metal_forming}

\input{content/03_dl}
\input{content/04_approach}
\input{content/05_evaluation}
\input{content/06_discussion}
\input{content/07_related_work}
\input{content/08_conclusion}
\input{content/10_acknowledgement}

\input{preamble/gloss_print}
\bibliographystyle{IEEEtran}
\bibliography{ref}

\end{document}

%% file: preamble/flags.tex
\newif\ifremovetodos
\removetodosfalse

\newif\ifdoubleblind
\doubleblindtrue 
\doubleblindfalse

\newif\ifextended
\extendedfalse

\newif\ifgloss 
\glossfalse

\ifdoubleblind
\else
\removetodostrue
\fi 

%% file: preamble/packages_IEEE.tex
\usepackage{cite}
\usepackage{amsmath,amssymb,amsfonts}
\usepackage[dvipsnames, table]{xcolor}
\PassOptionsToPackage{table,dvipsnames}{xcolor}
\usepackage[hypertexnames=false,hidelinks]{hyperref}
\usepackage{orcidlink}
\usepackage{algorithmic}
\usepackage{graphicx}
\usepackage{textcomp}
\usepackage[normalem]{ulem}
\usepackage{multirow}
\usepackage{booktabs}
\usepackage{fancyvrb}
\usepackage{rotating}
\usepackage{tabularray}
\usepackage{tablefootnote}
\usepackage{printlen}
\usepackage{listings}
\usepackage[]{threeparttable}
\usepackage{listings}
\usepackage{xspace}
\usepackage{paralist}
\usepackage{xargs}
\usepackage{siunitx}
\usepackage{svg} 
\svgsetup{inkscapepath=svgsubdir}

\usepackage{subcaption}
\def\BibTeX{{\rm B\kern-.05em{\sc i\kern-.025em b}\kern-.08em
    T\kern-.1667em\lower.7ex\hbox{E}\kern-.125emX}}
\usepackage{tikz}
\usetikzlibrary{calc}

\usepackage{tabularx} 
\usepackage{pgfcalendar} 
\usepackage{cleveref} 
\usepackage[most,minted]{tcolorbox} 
\usepackage{fontawesome} 
\ifgloss
\usepackage[automake,toc,acronym,shortcuts,nonumberlist]{glossaries-extra} 
\fi

\ifremovetodos
\usepackage[disable]{todonotes}
\else
\usepackage{todonotes}
\fi

%% file: preamble/custom_commands.tex
\definecolor{c1}{RGB}{0,73,114}
\definecolor{c2}{RGB}{238,241,251}
\definecolor{MidnightBlue}{rgb}{0.0, 0.2, 1}
\definecolor{charcoal}{rgb}{0.21, 0.27, 0.31}

\ifremovetodos
\newcommand{\instr}[1]{}
\newcommand{\warn}[1]{}
\else
\newcommand{\instr}[1]{\xspace{\begin{infobox}{#1}\end{infobox}\xspace}}
\newcommand{\warn}[1]{\xspace{\begin{warningbox}{#1}\end{warningbox}\xspace}}
\fi

\ifdoubleblind
	\newcommand{\softname}[1]{\textit{Tool}~}

\else

        \newcommand{\softname}[1]{\textit{ftio}~}
	
\fi

\newcount\myjuliandate
\newcount\myjuliantoday

\newcommand{\difftoday}[3]{
    \pgfcalendardatetojulian{\year-\month-\day}{\myjuliantoday}%
    \pgfcalendardatetojulian{#1-#2-#3}{\myjuliandate}%
    \advance\myjuliandate by-\myjuliantoday\relax
    \ifnum\myjuliandate>5
        \textcolor[rgb]{0,0.5,0}{\textbf{\makebox[1cm]{\the\myjuliandate}}} 
    \else
        \ifnum\myjuliandate>0
            \textcolor{orange}{\textbf{\makebox[1cm]{\the\myjuliandate}}}
        \else
            \textcolor{red}{\textbf{\makebox[1cm]{\the\myjuliandate}}}
        \fi
    \fi
}

\crefname{equation}{Eq.}{Eqs.}
\crefrangelabelformat{equation}{(#3#1#4--#5#2#6)}
\crefrangeformat{equation}{#3Eqs. (#1)#4 to #5(#2)#6}
\crefmultiformat{equation}{#2Eqs. (#1)#3}{ and #2(#1)#3}{, #2(#1)#3}{ and #2(#1)#3}
\crefrangemultiformat{equation}{#3Eqs. ((#1))#4 to #5((#2))#6}{ and #3(#1)#4 to #5(#2)#6}{, #3(#1)#4 to #5(#2)#6}{ and #3(#1)#4 to #5(#2)#6}
\crefname{figure}{Figure}{Figures}
\Crefname{figure}{Figure}{Figures}
\crefname{section}{Section}{Sections}
\Crefname{section}{Section}{Sections}
\crefname{table}{Table}{Tables}
\Crefname{table}{Table}{Tables}

\DeclareTCBListing{shellCommand}{ !O{||} !O{\footnotesize}}{
    breakable,
    listing engine=minted,
    minted style=vs,
    boxsep=1pt,left=2pt,right=1pt,top=4pt,bottom=4pt,
    minted language=console,
    minted options = {escapeinside=#1,mathescape=true, fontsize=#2, tabsize=2, breaklines,  numbersep=4mm,}, 
    listing only,
    boxrule=1pt
}

\newtcblisting{python}{
	breakable,
	enhanced,
	listing engine=minted,
	minted style=vs,
	minted language=python,
	minted options = {linenos,fontsize=\small, numbersep=4mm,},
	left=8mm,
	overlay={
			\begin{tcbclipinterior}
				\fill[gray!25] (frame.south west) rectangle ([xshift=7mm]frame.north west);
			\end{tcbclipinterior} },
	listing only,
	boxrule=1pt,
	listing options={ numbers=right, basicstyle=\small\ttfamily,}
}

\newtcblisting{onejsonl}{
	breakable, enhanced,
	listing engine=minted,
	minted style=tango,
	minted language=js,
	minted options = {fontsize=\small, numbersep=4mm},
	left=18pt,right=1pt,top=4pt,bottom=4pt,
	overlay={
			\begin{tcbclipinterior}
				\fill[gray!25] (frame.south west) rectangle ([xshift=7mm]frame.north west);
			\end{tcbclipinterior} },
	listing only,
	boxrule=1pt,
	listing options={ numbers=right, basicstyle=\small\ttfamily,}
}

\newtcblisting{jsonl}[1][\tiny]{%
	breakable, enhanced,
	listing engine=minted,
	minted style=tango,
	minted language=js,
	minted options = {escapeinside=||,mathescape=true,fontsize=#1, tabsize=2,breaklines,  numbersep=4mm,},
	boxsep=1pt,left=0pt,right=1pt,top=1pt,bottom=1pt,
	overlay={
			\begin{tcbclipinterior}
				\fill[gray!25] (frame.south west) rectangle ([xshift=7mm]frame.north west);
			\end{tcbclipinterior} },
	listing only,
	boxrule=1pt,
	listing options={ numbers=right, basicstyle=\small\ttfamily,}
}

\newtcolorbox{mywp}[2]{
	before skip=1.2cm,      
	before upper={\itshape},   
	fonttitle=\large\bfseries,       
	colback=white,             
	colframe=black,            
	title={\hspace*{-10pt}#1},       
	left=10pt,                  
	right=10pt,                 
	top=1pt,                   
	bottom=1pt,                
	separator sign none,        
	label={#2},
}

\newtcolorbox[auto counter,number within=section]{warningbox}[1][]{
	enhanced jigsaw,colback=white,colframe=orange!70,coltitle=orange!70,
	fonttitle=\bfseries\sffamily,
	detach title,
	leftrule=0.1\columnwidth,
	underlay unbroken and first={\node[below,text=white,anchor=center]
			at ([xshift=-0.05\columnwidth]interior.base west) {\Huge\faWarning};},
	breakable,pad at break=1mm,
	#1,
	code={\ifdefempty{\tcbtitletext}{}{\tcbset{before upper={\tcbtitle\par\medskip}}}},
}

\newtcolorbox[auto counter,number within=section]{infobox}[1][]{
	enhanced jigsaw,colback=white,colframe=charcoal,coltitle=charcoal,
	fonttitle=\bfseries\sffamily,
	detach title,
	leftrule=0.1\columnwidth,
	underlay unbroken and first={\node[below,text=white,anchor=center]
			at ([xshift=-0.05\columnwidth]interior.base west) {\Huge\faInfoCircle};},
	breakable,pad at break=1mm,
	#1,
	code={\ifdefempty{\tcbtitletext}{}{\tcbset{before upper={\tcbtitle\par\medskip}}}},
}

\usepackage{adjustbox}
\usepackage{tabularx}
\usepackage{booktabs}
\newcolumntype{L}[1]{>{\raggedright\arraybackslash}p{#1}} 
\newcolumntype{C}[1]{>{\centering\arraybackslash}p{#1}} 
\newcolumntype{R}[1]{>{\raggedleft\arraybackslash}p{#1}} 


%% file: preamble/ieee_settings.tex
\makeatletter
\newcommand{\linebreakand}{%
  \end{@IEEEauthorhalign}
  \hfill\mbox{}\par
  \mbox{}\hfill\begin{@IEEEauthorhalign}
}
\makeatother

%% file: preamble/authors_IEEE.tex
\ifdoubleblind
\else

\author{
\IEEEauthorblockN{
Ahmad Tarraf\IEEEauthorrefmark{1},
Koutaiba Kassem-Manthey\IEEEauthorrefmark{2},
Seyed Ali Mohammadi\IEEEauthorrefmark{1},
Philipp Martin\IEEEauthorrefmark{3},\\
Lukas Moj\IEEEauthorrefmark{4},
Semih Burak\IEEEauthorrefmark{3},
Enju Park\IEEEauthorrefmark{4},
Christian Terboven\IEEEauthorrefmark{3},
and Felix Wolf\IEEEauthorrefmark{1}
}

\IEEEauthorblockA{\IEEEauthorrefmark{1}\textit{Department of Computer Science, Technical University of Darmstadt}, Darmstadt, Germany}

\IEEEauthorblockA{\IEEEauthorrefmark{2}\textit{GNS Gesellschaft für numerische Simulation mbH}, Braunschweig, Germany}

\IEEEauthorblockA{\IEEEauthorrefmark{3}\textit{Chair for High Performance Computing, IT Center, RWTH Aachen University}, Aachen, Germany}

\IEEEauthorblockA{\IEEEauthorrefmark{4}\textit{GNS Systems GmbH}, Braunschweig, Germany}


\IEEEauthorblockA{Contact: ahmad.tarraf@tu-darmstadt.de}

}

\fi

%% file: content/00_abstract_IEEE.tex
\begin{abstract}

Numerical simulations have revolutionized the industrial design process by reducing prototyping costs, design iterations, and enabling product engineers to explore the design space more efficiently. However, the growing scale of simulations demands substantial expert knowledge, computational resources, and time. A key challenge is identifying input parameters that yield optimal results, as iterative simulations are costly and can have a large environmental impact.
This paper presents an AI-assisted workflow that reduces expert involvement in parameter optimization through the use of Bayesian optimization.  Furthermore, we present an active learning variant of the approach, assisting the expert if desired. A deep learning model provides an initial parameter estimate, from which the optimization cycle iteratively refines the design until a termination condition (e.g., energy budget or iteration limit) is met.
We demonstrate our approach, based on a sheet metal forming process, and show how it enables us to accelerate the exploration of the design space while reducing the need for expert involvement. 
\end{abstract}


%% file: content/01_introduction.tex
\section{Introduction}
\label{sec:introduction}

\begin{figure}[hthp]
	\centering
	\includegraphics[width=.4\textwidth, trim=0 15 365 15, clip]{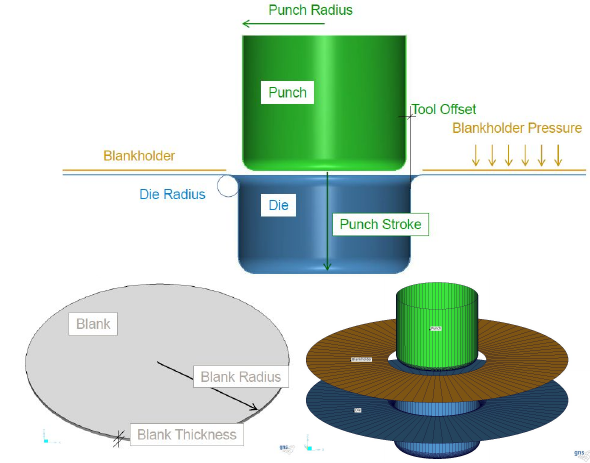}
	\caption{Forming of a cylindrical cup.}
	\label{fig:CCup_into}
\end{figure}

The design of engineering systems is an iterative process that involves
analyzing and evaluating successive designs until an acceptable solution is
reached~\cite{JSA04}. Design optimization problems abound in all engineering
disciplines, including wing design in aerospace engineering, process control in
chemical engineering, structural design in civil engineering, circuit design in
electrical engineering, and mechanism design in mechanical
engineering~\cite{martins2021engineering}. One such problem, in the manufacture
of car body parts or household appliance components, is finding the optimal
parameter configuration for sheet metal forming by deep
drawing~\cite{tschaetsch2006metalforming}. Interestingly, due to the growing
demand for miniaturization, the production of sheet-formed products has
experienced substantial growth in recent years~\cite{OMI22}. Industries like
automotive and packaging mass-produce hollow shapes using cost-effective
methods such as the deep drawing process~\cite{OMI22}. Deep drawing is a
standard technique for deforming plane sheets of metal in a controlled way. The
main idea is to draw the metal sheet into an appropriately shaped die by the
mechanical action of a punch, as illustrated in \cref{fig:CCup_into}.

During the forming process, defects such as fractures, excessive local
thinning, or wrinkling can occur due to limited formability or improper
parameter settings~\cite{OMI22}. Consequently, accurate prediction and control
of the parameters (e.g. surface roughness of the metal and friction during the
forming process, and spring-back) in forming processes is very
important~\cite{OMI22,WP20,KB21}.
Numerical simulation of metal forming processes, as part of virtual
manufacturing, plays a crucial role in reducing lead time, with the finite
element method being the most widely used technique for simulating sheet metal
forming processes~\cite{DB10}. Simulation codes for sheet metal forming are
well-established tools for predicting the workpiece shape and properties for
given process parameters.
%
The task of the domain expert can be seen as an inverse problem: given the
target shape, required properties, and quality standards, they must determine
the process parameters that achieve the desired outcome. %

Determining design parameters for the sheet metal forming process involves a
large number of variables~\cite{LW08}, resulting in an extremely
high-dimensional search space. Additional constraints, such as budget limits,
require the domain expert to consider multiple cost sources, including energy,
material, compute time, and expert involvement. Often, these constraints also
become secondary objectives in the optimization cycle, further expanding the
search space. Considering that sheet metal forming is a typical multi-objective
problem with conflicting relationships between multiple objective
functions~\cite{LW08,OS04,TJ05,Tk04}, brute-force, factorial, or classical
search and sampling methods become impractical or highly inefficient, making
the determination of optimal design parameters extremely complex and costly,
and often requiring designers to rely on approximate solutions.
Numerical simulations can replace costly and time-consuming physical test
series~\cite{SH04}, saving resources and significantly reducing the time
required for processes. However, a vast number of possible configurations can
lead to a combinatorial explosion in required simulations, making the search
for near-optimal design parameters computationally demanding.

Yet, these simulations are among the most computationally demanding
applications in high-performance computing (HPC), consuming significant energy,
though generally far less than the actual processes they model. Nevertheless,
power consumption is a major contributor to the total cost of ownership in HPC.
For the HPC center of the RWTH Aachen University, power consumption contributed
28\% of the total costs as the second highest contributor, and hardware-related
costs as the first with roughly half of the total, while all other cost factors
are relatively minor~\cite{bischof2012brainware}. The cost of HPC systems
affects the budget companies have for simulations, limiting the number of
simulations they can afford to run. To minimize costs, the experts should find
the optimal design with the smallest possible number of simulations, which
requires considerable knowledge and years of experience. This becomes even more
challenging due to the numerous combinations of process parameters and the
time-consuming nature of finite element analysis~\cite{LW08}.

\begin{figure}[t]
	\centering
	\includegraphics[width=\columnwidth]{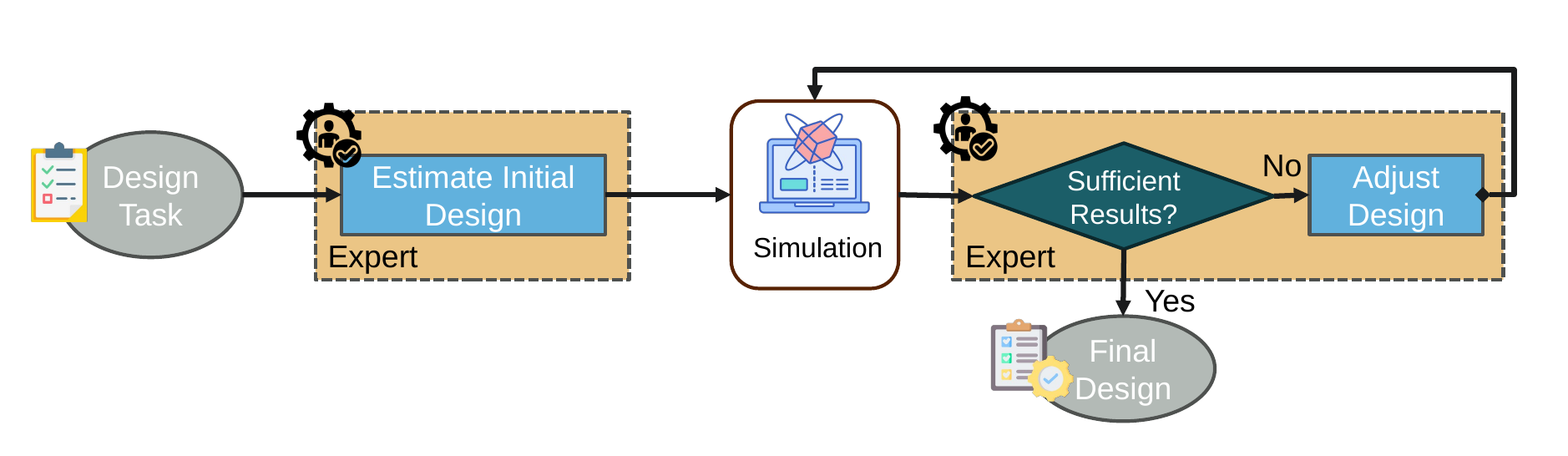}
	\caption{Traditional Workflow extensively relying on an expert to specify promising input design configurations for the simulation to obtain the acceptable target parameters.}
	\label{fig:old_workflow}
\end{figure}
\begin{figure}[t]
	\centering
	\includegraphics[width=\columnwidth,]{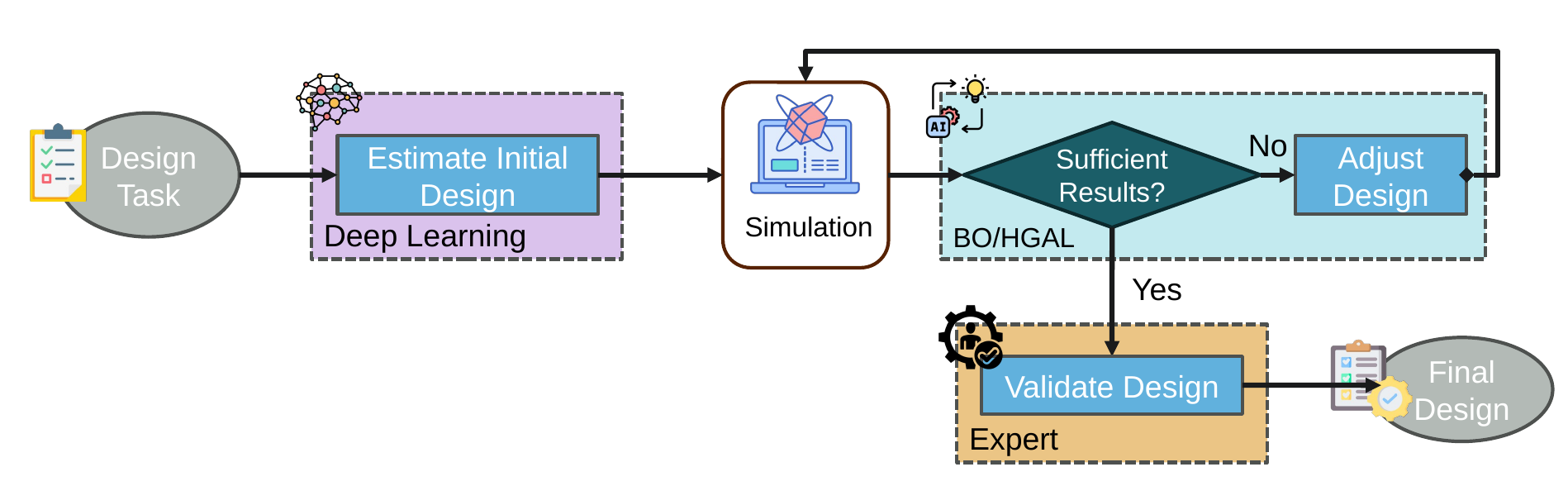}
	\caption{Improved workflow utilizing AI methods to reduce the expert involvement, opposed to Figure~\ref{fig:old_workflow}.}
	\label{fig:new_workflow}
\end{figure}

This paper proposes an AI-assisted workflow to reduce the expert involvement in both the initial design estimation and the optimization loop for finding a satisfying configuration.
\cref{fig:old_workflow} illustrates a traditional workflow that heavily relies on an expert to guide the design  processes, while \cref{fig:new_workflow} shows our approach.
As shown in \cref{fig:new_workflow}, rather than simply guessing, a deep
learning model, trained with a dataset consisting of past runs, predicts the
ideally optimal values of the design parameters for a new part. Due to effects
such as new geometries, materials, or constraints, the suggested parameters by
the deep learning component may be outside the acceptable ranges. The workflow
offers two approaches to refine the results: (1) human-guided active learning
 or (2) Bayesian optimization to identify input parameter
configurations for the simulation that yield the desired design parameters.
Although a full replacement of the domain expert is striven for, to prevent
actual manufacturing costs, the expert might want to verify the output design
configuration as the last step, as shown in \cref{fig:new_workflow}.

Using Bayesian optimization, the input parameters are iteratively adjusted to find optimal design parameter configurations until the specified end conditions are met. We employ  a Gaussian Process Latent Variable Model (GPLVM), which is iteratively improved with new samples from the simulations, while simultaneously guiding the input parameter configuration based on an acquisition function.
We demonstrate due workflow based on a sheet metal forming simulation process, highlighting how the approach enables the automation, acceleration, and improvement of design parameter explorations, while saving resources such as personnel and time. In short, our contributions are:
\begin{itemize}
	\item Designing an AI-assisted workflow to reduce expert involvement in the design
	      process while simultaneously decreasing personnel costs, runtime, and the
	      number of simulations, thereby reducing the environmental impact.
	\item Presenting two methods to to identify promising design parameter
	      configurations: human-guided active learning and Bayesian optimization based on
	      GPLVM models. Our approach further reduces computational cost and energy
	      consumption by early termination of non-promising configurations, and
	      accelerates key algorithm components through heterogeneous computing.
	\item Providing various improvements to the AI workflow to enhance performance and
	      reduce time-to-solution. Using a mixture-of-experts approach, we show that
	      simulation input parameters can be optimally selected even for new parts.
	      Additionally, parallel sampling reduces the time required to reach the best
	      solution.
	\item Demonstrating the workflow based on a deep-drawing simulation software
	      OpenForm, and various sheet metal parts, highlighting how our approach allows
	      us to reduce expert involvement and improve sample selection.
\end{itemize}

In what follows, we start by explain the typical sheet metal forming in
\cref{sec:sheet_metal_forming}. Afterwards, we describe how the deep learning
component suggests an initial configuration for the simulation
in~\cref{sec:dl}. To improve the results, active learning is employed either
fully automated, using Bayesian optimization, or with a human in the loop, as
described in~\cref{sec:BO}. We evaluate our solution in~\cref{sec:evaluation}
and provide a discussion of the results. \cref{sec:related} provides the
related work, and \cref{sec:conclusion} finally provides a conclusion and our
future work.




%% file: content/02_sheet_metal_forming.tex
\section{Sheet Metal Forming}
\label{sec:sheet_metal_forming}

Sheet metal forming is a process for producing metal parts of a desired geometry from a primarily planar metal sheet, as it is used throughout the automotive, consumer electronics, and houseware industries. 
Many questions must be addressed before deciding to manufacture a designed part. This includes selecting the appropriate forming process, such as deep drawing, hydroforming, or injection molding. Next, the process parameters and resources must be determined, including the required number of forming steps, die forces, and forming velocities. Furthermore, it is essential to identify the optimal initial workpiece configuration, considering factors such as sheet thickness, contour, and material properties
These parameters significantly influence the part's feasibility, the overall manufacturing costs, and the carbon footprint of the final product. Therefore, digital prototyping and numerical forming simulation can help companies maintain sustainability and reduce costs.

During numerical forming simulations, the forming process is replicated virtually in several short time or displacement steps until the whole process (i.e., the total motion of the tools) has been completed. In every step, the current deformed state of the workpiece is updated by considering the acting forces, the applied boundary conditions, and the material properties of the part. During and at the end of the simulation, various properties are monitored and displayed, including the feasibility of the part, i.e., the occurrence of cracks, wrinkles, or excessive thinning, the applied die forces, and the shape stability of the final part.

\subsection{Deep Drawing}
\label{subs:deepdrawing}
\begin{figure}[t]
\centering
\begin{subfigure}[b]{0.48\columnwidth}
\includegraphics[width=\columnwidth]{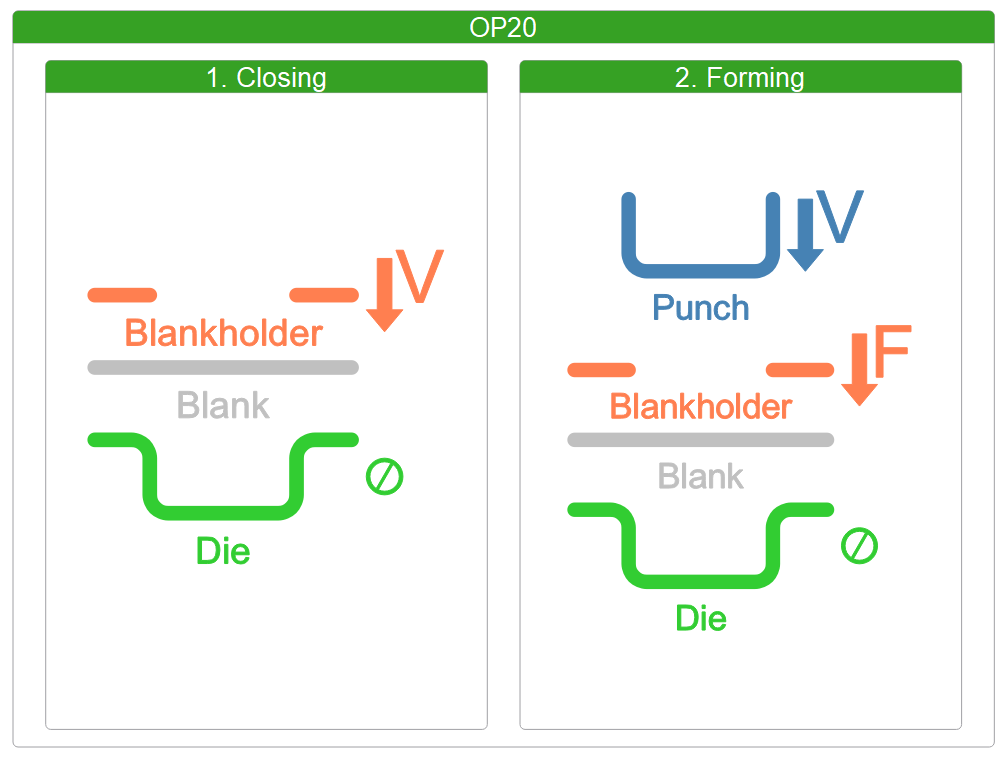}
\end{subfigure}
\hfill
\begin{subfigure}[b]{0.48\columnwidth}
\includegraphics[width=\columnwidth]{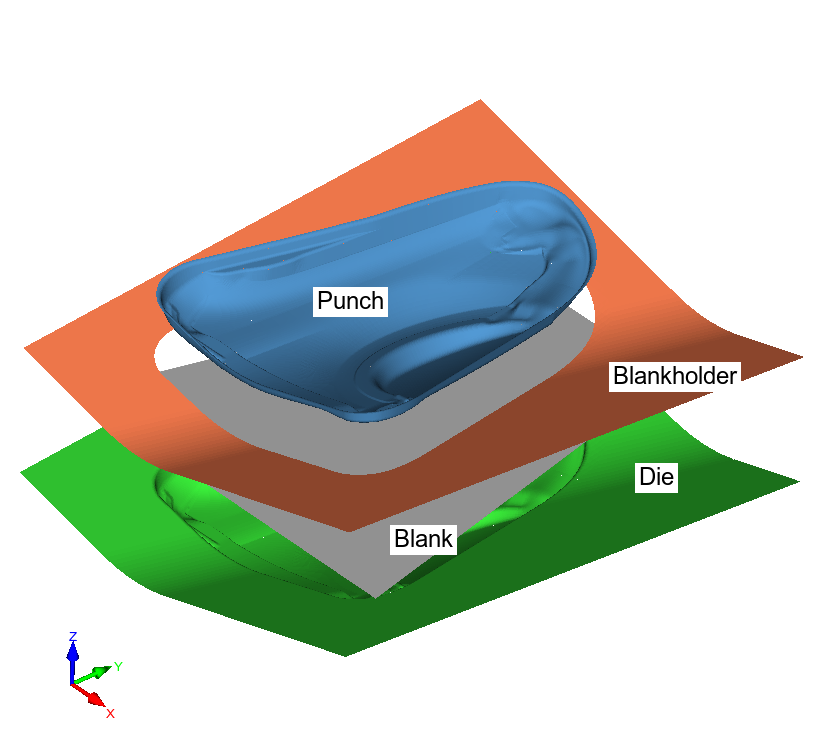}
\end{subfigure}
\caption{Deep-drawing process (left) and tools (right).}
\label{fig:deepdraw}
\end{figure}

The deep drawing process, shown in Figure~\ref{fig:deepdraw}, is a forming process in which a mostly plane metal sheet, called the blank, is drawn into a form (known as the die) by the mechanical action of a counterpart of the die, the so-called punch. Deep drawing thus describes a shape transformation process widely used to produce thin metal parts for the automotive, consumer electronics, and household appliance industries.  

During the first step of the deep drawing process, the closing step, an arbitrarily shaped (in many cases rectangular or circular) blank is placed onto the die, and a so-called blank holder or binder tool presses the blank against the die and thus, among other things, prevents the blank from sliding away during the forming process. 
In the next process step, the forming step, the blank material is drawn into the die by the moving punch until the latter closes, i.e.\ until the remaining distance between both tools (die and punch) is not greater than the blank thickness. 
Both the closing and the forming steps can be repeated in several drawing steps, allowing for the production of complicated parts. At the end of a drawing step or the entire deep drawing process, the part's boundary may be trimmed (trimming step), and some further follow-up operations, such as hole punching, plunging, and folding, may be necessary. Depending on the complexity and size of the part to be produced, a punching frequency of six to ten pieces per minute can usually be achieved in a deep drawing process.

\subsubsection{Control Parameters}
\begin{figure}[tbp]
    \centering
    \includegraphics[width=0.98\columnwidth]{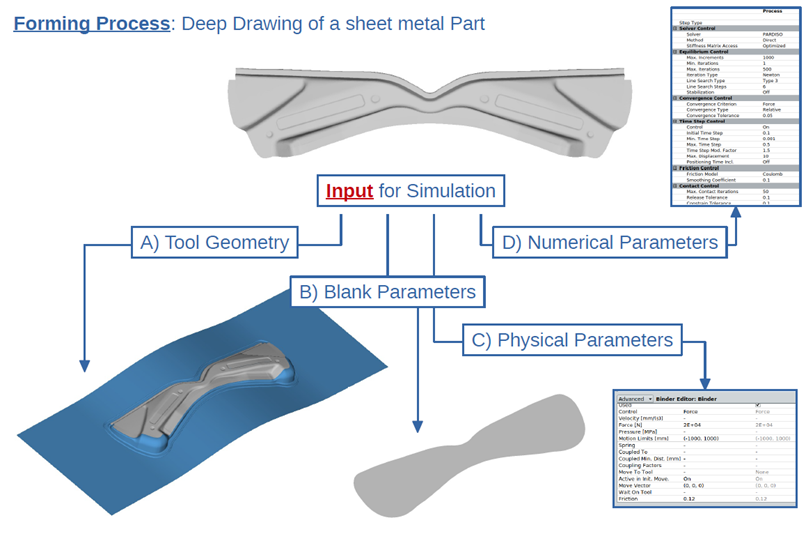}
    \caption{Input parameters for the numerical simulation. }
    \label{fig:input}
\end{figure}
\begin{figure}[htp]
\centering
\includegraphics[width=0.75\columnwidth]{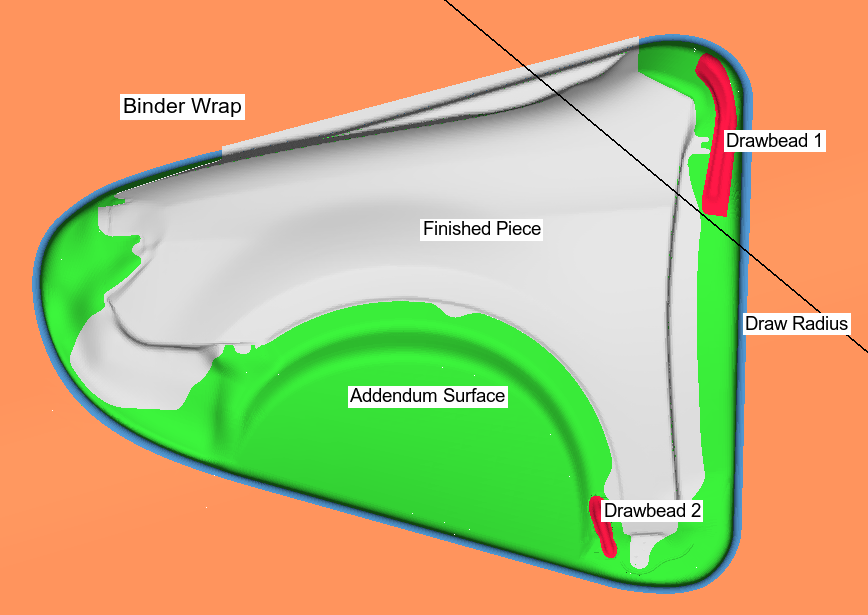}
\caption{Design of addendum, draw radii, and drawbeads for a front fender (with details in a cross-section).}
\label{fig:addendum}
\end{figure}

The deep drawing process is a shape transformation process based on material retention, i.e., the increase in the blank surface from the 2D flat sheet toward a formed 3D part is done by stretching the material at the expense of its thickness. Therefore, the parameters that influence the result of a deep-drawn part can be classified into two categories: parameters related to the forming tools and those related to the blank to be formed.
During the forming process, the flat blank is forced into the targeted geometry through continuous contact with the tool surfaces (the so-called operative surface or addendum); at the same time, the increase in blank surface area is achieved by controlling the material movement with the forming tools. Considering these, the control parameters related to the forming tools and influencing the result of the forming process are shown in \cref{fig:input}. 
The control parameters of the forming tools include the design of the addendum surface, the drawing radii over which the blank material flows, the pressure load applied by the blank-holder tool, and the configuration of the draw beads as visualized in \cref{fig:addendum}. 
Typically, the addendum surface and the drawing radii are specified first, followed by the pressure load of the blank holder tool. 
It follows the roughness (defined by the friction coefficient) of the tool surfaces (all tools).
In some cases, neither the binder load nor the tool friction is sufficient to restrain the blank and thus to achieve the necessary material extension. In such cases, geometric elements called draw beads or draw bars are incorporated into the binder and die surfaces to provide additional restraining effects.

\begin{figure}[bp]
\centering
\includegraphics[width=.85\columnwidth]{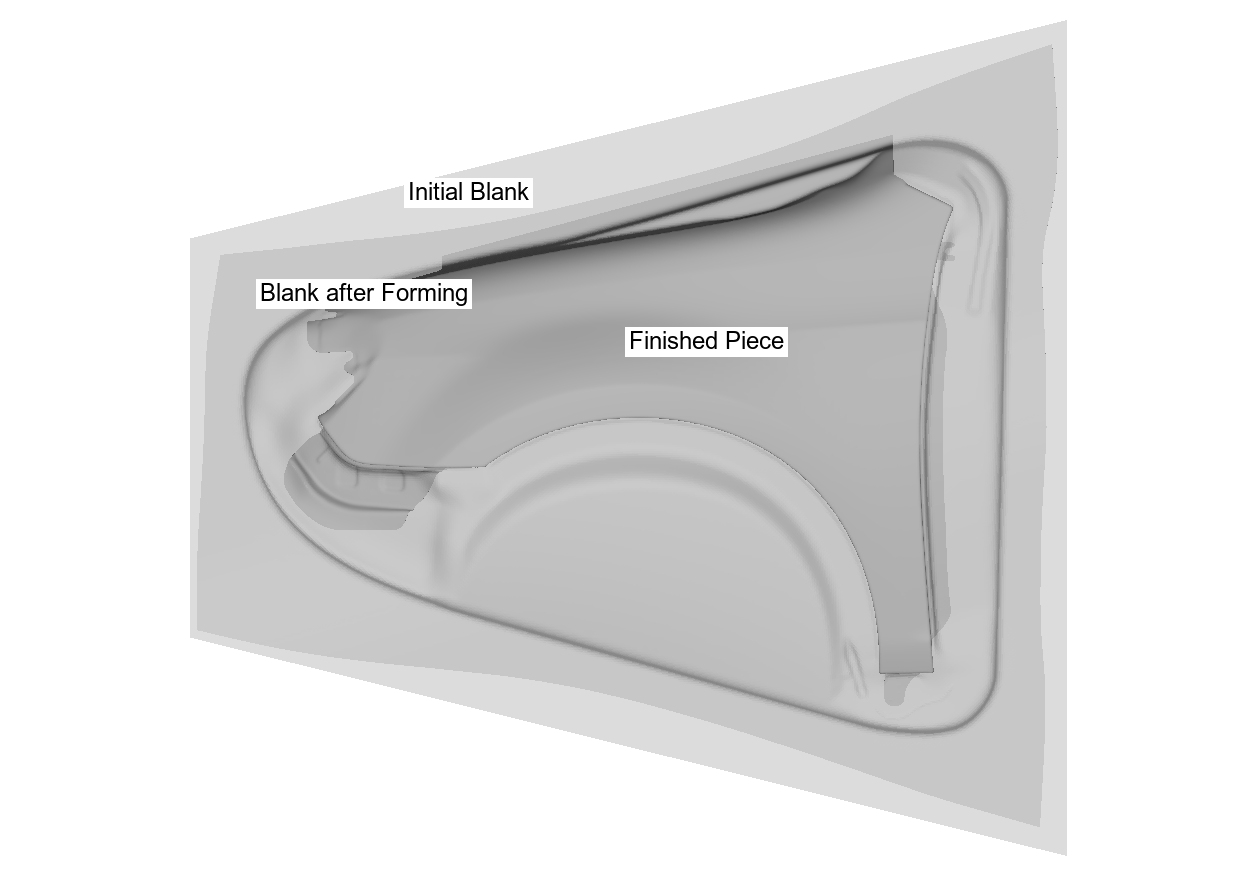}
\caption{Initial blank geometry and deformed workpiece for a front fender.}
\label{fig:initialblank}
\end{figure}

As mentioned above, the result of a deep drawing process depends on both the material flow during the process as well as the material extension. The control parameters related to the formed blank are: the blank thickness, material, and its initial outline (shape of the boundary) (cf.  \cref{fig:initialblank}).
The blank material is specified by the material's mechanical properties, such as yield stress, elongation at break, and R-values (ratio of in-plane to out-of-plane strains).

\begin{figure}
    \centering
    \includegraphics[width=0.98\columnwidth]{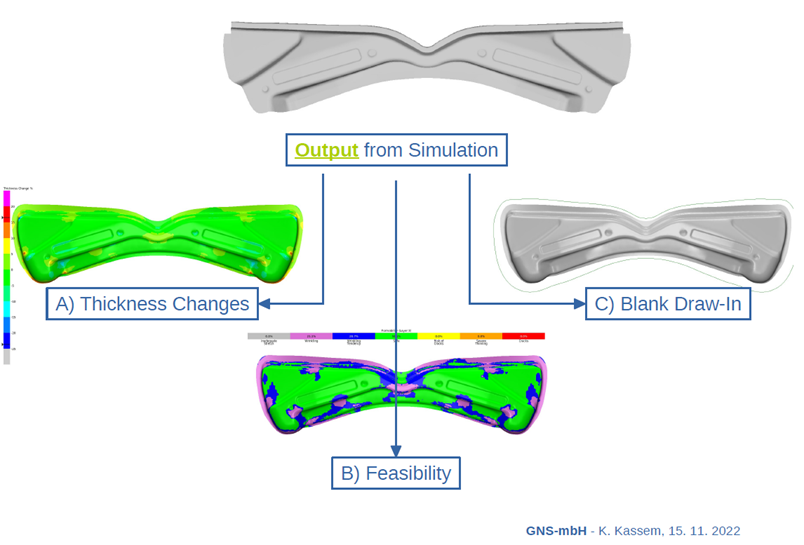}
    \caption{
    Target parameters of the simulation. 
    }
    \label{fig:target_parameters}
\end{figure}

In general, several targets and constraints must be met at the end of a deep drawing process. 
First, the targeted geometry, even after removing the part from the forming tools, must be met, i.e., minimizing large spring-back effects after unloading the part.
Second, the feasibility of the part needs to be ensured ($L1$ till $L7$), i.e.\ the absence of defects like cracks, wrinkles, or severe thinning in the final part.
Third, the presence of a minimum residual thickness must be ensured throughout the entire part. In some cases, it is also necessary not to exceed a maximum thickening.
All these requirements influence the stiffness, fatigue, and crash behavior of the part, thereby determining whether the manufactured part can be used successfully for its intended purpose.
Other considerations, such as reducing press loads, energy consumption, or material costs, are also often of great interest and can be part of a process optimization procedure.
Figure~\ref{fig:target_parameters} visualizes the degree of fulfillment of these requirements in the simulation software OpenForm, which will be introduced next.

\begin{table}[bhp]
\centering
\caption{Summary of the input and target parameters. 
Values in square brackets indicate units or enumerated options; unspecified values are numerical.}
\label{tab:parameters}
\scriptsize
\begin{tabularx}{\columnwidth}{X X X}
\toprule
\textbf{Fixed Design Parameters} & \textbf{Variable Design Parameters} & \textbf{Target Parameters} \\
\midrule
\textbf{Material Properties} \newline
\textbullet~Yield strength ($Rp$) \newline
\textbullet~Tensile strength  \newline
\textbullet~Elongation at break [\%] \newline
\textbullet~Stretch ratio during rolling  \newline
\textbf{Workpiece Geometry} \newline
\textbullet~Length width ratio  \newline
\textbullet~Width depth ratio  \newline
\textbullet~General shape [square, rectangle, bar, circle, ellipse, triangle] \newline
\textbullet~Part complexity [\%] \newline
\textbullet~Stroke length  \newline
\textbf{Fixed Design Parameter} \newline
\textbullet~Number forming steps 
&
\textbf{Process Parameter} \newline
\textbullet~Blankholder pressure ($p$)\newline
\textbullet~Draw beads restraining force average ($db$)  \newline
\textbullet~Number of draw beads with respect to workpiece edge [\%] \newline
\textbf{Lubrication} \newline
\textbullet~Friction between die and workpiece ($Fr$)\newline
\textbf{Blank Geometry} \newline
\textbullet~ Blank thickness ($D$)
&
\textbf{Springback} \newline
\textbullet~Ratio of length to springback height \newline
\textbf{Feasibility} [$\sum$=100\%] \newline
\textbullet~Inadequate stretch [\%] \newline
\textbullet~Wrinkling  [\%] \newline
\textbullet~Wrinkling tendency [\%] \newline
\textbullet~Safe  [\%] \newline
\textbullet~Risk of cracks  [\%] \newline
\textbullet~Severe thinning  [\%] \newline
\textbullet~Cracks [\%] \newline
\textbf{Change of thickness} \newline
\textbullet~Max thickening [\%] \newline
\textbullet~Max thinning [\%] \newline
\textbf{Blank Draw-In} \newline
\textbullet~Geometric boundary conditions [\%] \\
\bottomrule
\end{tabularx}
\end{table}
\cref{tab:parameters} shows a summary of process parameters, which can be categorized into three groups: Fixed process parameters, variable process parameters, and process target parameters. 
 The fixed design parameters represent parameters that should not be changed due to the process and requirements. Typically, these parameters are only altered in exceptional circumstances, for instance, when desired outcomes cannot be achieved through adjustments to variable process parameters. Changing fixed process parameters is often associated with high costs or may be constrained by product requirements. For instance, the yield strength $Rp$ can be only altered according to the material files. Other examples of fixed process parameters include material composition, die geometry, or blank thickness. 
In contrast, variable process parameters are those parameters that can be adjusted or optimized to achieve desired outcomes in the forming process. These parameters may include parameters such as pressure, lubrication, and draw beads. They are subject to modification based on the specific requirements of the forming operation and the desired properties of the end product. 
 In this paper, we refer to them as \emph{input design parameters}, which are predicted later with the approach in \cref{sec:dl,sec:approach} to yield optimal target parameters. 
The constraints represent range restrictions of the process parameters to ensure the practical feasibility. Finally, target parameters represent the desired quality characteristics or performance metrics of the formed product. These parameters serve as objectives or goals to be achieved by optimizing variable process parameters. Examples of process target parameters include mainly the feasibility and formability of the product, where the goal is to maximize the safe region and minimize inadequate stretches, risk of cracks, and wrinkling. They define the criteria for evaluating the success of the forming process and guide the selection of optimal process parameters.
Note that the feasibility target values are referred to as $L1$ through $L7$, corresponding to the fields of inadequate stretch to cracks (from top to bottom), respectively. 
Furthermore, their sum yields 100\%. Blank Draw-In considers geometric boundary conditions that require modification of the geometry itself, and is thus not considered for this paper.

\subsection{Simulation Software OpenForm}\label{subsec:openform}

\begin{figure}
    \centering
    \includegraphics[width=0.98\columnwidth]{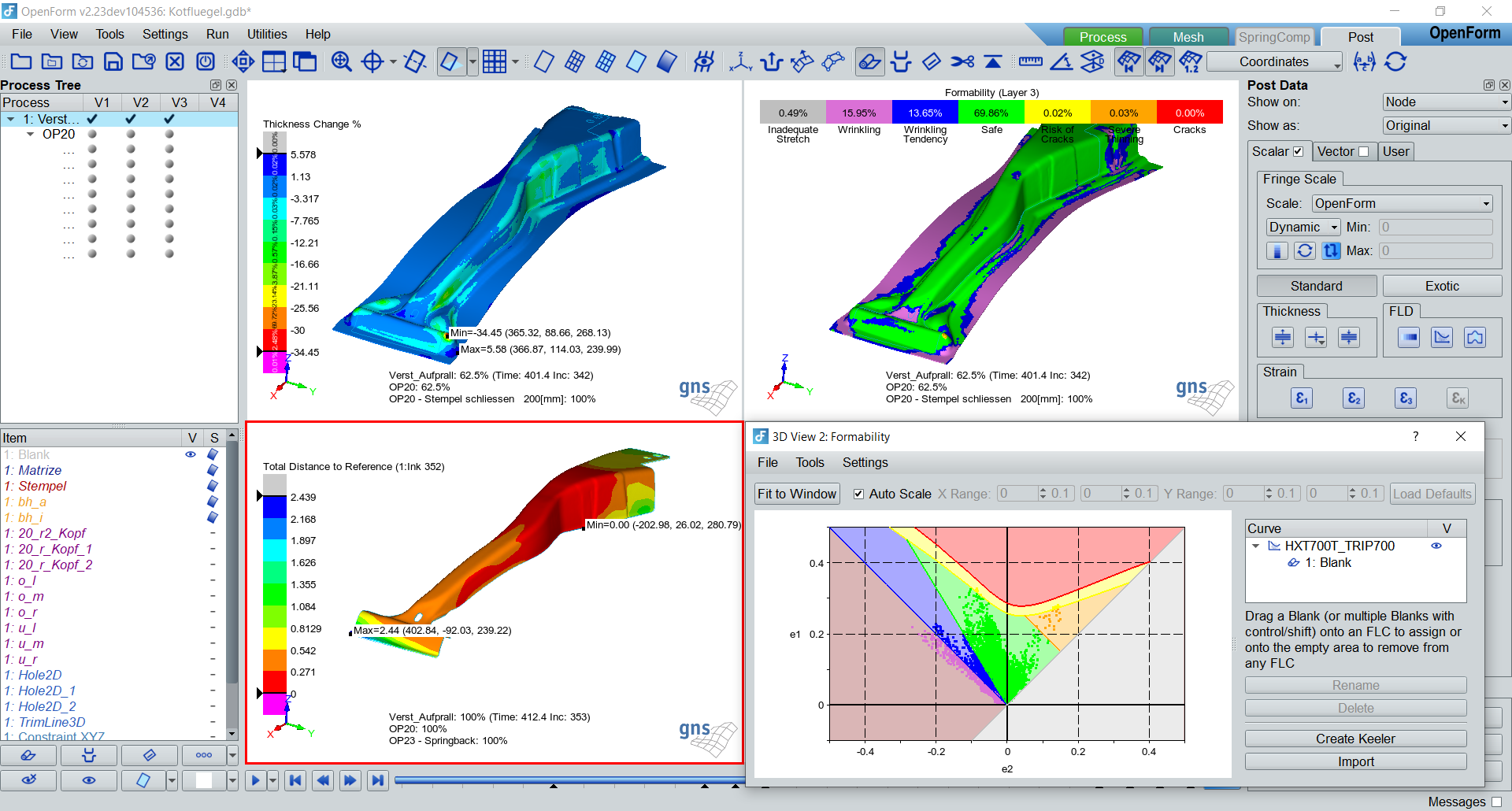}
    \caption{OpenForm GUI.}
    \label{fig:ofm_gui}
\end{figure}

OpenForm is a simulation software system\footnote{\url{https://www.gns-mbh.com/products/openform/}}
 that enables a solver-independent setup for the numerical simulation of metal forming processes. It is based on the OpenForm Process Language (OFPL) and includes a solver-independent evaluation component, enabling the comparison of results across a wide range of commercially available simulation tools, as well as measurement software. 
In addition, an in-house developed Finite Element solver for the numerical simulation of metal forming processes called OFSolv 
is integrated into the OpenForm package.
\cref{fig:ofm_gui} shows the OpenForm GUI, which allows control of the simulation and visualization of the results.

Based on the implicit Finite Element Method, OFSolv is designed to solve highly nonlinear problems, including large deformations and rotations, elastoplastic material behavior, and contact problems. 
The run-time efficiency of the software is achieved through both adaptive process time control and adaptive mesh refinement during the simulation.
The code is highly vectorized and parallelized for shared memory computing based on OpenMP. The primary parameter influencing computation time is the mesh size, i.e., the number of nodes/elements in the deformable mesh; therefore, parallelization is primarily realized by processing very long node/element vectors and arrays (parallel DO loops). By using the above-mentioned adaptive mesh refinement during a simulation, guided by the principle “as coarse as geometrically possible and as fine as physically necessary”, the task of code parallelization becomes quite challenging.

The basic workflow of such an implicit Finite Element (FE) simulation can be described in three steps repeated iteratively for every time step of the process.
In the first phase, the \textit{assembly phase},  the FE simulation constructs an overall system stiffness matrix from the individual element stiffness matrices and the associated load vector, which accounts for approximately 30\% of the total computation time. Then, in the \textit{solving phase}, the tools solve the very large system of equations of the form $A \cdot x-b = 0$, taking about 50\%-60\% of the total computation time. In the last phase, named \textit{recovery phase}, the FE simulation computes derived quantities based on the above solution vector, checks the convergence and contact state, and updates internal state variables. This phase accounts for approximately 15\% of the total computation time.
The deep drawing simulation of a common realistic car body part with a drawing depth (punch traveling distance) of 100\,mm (a rather flat part) usually requires between 100 and 200 time steps (often also called increments) and about 20 to 40 iterations per time step, so the equation system must be assembled and solved between 2,000 and 8,000 times. The typical size of the equation systems in such a process ranges from 500,000 to 3 million rows (and equally many columns).

%% file: content/03_dl.tex
\section{Deep learning Component}
\label{sec:dl}
As illustrated in \cref{fig:new_workflow}, our AI-assisted workflows utilizes two main components: 
a deep learning component and an active learning loop. In this section, we handle the deep learning component, emphasizing its role in our workflow. 

\begin{figure}[btp]
    \centering
    \includegraphics[width=.8\columnwidth,clip,trim=0 40 180 50]{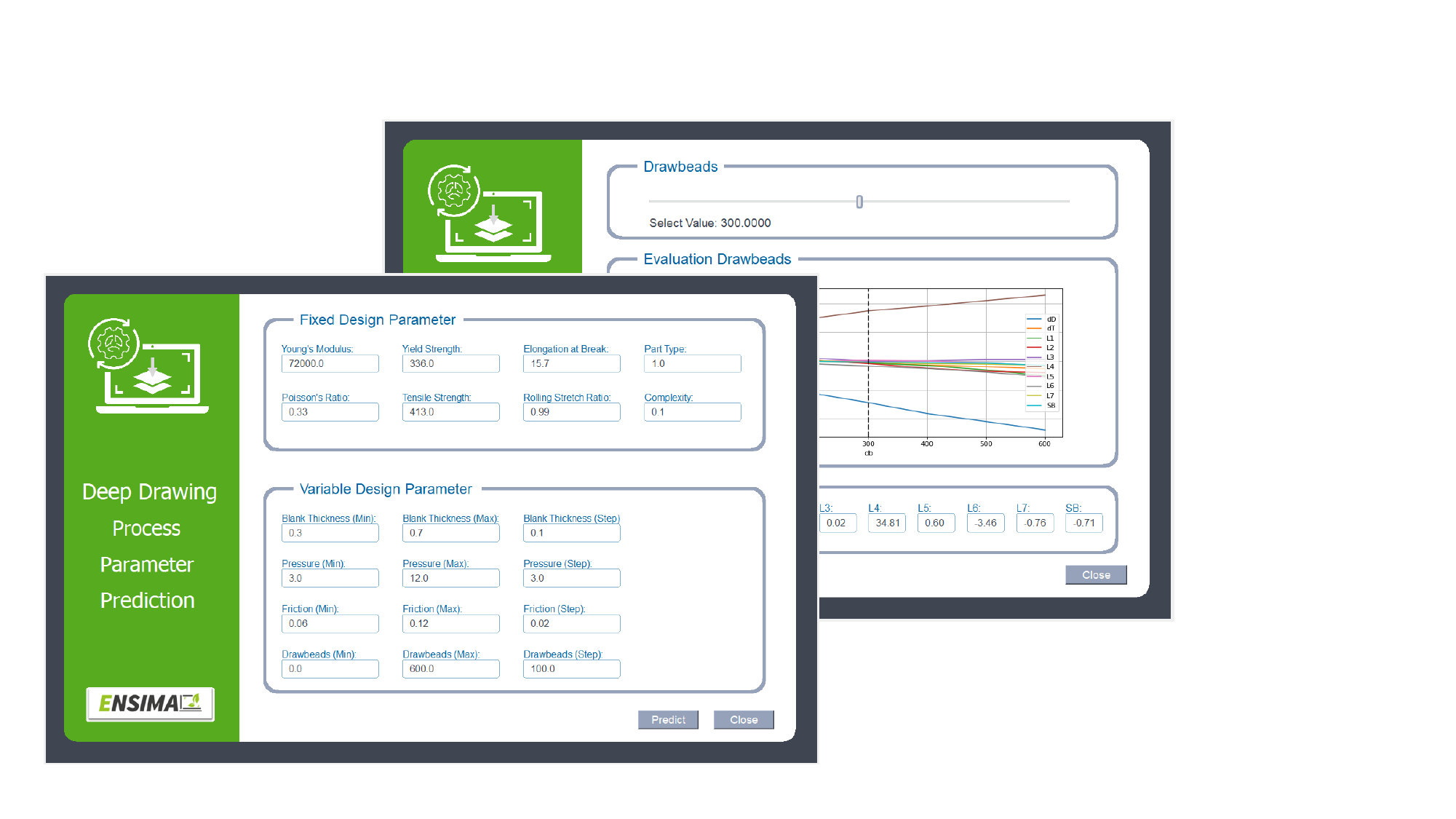}
    \caption{GUI to control the workflow.}
    \label{fig:gui}
\end{figure}

As running numerical simulations is both costly and time-consuming, the initial step of our workflow uses a deep neural network (DNN) that attempts to predict the ideally optimal input configurations that yield target parameters within acceptable ranges.
This component suggests multiple potential initial process parameters based on ranges for the target parameters (e.g., $L1$ till $L7$) and constraints (e.g., valid ranges for the input parameters). Ideally, these input process parameter configurations should contain the optimal solution. 
A GUI allows specifying these ranges and presents the results to the user as shown in \cref{fig:gui}. 
Using one Finite Element simulation, the workflow verifies the predicted process parameters. If the targets are not reached, the user can start the optimization loop through the GUI (blue box in \cref{fig:new_workflow}) to find the optimal solutions, as handled later in \cref{sec:approach}. 

\subsection{Design Parameters}
Staring with a well-chosen set of initial design parameters can help streamline the process and reduce the computational cost. The objective of the deep learning model learns to predict optimal values for these parameters based on the input parameters and historical data, aiming to maximize the likelihood of achieving the desired quality characteristics or performance properties of the formed product. The deep learning model receives as inputs material properties, abstracted geometry, process parameters, and lubrication, while the variable design parameters are treated as outputs. 
\cref{tab:deep_learning_parameters} presents an overview of the parameters employed as inputs and outputs in the deep learning model. 
\begin{table}[bhp]
\centering
\caption{Summary of input and output parameters of deep learning model.}
\label{tab:deep_learning_parameters}
\scriptsize
\begin{tabularx}{\columnwidth}{X X}
\toprule
\textbf{Inputs} & \textbf{Outputs}  \\
\midrule
\textbf{Material property} \newline
\textbullet~Young's modulus\newline
\textbullet~Poisson's ratio\newline
\textbullet~Yield strength\newline
\textbullet~Tensile strength\newline
\textbullet~Elongation at break\newline
\textbullet~Stretch ratio during rolling\newline
\newline
\textbf{Geometry} \newline
\textbullet~Blank thickness\newline
\textbullet~Part type\newline
\textbullet~Complexity\newline
\newline
\textbf{Lubrication} \newline
\textbullet~Friction coefficient\newline

\textbf{Process parameter} \newline
\textbullet~Drawbead forces\newline
\textbullet~Blankholder pressure\newline
&
\textbullet~Thinning \newline 
\textbullet~Thickening \newline 
\textbullet~Inadequate stretch \newline 
\textbullet~Wrinkling \newline 
\textbullet~Wrinkling tendency \newline 
\textbullet~Safe region \newline 
\textbullet~Risk of cracks \newline 
\textbullet~Severe thinning \newline 
\textbullet~Cracks \newline 
\textbullet~Spring back \newline 
\\
\bottomrule
\end{tabularx}
\end{table}

Although most input parameters are geometry-agnostic (e.g., material properties, friction coefficient), this has significant advantages enabling generalization across geometry variants while avoiding geometry-specific retraining, as it allows the prediction of process parameters for new geometry variants that are not yet part of the training dataset without the need to explicitly train on these new geometry variants. Instead of using full CAD geometry, a set of representative scalar features is encoded:
\begin{itemize}
\item Blank thickness: initial sheet thickness before forming.
\item Part type: a categorical classification identifying the general shape class of the part (e.g., flat, beam, deep).
\item Complexity: a scalar indicator describing how geometrically intricate the part is.
\end{itemize}

By framing the problem in this way, the deep learning model can effectively learn the complex relationships between the input parameters and output parameters, enabling the automated optimization of the forming process to meet specified quality standards or performance criteria.
Unlike the approach in \cref{sec:approach}, which considers the geometry of the part to be formed, the deep learning model uses an abstracted geometry (i.e., complexity, as shown in \cref{tab:deep_learning_parameters}). This approach enables the capture of geometric effects while maintaining flexibility across different geometries, thus enhancing the adaptability and generalization of the predictive model.

\subsection{Architecture}
The deep learning component consists of fully connected feed-forward neural networks (Multilayer Perceptrons, MLPs), implemented using the Keras API within the TensorFlow framework. Unlike the classical approach of training one MLP per output parameter, we constructed a single MLP to predict multiple target parameters simultaneously, as these parameters are interdependent and exhibit correlated behaviors. This approach enables the network to capture relationships between outputs, improving prediction accuracy and consistency. 
The network has the following architecture: 
\begin{itemize}
\item Input layer: dimension = 12 (including material properties, geometry, lubrication, process parameter)
\item Hidden layers: six fully connected layers with different numbers of neurons and ReLU activations. The numbers of neurons in these hidden layers are 256, 20, 420, 100, 184, and 300, respectively.
\item Output layer: dimension = 10 (corresponding to the number of output parameters) with linear activation
\end{itemize}

Hyperparameters, including the number of hidden layers, units per layer, dropout rate, and learning rate, were optimized using the Keras Tuner library. Specifically, the tuner used random search over the search space, evaluating 30 different hyperparameter configurations, with each configuration trained for up to 50 full passes over the training dataset. The Adam optimizer was used, with the learning rate explored over the range $1 \times 10^{-4}$ to $1 \times 10^{-2}$. Training employed a batch size of 64 and a maximum of 100 epochs, with early stopping to prevent overfitting and reduce unnecessary computation with a patience of 10 consecutive epochs. During the fine-tuning process, the model achieving the lowest validation loss was selected for evaluation. As a result of the hyper parameter tuning process, the selected model comprised six fully connected hidden layers with a learning rate of 0.001. 

\subsubsection{Experiment}
The dataset consists of 859 samples generated from validated forming simulations. These samples include 10 different part shapes and a wide range of process parameters. The dataset was split into a training set (80\%), a validation set (10\%), and a test set (10\%). Additionally, a separate experiment using three new part shapes was conducted, with a total of 315 samples, to evaluate the model's ability to generalize to unseen part geometries. \cref{tab:dl_result} summarizes the performance of the deep learning model when the model was trained on 859 samples and evaluated on three previously unseen parts. The model demonstrates that ability to generalize effectively to new part geometries. This is likely because the model has successfully captured the underlying structural behaviors and interdependencies among parameters across the various part shapes. More details are provided in~\cite{park_accelerating_2025}. 

\begin{table}[bhp]
\centering
\caption{Evaluation results for train, validation, and test sets, along with three new parts: \emph{Size} indicates the number of samples (variants). \emph{MAE} is Mean Absolute Error, \emph{MSE} is Mean Squared Error, and \emph{RMSE} is Root Mean Squared Error.}
\label{tab:dl_result}
\scriptsize
\begin{tabularx}{\columnwidth}{p{1.6cm} X X X X}
\toprule
\textbf{Dataset } & \textbf{Size} & \textbf{MAE} 
& \textbf{MSE} & \textbf{RMSE} \\
\midrule
\text{Train} \newline
\text{Validation} \newline
\text{Test} \newline
\text{New parts} 
&
\text{687} \newline 
\text{86} \newline 
\text{86} \newline 
\text{315} 
&
\text{7.08} \newline 
\text{6.87} \newline 
\text{6.61} \newline 
\text{6.90} 
&
\text{153.84} \newline 
\text{155.17} \newline 
\text{131.41} \newline 
\text{133.80} 
&
\text{12.40} \newline 
\text{13.46} \newline 
\text{11.46} \newline 
\text{11.57} 
\\
\bottomrule
\end{tabularx}
\end{table}


%% file: content/04_approach.tex
\section{Optimization Loop of the Workflow}
\label{sec:approach}
\label{sec:BO}
As the optimal solution might not be found with the initial configuration obtained with the deep learning component from~\cref {sec:dl}, our approach employs active learning (in particular Bayesian optimization~\cite{RG23,BO1shahriari2015taking,BO2frazier2018tutorial,BO3wang2023recent,BO4wilson2018maximizing}) to select the most promising points from the input space that yield the best target parameters. 
In this section, we describe these aspects, as well as the human-guided optimization, in detail. 

To achieve higher accuracy while using less training data, our approach employs active learning, a methodology that enables a machine learning algorithm to select its own training data~\cite{settles2009active}. In particular, we use Bayesian optimization.
 In Bayesian optimization, previous simulations (i.e., the labeled data consisting of input parameters and their corresponding outputs) are used to train a surrogate model that approximates the true objective function.
At each iteration, an acquisition function evaluates the specified input points and assigns a score that reflects how promising each point is for finding the global minimum or maximum of the objective function.

\begin{figure}[tbp]
\centering
\includegraphics[width=\columnwidth,clip, trim= 0 60 210 110]{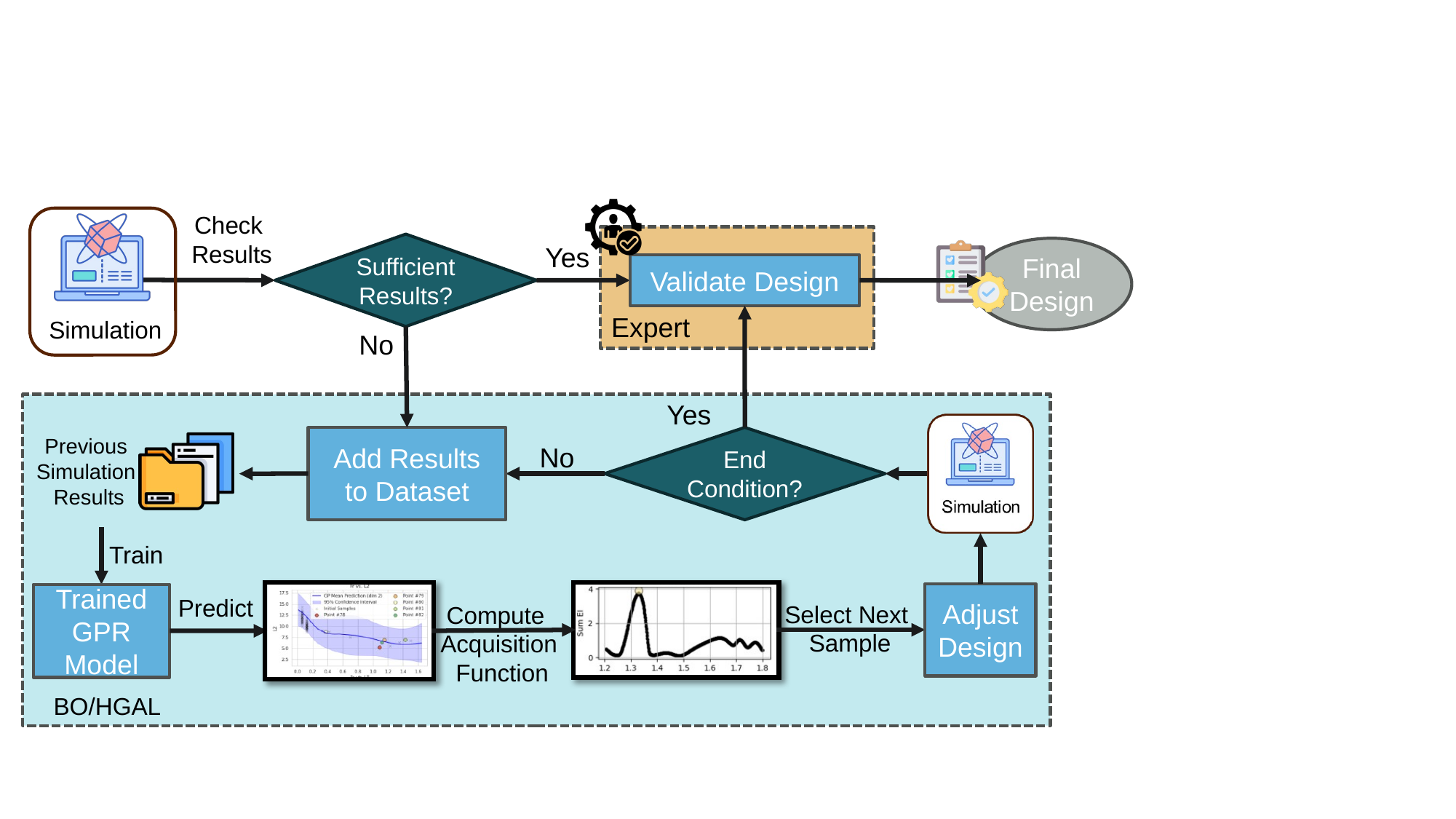}
\caption{Optimization loop supporting Bayesian optimization or human-guided active learning for selecting promising input configurations.}
\label{fig:activelearning}
\end{figure}
Our approach supports various configurations, models, and latent dimension transformations.  \cref{fig:activelearning} illustrates the basic approach, which simply employs a multivariate Gaussian Process (GP) surrogate model that directly uses the input configuration to predict the output parameters. 
As shown, if insufficient results are obtained from the deep learning component (\cref {sec:dl}), the user can trigger the optimization loop from the GUI or console. This loop is executed iteratively until the total specified iterations or one of the end conditions (\cref{subsec:end_conditions}) is reached. At each iteration, the results (input parameter configurations and output values from the simulation) are appended to the file that contains the results from previous simulations. This file is used to train the GP model as described in \cref{subsec:model}. Based on the trained model, the acquisition function (\cref{subsec:acq_func}) is created and evaluated at specific points from the input space (\cref{subsec:input_space}). The most promising points are then selected, while balancing exploration and evaluation as described in \cref{subsec:acq_func}. Afterwards, the input configurations are embedded into the required simulation files (see \cref{subsec:adapting_the_design}), and the numerical simulation with OFsolve is executed. Once the simulation finishes, OpenForm is launched to extract the target parameters from the simulation results. These results, along with the input configuration, are appended to the result file again, and the loop continues with the next iteration until the specified total iterations or end conditions (\cref{subsec:end_conditions}) are reached. In what follows, we examine our approach in detail, focusing on these components.  Note that Python code controls the entire optimization process, including launching the simulation in a dedicated process, handling end conditions, and training models.

\subsection{Surrogate Model}
\label{subsec:model}
As our use case involves multiple input parameters and target values, we employ a multivariate Gaussian Process surrogate model. This surrogate model is implemented using two different backends: GPyTorch for scalable multi-task modeling with latent inputs and outputs, and scikit-learn for simpler cases. By default, our approach utilizes the GPyTorch version, as this enables easy switching between the devices used for training (CPUs and GPUs). 

A Gaussian Process is a stochastic process in which every finite subset of variables follows a multivariate Gaussian distribution.
In principle, a stochastic process is a collection of random variables, specified by providing the probability distribution for every set of finite variables in a consistent manner. Hence, a Gaussian Process can be seen as a generalization of the Gaussian distribution to functions, and it is fully specified by a prior mean function $\boldsymbol{\mu}(\mathbf{x})$ and covariance function $\mathbf{K}(\mathbf{x},\mathbf{x}')$ for any two input vectors $\mathbf x,\mathbf x' \in  \mathbb{R}^d$:
\begin{equation}
\mathbf{f}(\mathbf{x}) \sim \mathcal{GP}(\boldsymbol{\mu}, \mathbf{K})
\label{eq:gp}
\end{equation}
With: 
\begin{equation}
\begin{aligned}
\boldsymbol{\mu}(\mathbf{x}) &= \mathbb{E}[\mathbf{f}(\mathbf{x})] \\
\mathbf{K}(\mathbf{x},\mathbf{x}') &= \mathbb{E}\big[(\mathbf{f}(\mathbf{x})-\boldsymbol{\mu}(\mathbf{x}))(\mathbf{f}(\mathbf{x}')-\boldsymbol{\mu}(\mathbf{x}'))^\top\big] 
\end{aligned}
\end{equation}

As \cref{eq:gp} implies, the matrix of true function values $\mathbf Y = [\mathbf{f}_1, \dots, \mathbf{f}_n]^\top \in \mathbb{R}^{n\times m}$ at the training inputs $\mathbf{X} = [\mathbf x_1, \dots, \mathbf x_n]^\top\in \mathbb{R}^{n\times d}$ follows a multivariate Gaussian distribution. 
For the multivariate case, the prior mean function $\boldsymbol{\mu}(\mathbf{x}) = \mathbb{E}[\mathbf{f}(\mathbf{x})]$ maps 
each input vector $\mathbf{x} \in \mathbb{R}^d$ to the expected output vector, while the covariance function $\mathbf{K}(\mathbf{x},\mathbf{x}') \in \mathbb{R}^{m \times m}$ describes the uncertainty and correlations between outputs.
The key idea of Gaussian process regression is to make predictions without explicit parameterization. 
For that, the GP prior together with the observed training data is used to make predictions and the unseen test points. 
With the true function values  $\mathbf{Y}$ at the training input $\mathbf X$, and the function values $\mathbf{Y}_{\ast}$ corresponding to set of test inputs $\mathbf{X}_\ast$,  Gaussian process regression computes posterior distribution of 
$\mathbf{Y}_{\ast}$  conditioned on the training data ($\mathbf X$ and $\mathbf Y)$.
\begin{equation}
\begin{aligned}
p(\mathbf{Y}_\ast \mid \mathbf{X},\mathbf{Y}) =\  &\mathcal{N}\Big(
\boldsymbol{\mu}_\ast + \mathbf{K}_\ast^\top \mathbf{K}^{-1} (\mathbf{Y}-\boldsymbol{\mu}),\\
&\mathbf{K}_{\ast\ast} - \mathbf{K}_\ast^\top \mathbf{K}^{-1} \mathbf{K}_\ast 
\Big)
\end{aligned}
\label{eq:post}
\end{equation}
With:
\begin{itemize}
    \item $\boldsymbol{\mu}\in \mathbb{R}^{n \times m}$ and $\boldsymbol{\mu}_\ast \in \mathbb{R}^{n_\ast \times m}$: prior means at the training and test points
    \item $\mathbf{K}=  \mathbf{K}(\mathbf{X}, \mathbf{X}) \in \mathbb{R}^{(n m) \times (n m)}$: block covariance matrix for the training points, where each block $\mathbf{K}(\mathbf{x}_i, \mathbf{x}_j) \in \mathbb{R}^{m \times m}$ represents the covariance between outputs at inputs $\mathbf{x}_i$ and $\mathbf{x}_j$.
    \item $\mathbf{K}_\ast = \mathbf{K}(\mathbf{X}, \mathbf{X}_\ast)$: covariance between training and test points 
    \item $\mathbf{K}_{\ast\ast} = \mathbf{K}(\mathbf{X}_\ast, \mathbf{X}_\ast)$: covariance  between test points
\end{itemize}

Note that \cref{eq:post} represents the simple noise-free case.  In practice, the true function $\mathbf{f}$ is unknown, and we only have noisy observations  $\mathbf{f} + \boldsymbol{\epsilon}$ with $ \boldsymbol{\epsilon} \sim \mathcal{N}(\mathbf 0, \sigma_n^2 \mathbf I)$.
In simple terms, Gaussian process regression provides a Gaussian distribution for each test point, providing both a mean and a measure of uncertainty. The uncertainty is especially valuable for active learning as it is used to guide the selection of the next samples. 

Our approach provides three configurations to model multiple correlated outputs with different treatments of the mean and covariance functions: \textit{independent}, \textit{coupled mean}, and \textit{LCM} (Linear Model of Coregionalization). All configurations use the Matern kernel as the base covariance function but differ in how they model the mean structure and correlations between outputs.
In the \textit{independent} configuration, each output has an independent zero-mean function ($\boldsymbol{\mu}(\mathbf{x}) = \mathbf{0}$), and all outputs share a single Matern kernel. This configuration assumes that outputs share a covariance structure but not mean values. No latent structure is used for the outputs; however, an optional latent input transformation via a neural network encoder can be applied to capture nonlinear dependencies in the inputs. 
In contrast, both the \textit{coupled mean} and \textit{LCM} configurations incorporate latent coupling across outputs in the mean function, in addition to a latent input transformation. 
Thus, instead of each output having its own independent mean, all output means are generated from a shared latent space via a small neural network projection, which enforces a shared structure among outputs rather than treating them independently.

The \textit{coupled mean} configuration employs a single Matern kernel shared across outputs, hence, correlating outputs only through the mean. Conversely, the \textit{LCM} configuration introduces multiple latent Gaussian processes, each associated with its own Matern kernel, and models each output as a linear combination of these latent functions.
In particular, the latent output space serves two purposes for this configuration: 1) as previously, it is used in the latent mean projection to couple the output means, and 2) it defines the latent functions in the LCM module that are linearly combined to produce the outputs. 
This provides a more expressive representation capable of modeling complex inter-output correlations. By default, we use the  \textit{LCM} configuration as this yields the best results and the target parameters from the forming process are strongly correlated. For the remainder of the paper, we focus solely on the \textit{LCM} case, as the other configurations are special cases of this setup. 

The \textit{LCM} configuration is realized using GPyTorch, including the LCMKernel with latent output support for the covariance, MaternKernel for each latent function, a small neural network for the latent mean, and a latent input encoder (neural network) that nonlinearly transforms the original inputs before feeding them to the GP. Furthermore, \cref{eq:post} is realized using the MultitaskMultivariateNormal module from the same library.

By examining the data from various simulation runs, we observed that there are hidden or non-observable variables that contribute to a specific structure in the data. On one hand, this results from the fact that we often consider only a subset of the design parameters and ignore the fixed design parameters. On the other hand, this also results from the fact that the forming process is a strongly nonlinear simulation, which introduces latent factors that influence the results. To compensate for these aspects, the latent transformations enable the capture of correlations between outputs or complex, nonlinear effects of the inputs that are not explicitly modeled in the observed data, allowing the Gaussian Process models to provide more accurate predictions. By default, the latent inputs are twice as many as the original inputs, while the latent output dimensions are the same as the original output dimensions. A model class wraps these transformations alongside the configuration for the mean and covariance, such that the inputs and outputs to the model are those from the original space (i.e., input and target parameters of the forming process).

During training, the model optimizes the hyperparameters of the covariance (Matern) kernels (e.g., lengthscale, variance, etc.), the latent input encoder (i.e., neural network weights for transforming the inputs), the latent coupled mean network weights (i.e., the neural network projecting the latent representation to output means), and the latent output projection weights that combine the latent Gaussian processes in the LCM kernel to produce the final correlated outputs. During this process, the training data plays a key role in our approach. The result file, which gathers input and target parameters from previous sheet metal forming simulations, hosts these values for various parts (e.g., doors, car roofs, front fenders). To train our model with this  dataset, our approach supports three mechanisms:
\begin{inparaenum}
    \item Use all the data contained in the result file
    \item Filter the points according to their complexity 
    \item Filter the points according to the part
\end{inparaenum}
Using all the data would train the model with incorrect values, and this is thus avoided. The second option is usually chosen, especially for new parts, no data is available. In a typical process, after a few iterations with the second option, the third option can be used, as Bayesian optimization with a GP model is usually efficient even with a few samples. Alternatively, the third option can be used by randomly selecting the next samples until enough points are available. Considering that numerical simulation can be very expensive, none of the three mechanisms provides optimal results, especially for new parts. 
To overcome this limitation, a mixture of experts can be used, as described in \cref{subsec:moe}, to guide the search when not enough training points are available for new parts. Regardless of the model used, the next step in Bayesian optimization is to evaluate the acquisition function using the model to determine the most promising input configuration that yields the desired target values. For that, the input space hosting all possible input configurations is needed, which is generated as explained next.

\subsection{Input Space}
\label{subsec:input_space}
The size of the sampling space plays a key role in our approach. Logically, the range and accuracy (i.e., precision) of the search space directly influence the results, as they limit the candidates for optimal results. For each input variable, its type (continuous or discrete) and optional range constraints can be specified. Next, the result file is read, and the ranges (for continuous variables) or valid values (for discrete variables) are extracted. Our approach adds an expansion factor (default 10\%) to these ranges to increase the exploration ranges during optimization. In the next steps, the constraints are considered to shrink or expand the continuous ranges, or discard or add discrete values. 

To compute the space of input configurations $\mathbf X_\ast$, we provide two methods: \emph{Linear} or \emph{combination}. The linear (default) method takes a fixed number of steps (default $n_\ast=10000$) to create an array 
bounded by the constraints and ranges (with 10\% expansion factor)  for each dimension and simply stacks them together to create $\mathbf {X}_\ast$. In contrast, 
the combination method takes the full Cartesian product across all dimensions to generate all possible combinations of per-dimension values. 
As with the linear method, it respects the ranges and constraints of each dimension.
However, the combination method yields a huge number of possible input configurations, which scales exponentially with the number of dimensions. 

To limit the number of possible input configurations, our approach supports an option to randomly select a specified number of samples from the pool of available options. Yet, as this solution might miss optimal  configurations, a more robust alternative is to shrink the step count per dimension to increase control over the selection process. Additionally, 
To limit the size with both the linear and the combination method, the precision of the $\mathbf {X}_\ast$ can be specified. 
We use the linear method in this paper as the default, as it yields sufficient results. Note that in both cases, we evaluate the expected improvement in batches with the surrogate model to decrease the memory footprint. 

\subsection{Selecting the Next Sample}
\label{subsec:acq_func}
\label{subsec:next_sample}
A key aspect of Bayesian optimization is selecting the next sample point, i.e., the input configuration that will yield the global optimum of an unknown objective function (in our case, the minimum). 
However, as executing the simulation (true objective function) is too expensive, the optimization process uses the surrogate model from the previous section to approximate the true objective function while providing uncertainty in unexplored regions. To guide the selection of the most promising sample,  acquisition functions are used, which are  inexpensive functions that allow for evaluating the potential of a point in yielding the global optimum. 
Given an acquisition function and a candidate point $\mathbf x_\ast \in \mathbf{X} _\ast$, the prediction distribution according to \cref{eq:post} is. Consequently, the prediction (posterior) mean $\boldsymbol \mu(\mathbf{x}_\ast)\in \mathbb{R}^m$ and covariance  matrix $\mathbf{K}(\mathbf{x}_\ast)\in \mathbb{R}^{m\times m}$ are:
\begin{equation}
\begin{aligned}
\boldsymbol{\mu}(\mathbf{x}_\ast) &= \boldsymbol{\mu}_\ast + \mathbf{K}_\ast^\top \mathbf{K}^{-1} (\mathbf{Y} - \boldsymbol{\mu}), \\
\mathbf{K}(\mathbf{x}_\ast) &= \mathbf{K}_{\ast\ast}(\mathbf{x}_\ast,\mathbf{x}_\ast) - \mathbf{K}_\ast^\top \mathbf{K}^{-1} \mathbf{K}_\ast,
\label{eq:posterior}
\end{aligned}
\end{equation}

Several sampling strategies exist in this context, including the upper confidence bound~\cite{srinivas2009gaussian}, probability of improvement, and expected improvement (EI). For our implementation, we select expected improvement because, similar to the upper confidence bound, it balances exploration (examining uncertain regions) and exploitation (focusing on regions with high predicted values). Unlike the probability of improvement, which treats all improvements equally and can often get stuck in local optima, the expected improvement incorporates the magnitude of the improvement into the acquisition function. There are two approaches to compute the expected improvement: Either consider the full covariance matrix and estimate the expected improvements jointly across outputs using Monte Carlo sampling with a specified number of points, or compute the marginal variance ($\boldsymbol \sigma^2=diag(\mathbf{K}(\mathbf{x}_\ast))$ to optimize the outputs independently. The first option draws random samples (default $n_{mc} = 10000$) from the multivariate distribution given by \cref{eq:posterior} for each input and computes the  improvement over the current best output $f^\ast$ for each sample: 
\begin{equation}
\mathbf{EI}(\mathbf x_\ast) = \frac{1}{n_{mc}}\sum_{i=1}^{n_{mc}}max(\mathbf f_\ast^{(i)} - \mathbf f^\ast)
\label{eq:nc}
\end{equation}
For the latter option, the expected improvement is:
\begin{equation}
 \mathbf{EI}(\mathbf x_\ast) = (\mathbf f^{\ast} - \boldsymbol \mu)\boldsymbol \Phi(\frac{\mathbf f^{\ast} - \boldsymbol\mu}{\boldsymbol \sigma})+\boldsymbol\sigma\boldsymbol\phi(\frac{\mathbf f^{\ast} - \boldsymbol \mu}{\boldsymbol \sigma})
 \label{eq:ei_matginal}
\end{equation}
where $\boldsymbol\phi$ and $\boldsymbol\Phi$ are the standard normal probability density and cumulative distribution functions, respectively. While this method ignores the correlation between outputs, it is significantly faster to compute and lowers the overhead in successive steps.  For our purpose,  rather than optimizing towards a global minimum, we specify the target $f^\ast$ that contains the desired L1 to L7 values. As we desire high values for L4 and low values for L7, L1, and L6, we further scale the variables by an \textit{attention} vector, reversing the sign of L4 in the process. Consequently, our optimization problem finds the maximum value of L4 with mid priority, and minimizes  L7, L1, and L6 with a higher priority.

With the input space $\mathbf {X}_\ast \in \mathbb{R}^{n_\ast\times d}$ at hand (\cref{subsec:input_space}), the expected improvement can be computed (using \cref{eq:ei_matginal} or \cref{eq:nc}) to obtain $\mathbf{EI}(\mathbf X_\ast)\in \mathbb{R}^{n_\ast \times m}$. Consequently, each output can be optimized separately. As mentioned at the end of \cref{subsec:next_sample}, we exploit this aspect through scaling the important target values. Afterwards, the sum of $\mathbf{EI}$ across all outputs is computed to obtain a vector of size $n_\ast\times1$. Our algorithms select the next sample point as the one with the highest expected value. This ensures that the next sample is the most promising across all objectives. While this strategy appears purely exploitative, the iterative nature of Bayesian optimization naturally handles exploration. Once a sample is selected, the uncertainty in this area drops, suppressing the expected improvement and forcing the algorithm to search in other promising regions in the next iteration. To further balance the exploitation and exploration aspects of our approach, we introduce parallel samples as explained next.

\subsection{Parallel Samples}
\label{subsec:parallel_samples}
When the number of samples selected in an iteration is set to one, only one value is drawn from the sum of expected improvement across the outputs as described in \cref{subsec:next_sample}. To improve the exploration aspects of the approach, $p$ executions can be conducted in parallel (using futures) that draw and execute the simulation with other promising input configurations, in addition to the input configuration with the highest exploration from \cref{subsec:next_sample}. For that, our approach supports three selection strategies: \emph{Highest sum}, \emph{peak-based}, and \emph{crowding distance}. Both the \emph{highest sum} and \emph{peak-based} strategies operate on the sum of expected improvement across the outputs to select the next input configuration with either the highest values or the highest peaks, respectively. 
In contrast, the \emph{crowding distance} operates directly on $\mathbf{EI}$. This strategy computes the crowding distance~\cite{996017} for each candidate point $x_\ast$:
\begin{equation}
\text{CD}_i = \sum_{j=1}^{m} \frac{EI_{i+1,j} - EI_{i-1,j}}{EI_{\max,j} - EI_{\min,j}}, 
\quad i \in [2, \dots, n_\ast-1]
\label{eq:crowding_distance}
\end{equation}
With $EI_{\max,j}$ and $EI_{\min,j}$ representing the max and min values along the target dimension. Furthermore, the crowding distance at the first and last point ($n_\ast$) is set to infinity. As observed in \cref{eq:crowding_distance}, the greater the isolation of the candidates, the higher their crowding distance, thereby increasing spatial diversity across the parallel samples. 

Parallel samples improve the exploration aspect of our approach while potentially reducing the time-to-solution. However, considering that the best sample has already been selected based on the highest exploitation (\cref{subsec:next_sample}), selecting alternative input configurations from the same expected improvement might only yield suboptimal results. 
While considering multiple samples simultaneously bears this risk, it can significantly reduce the time to solution. Moreover, considering the scaling behavior of OFsolve, this approach can also lower energy consumption, since running a single OFsolve instance on all available cores scales less efficiently than running several instances in parallel, each using a fraction of the total cores, as the results later demonstrate. 
Additionally, the exploration of parallel samples may introduce new, promising regions due to the strongly non-linear behavior of the simulation. 
As an alternative to letting the algorithm select the most promising sample (see \cref{subsec:next_sample}), a human can intervene to guide the optimization, as described next. 

\subsection{Human Guided Active Learning}
\label{subsec:HGAL}
Our approach enables a human to intervene in the optimization loop, guiding the selection of the next input sample. 
To aid the human during this process, our approach computes the \emph{expected improvement} as in \cref{subsec:next_sample}
and provides the results to the user. More specifically, a plot for the expected improvement is generated, highlighting the last selected samples. Furthermore, the results from Bayesian optimization are also shown for each input and target output pair. The user then specifies the next desired input sample, and the optimization loop continues as usual (see \cref{fig:activelearning}). That is, the design is adjusted (\cref{subsec:adapting_the_design}) based on the provided input. Simulations with OFsolve are then executed, OpenForm extracts the target parameters from the simulation results, and the input configuration, together with the obtained target values, is appended to the results file. If the end conditions are not met, this file is used again to train the surrogate model in the next iteration, and the user is prompted again to provide the input based on the new values obtained for the expected improvement. 

\subsection{Adapting the Design}
\label{subsec:adapting_the_design}
Regardless of how the next sample (i.e., input configuration) is obtained, the next step in the optimization loop is to adapt the design and execute the simulation. For that, several files that host the parameters are modified. Depending on the parameter, this process varies significantly. For the friction $Fr$, the drawbeat force $db$, and the blankholder pressure $p$,  a single entry in the simulation file (.dat) needs to be modified. In contrast, for the blank thickness $D$, the 3D point cloud needs to be adjusted by modifying the Z-dimension. For discrete parameters, such as yield strength $Rp$, configuration files must be available that match the specified value. Consequently, our approach adapts the design based on the specified input parameters and then launches the simulation. 

OFsolve generates a set of output files upon completion of the simulation. In a traditional workflow, these files are examined by an expert inside the OpenForm GUI (see \cref{fig:gui}). The expert inspects if the target parameter has been reached and decides whether an additional simulation with an alternative input configuration is required. Our approach eliminates the need for an expert in this step by automatically examining the results based on a session file. This file specifies the target parameters that are extracted using the console version of OpenForm from the OFSolve results. Afterwards, these results, alongside the input configuration for the simulation, are appended to a result file that gathers these values from previous simulations.

An essential aspect of our approach is its modularity: the model can be easily adapted as desired (\cref{subsec:model} and \cref{subsec:moe}) and the next sample point can be computed by the workflow  (\cref{subsec:next_sample}) or specified by an expert (\cref{subsec:HGAL}). The same principle applies to the simulation. In this paper, we focused entirely on OpenForm and OFsolve. A single class wraps the calls that adjust the design and execute the simulation. Consequently, this allows for easily swapping the simulation tool if desired. In particular, a class for adjusting the design based on the input parameters and a simulation wrapper (e.g., to run a license server once the simulation starts) are needed to adapt our approach to other simulation tools. 

\subsection{End Conditions}
\label{subsec:end_conditions}
The optimization loop terminates in two cases: either when the specified number of iterations is completed or an end condition is met. Currently, our approach supports three end conditions: \emph{no improvement}, \emph{constant minimum}, or \emph{energy budget}. The end condition \emph{no improvement} stops the optimization loop if the sum of the expected improvement across all numbers of samples and outputs stays constant for a specified number of steps (default 5). Similarly, the \emph{constant minimum} condition terminates the loop once the best target parameters obtained from the simulations remain constant for the defined number of iterations. In contrast, the \emph{energy budget} condition allows the user to specify an energy budget for the simulations. Once this is exhausted, the optimization loop ends. 
The end condition \emph{no improvement} is used for this paper. 

To estimate the energy consumption during simulation, we use the theoretical energy model: 
$
E = C \cdot V^{2} \cdot f \cdot n \cdot t
$, which uses the the execution time $t$ and accounts for $n$ cores. In this model, $C$ represents the capacitance, $V$ the supply voltage, and $f$  the operating frequency of the processor. For training the model, energy consumption is measured directly from the hardware using the NVIDIA Management Library (pynvml) when GPUs are available. When GPUs are not available, the same theoretical energy model is used to estimate the energy consumption. Since the model presented in this paper is always trained on a GPU cluster (later discussed in \cref{sec:evaluation}), the first approach based on direct measurement with pynvml is used. Note that these values are displayed to the user during each iteration, and a summary is printed at the end of the optimization process, which also reports the total CO$_2$ footprint.

\section{Extended Optimization Loop}
\label{sec:extended_approach}
The approach handled in \cref{sec:approach} represents the basic approach that uses a \emph{single} 
multivariate latent-variable Gaussian Process (MLVGP) surrogate model.  Two key improvements can be applied to the approach:
\begin{inparaenum}
    \item Improve the selection of the next sample if not enough simulation results are available to create an MLVGP model, and 
    \item terminate the simulation using approximated computing.
\end{inparaenum}
The first improvement is necessary, in case a new part is considered or if there are not enough points available to train an MLVGP model. 
In this case, our approach employs a mixture of experts (\cref{subsec:moe}) that embeds the geometry of all simulated parts (\cref{subsec:embedding}) to select the relevant experts (\cref{subsec:gating}), guiding the search for the next sampling point. To save energy and avoid waiting for insufficient results, the second improvement interacts with the simulation tool (OFsolve) to terminate simulations at runtime when the target parameters fall outside the desired ranges, as described in \cref{subsec:approx_computing}.

\subsection{Mixture of Experts}
\label{subsec:moe}
As described in \cref{subsec:model}, regardless of the configuration, our approach basically employs a multivariate MLVGP surrogate model to predict latent output from latent input, which are then mapped to observed inputs and outputs. For new parts which have not been simulated, or in case not enough simulations have been executed, \cref{subsec:model} suggests applying a filter to remove entries in the result file for parts that mismatch the current one. While  the complexity filter enables realizing this aspect, grouping parts based on complexity is often insufficient, as this simple metric only considers a marginal fraction of the actual geometry (total sizes for each dimension). Ideally, the part filter would be applied; however, without sufficient simulation runs, the only option would be to randomly select the next samples. To improve this aspect, a mixture of experts (MOE) can be used to guide the selection of the next input configuration to simulate. 

In principle, MOE aggregates various expert models and utilizes gating to determine which ones to employ. Rather than training a single model with the entire data set, MOE decomposes the data into subsets and trains each expert on a subset. In our case, as the result file aggregates simulations from different parts, we create an expert for each part using the same approach as in \cref{subsec:model}. This brings the advantage that all options handled in  \cref{subsec:model} can be directly incorporated into the MOE. After creating the experts, we select specific ones for the prediction (\cref{subsec:gating}), based on the similarity of the current part's geometry to those of the experts. This is explained next. 

\subsection{Geometric Encoder}
\label{subsec:embedding}
After generating an expert for each part contained in the result file (\cref{subsec:moe}), the next task is to select which expert to use. For this task, we exploit the aspect that the geometry of each part is available. 
Once a new part is simulated, its geometry can be compared to that of other parts to identify the most similar ones. However, the fact that parts contain different numbers of points, spaced differently along the space, complicates this aspect. To overcome this limitation, we utilize a PointNet-based embedding network, which maps the 3D geometric points into a fixed-size ($k_{emb}$) embedding vector $\mathbf e$. The embedding obtained using this geometric encoder allows us to find the geometric similarity between the parts. 

The encoder consists of two parts: a PointNet network that generates the embedding, and a classifier that maps the embedding to the parts. The classifier consists of a single linear layer that transforms the $k_{emb}$ sized embedding vector into the number of parts $k_{experts}$. In contrast, the PointNet network uses a series of linear transformations followed by ReLU activations to extract features from each 3D point.
After processing all points, PointNet applies global max pooling to combine the individual point features into a single embedding vector of size  $k_{emb}$ for each part. This embedding is then fed into the classifier to produce the final part predictions. Once the optimization loop begins, both components are trained only once, as the geometric points remain constant for successive iterations. Once a new part is simulated, the encoder is used to generate an embedding vector for it. This vector is then compared to the embeddings of the parts from the MOE as explained in \cref{subsec:gating} to select the experts to use. 

The size of the embedding vector plays a key role in this comparison. A value that is too low might result in using parts  that are not similar to the current ones. On the contrary, a too high value induces unnecessary training and prediction overhead. For this purpose, our approach provides two options: \emph{up\_sample} or \emph{down\_sample}. 
In both cases, the length of geometric points across all parts is examined. 
In the \emph{down\_sample} mode, $k_{emb}$ is set to the minimum length, while the \emph{up\_sample} mode uses the maximum length across the points. Afterwards, to increase the fairness, the \emph{down\_sample} randomly selects $k_{emb}$ geometric points for each part. In contrast, the \emph{up\_sample} strategy first tries to deterministically double the number of points until their number exceeds $k_{emb}$, after which again $k_{emb}$ are randomly selected.

The approach from \cref{subsec:moe} to \cref{subsec:embedding} can be regarded as a supervised classification method, as it utilizes part-labeled data, dedicated experts, and an encoder trained on the parts. While our approach also supports an unsupervised mode that does not consider the part label, but rather uses the geometry of the parts to cluster them in groups using an autoencoder and trains experts for each group, we skip handling it in this paper for simplicity. 

\subsection{Selecting the Right Expert}
\label{subsec:gating}
The final step in the MOE approach is to select which experts to use. We implemented both hard and soft gating for this purpose. In hard gating, a single expert is selected by selecting the part that is most similar to the current one. Our approach first computes the Euclidean distance between the embeddings in the MOE ($\mathbf{e}_{a}$ with $a\in k_{experts}$) and the new  embedding of the current part $\mathbf{e}_{part}$:
\begin{equation}    
d_a= \sqrt{\sum_{i=1}^{k_{emb}} \left( e_{part,i} - e_{a,i} \right)^2 }
\label{eq:dist}
\end{equation}
Next, the minimum distance is found, and the corresponding expert is used  to select the next sample as described in \cref{subsec:next_sample}. In contrast, soft gating allows for the selection of multiple experts. Similar to hard gating, the distance is computed, but rather than selecting a single expert, the distance is used to compute the weights of the experts: $w_a = \frac{1/d_a}{\sum_{k=1}^{p} 1/d_a}$. 
After filtering experts who have a too low contribution (below 10\%) and computing $\tilde{w}_j$, 
all remaining experts are used to compute the mean and covariance matrix (\cref{eq:post}), and the results are scaled according to the weights: 
\begin{align}
\boldsymbol{\mu}_{\text{soft}} &= \sum_{j \in \mathcal{S}} \tilde{w}_j \, \boldsymbol{\mu}_j \\
\boldsymbol{\sigma}_{\text{soft}}^2 &= \sum_{j \in \mathcal{S}} \tilde{w}_j \, (\boldsymbol{\sigma}_j^2 + \boldsymbol{\mu}_j^2) - \boldsymbol{\mu}_{\text{soft}}^2
\end{align}

with $\mathcal{S} = \{ j \mid w_j > 0.1 \}$. We selected 10\%, as experts with small weights exhibit low similarity to the new part and contribute little to the aggregated prediction, while potentially introducing additional noise in the estimation.
Our approach provides an option to specify the number of iterations for which the MOE is used ($i_{moe}$). By default, we use soft gating during this process. At the end of each iteration, all experts used for prediction are retrained with a subset of data from the result file (organized by part) in addition to the results from the current simulation. 
After the specified number of iterations $i_{moe}$ is exceeded, a new expert for the new part is created, and the encoder from \cref{subsec:embedding} is retrained again. Consequently, the model for the new part becomes one of the experts in the MOE. For the next iterations, hard gating is employed, resulting in the explicit use of only a single model, namely the GPLVM model for this part. This is motivated by the fact, that Bayesian optimization typically requires only a few evaluations to yield good prediction.
In the future, we plan to reuse the MOE in subsequent iterations and enhance its training capabilities.

\subsection{Approximate Simulation via Early Termination}
\label{subsec:approx_computing}
As described in \cref{sec:introduction}, numerical simulations can be very expensive. In addition to using a model to guide the selection of the most promising input configurations through the prediction of the results, another way to reduce the computational cost is to terminate a simulation early if intermediate results indicate poor performance. For this, our approach allows us  to perform an intermediate check to examine the values of the target parameters. Using a specified percentage value and allowed ranges for the target parameters, a dedicated thread is launched once the OFsolve simulation progress reaches this value. This thread parses the results using OpenForm and signals the process handling the simulation to stop in case the target parameters exceed the allowed values.

While stopping the simulation at an early point risks discarding good results (i.e., the target parameters could still converge to good values), our use case allows for some exceptions. For instance, once the $L7$ parameter, which indicates cracks, attains a value different from zero, it is certain that it will not transition again below zero, as cracks can simply not be undone. For this purpose, we recommend executing the approximated computing check at values above or equal to 90\% with limits only on critical target parameters (i.e.,  $L7$, $L1$, and $L6$). 
In the future, we plan to further introduce approximated computing in our approach by modifying the precision or solving method of OFsolve.

For the experiments presented later in \cref{sec:evaluation}, we perform a single intermediate check at 90\%. This is motivated by the observation that this threshold is typically reached quickly, whereas the convergence toward the final solution (last 10\%) occurs more slowly. 
For the limits, the parameters $l_6$ and $l_7$ are set to the more restrictive values of $0.1$\% and $0.01$\% for the first use case (\cref{subsec:roof}), to moderately restrictive values of $0.5$\% and $0.01$\% for the second use case (\cref{subsec:seat}), and to the least restrictive values of $1$\% and $1$\% for the last use case (\cref{subsec:beam}). In all cases, $l_1$ is set to $15$\%. 
The workflow checks these limits in a separate thread once 90\% of the optimization process has been reached and terminates the current simulation if the conditions are satisfied.

%% file: content/05_evaluation.tex
\section{Evaluation}
\label{sec:evaluation}
In this section, we evaluate our approach using real-world designs from the industry. 
To examine our approach, we use a setup consisting of three machines: two local PCs and a single GPU node. The local machines are used by the experts to execute the simulations, while our AI workflow runs on the node. 
$S1$ is a local machine hosting a single Intel processor equipped with 6 cores on 1 socket  running at a base frequency of 1.6\,GHz with 2 threads per core (hyperthreading enabled).  In contrast, $S2$ features a single Intel processor with 4 cores at 2.8\,GHz on a single socket, each with 2 threads per core (hyperthreading enabled).  
The GPU node features two AMD EPYC 9255 processors, each with 24 cores per socket, 1 thread per core, and running at  1.5\,GHz (boost up to 4.3\,GHz). The machine also hosts 3 NVIDIA H200 NVL GPUs, each with 140\,GB of memory. For our experiments, we use just a single GPU. Note that our AI-Workflow  automatically detects the availability of GPUs and switches to use them instead of CPUs. We examined our approach based on three sheet metal forming parts from the industry.

\subsection{Use Case 1: Improving the Selection of Input Samples}
\label{subsec:roof}
\begin{figure}[bhp]
    \centering
        \includegraphics[width=.45\linewidth]{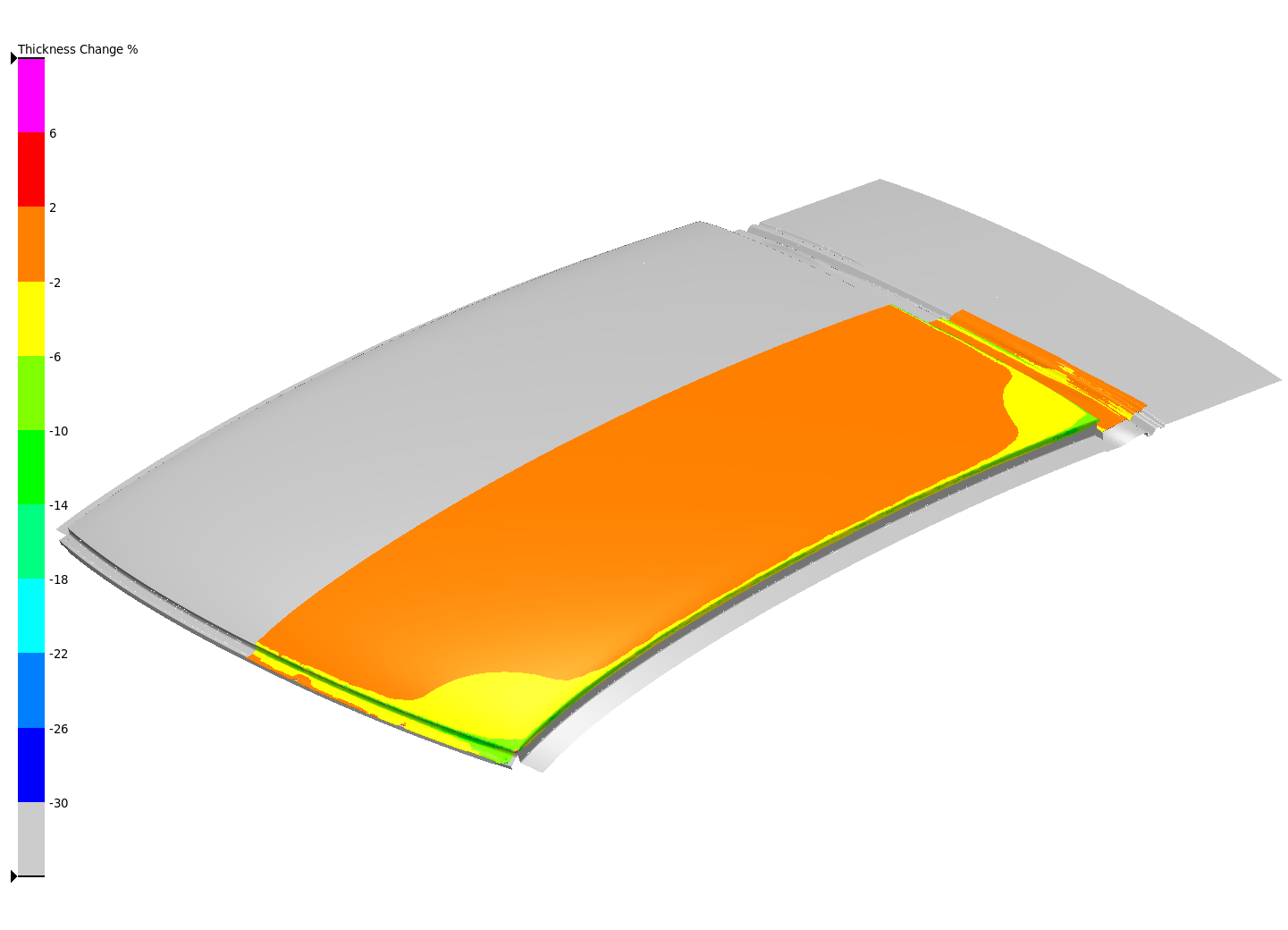}
    \includegraphics[width=0.45\linewidth]{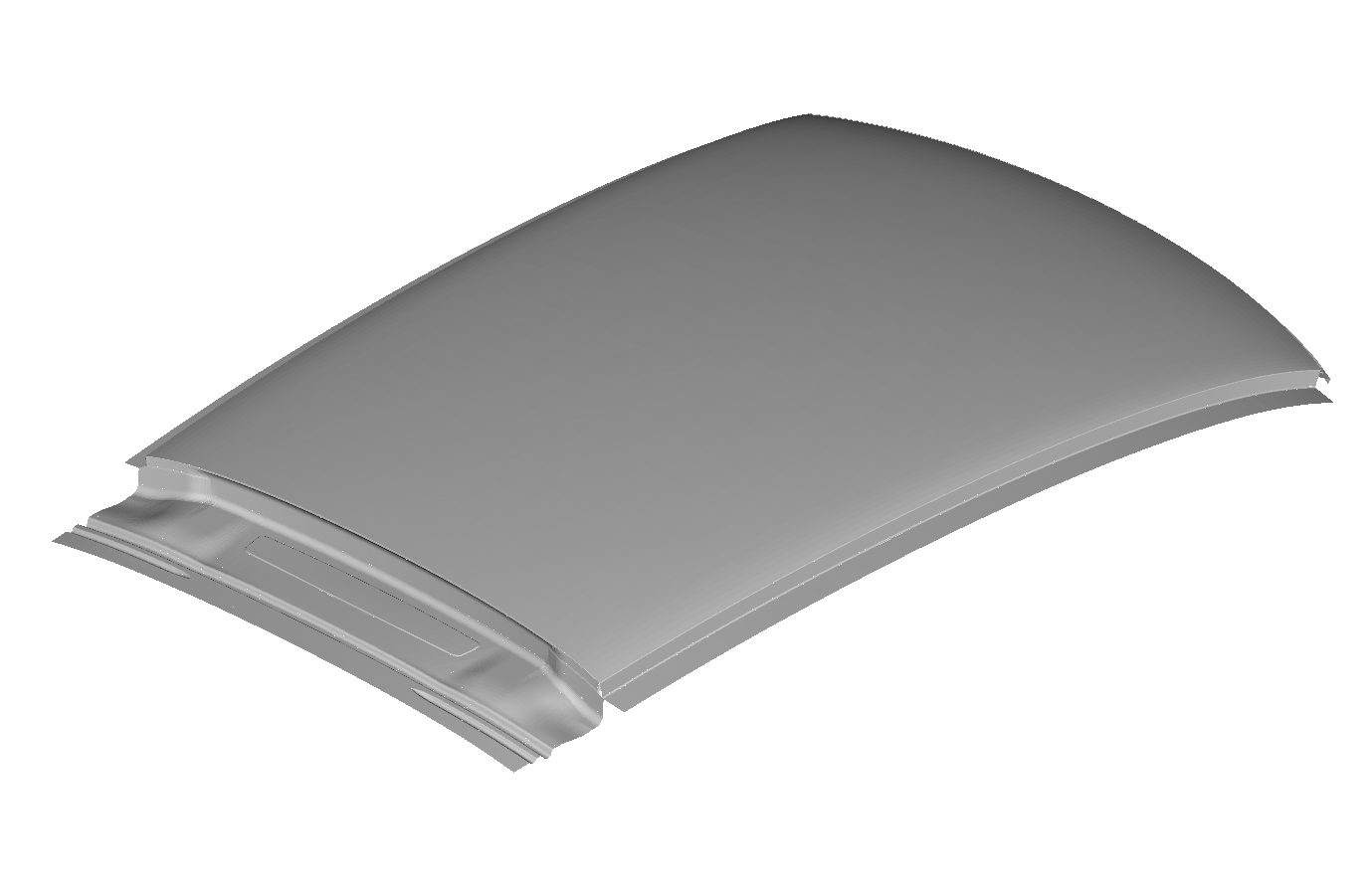}
    \caption{Roof part used for simulation.}
    \label{fig:roof}
\end{figure}

%
%
As a first example, we use a roof part for a car as illustrated \cref{fig:roof}. 
The geometry of the A-pillar is the most complex, hosting 723603 points. The geometric encoder reduces the number of points for all parts to $k_{emb}=5000$ as the \emph{down\_sample} mode is used (see \cref{subsec:embedding}).
%
%
We executed three simulations, each with a single sample per iteration (parallel samples, $ps=1$), varying the number of iterations, $i_{moe}$, until the MOE was reached (see \cref{subsec:gating}). In $sim_{1}$, the MOE is discarded in the 3rd iteration, in the 1st iteration in $sim_{2}$, and in the 5th iteration in $sim_{3}$. During the simulations with MOE and until $i_{moe}$, the following experts from the MOE were used using soft gating:
10.4\% B-pillar, 13.8\% wheelhouse adapter, 14.1\% cross member, 17.8\% front fender, 18.7\% rear hatch exterior, 22.7\% fuel cap housing. The remaining experts from the MOE were excluded, as the similarity (Euclidean distance) between their embedding and the current part's embedding was too low.
To compare the simulations from the AI workflow to those executed by humans, an unskilled user and an expert  conducted the simulation manually on system $S1$. While the AI workflow is completely automated, the user and expert had to manually adjust the configuration files to specify the input configurations. 

\begin{figure}[hbtp]
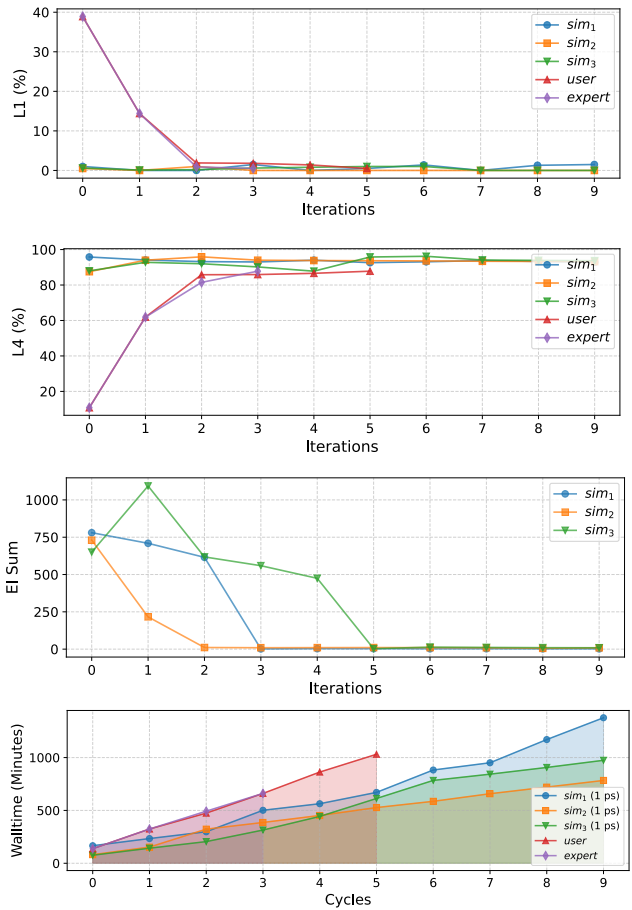

    \centering
    \includesvg[width=.95\linewidth]{figures/sim/MOE/DACH-VWS/L1_vs_iterations.svg}
    \includesvg[width=.95\linewidth]{figures/sim/MOE/DACH-VWS/L4_vs_iterations.svg}
    \includesvg[width=.95\linewidth]{figures/sim/MOE/DACH-VWS/EISum_vs_iterations.svg}
    \includesvg[width=.95\linewidth]{figures/sim/MOE/DACH-VWS/cumulative_cycles_walltime_vs_iterations.svg}
    \caption{Roof simulations with the AI-workflow using MOE (labeled $sim$) versus experts. The expert and unskilled user stop after 7 and 4 iterations, respectively, assuming that there is no further improvement.}
    \label{fig:dach_results}
\end{figure}

In \cref{fig:dach_results}, the results are presented, illustrating, from top to bottom, the two out of the seven target parameters (L1 and L4), the sum of the expected improvement over the outputs, and the walltime. 
While low values are required for L1, high ones are desired for L4 (the safe region). As observed, both experts and users require several iterations to find suitable input configurations. In contrast, all simulations attain improved values during the first iterations. Interestingly, the expert and unskilled user stop after 7 and 4  iteration, respectively, assuming that there is
no further improvement, though all simulations found higher ones. Interestingly, all simulations with the AI workflow already yielded good input configurations at the first iteration (iteration 0). If the simulations with the AI workflow had stopped at this iteration, the speedup to the user (a total of 1187.96 minutes at iteration 6) would be 7.1x, 14.5x, and 15.6x, for $sim_1$, $sim_2$, and $sim_3$, respectively. These values decrease to 3.9×, 8.1x, and 8.7x compared to the time when the expert (661.2 minutes at iteration 3) identified a good solution, although not as effective as the workflow’s solution.
As the plot for the sum of the expected improvement over the target parameters shows, the models assume that there is little enhancement after switching to the GPLVM model (after $i_{moe}$). With cycles denoting the number of iterations divided by parallel samples, the bottom figure illustrates the walltime. As expected, the simulations on the cluster are faster than those executed on system $S1$. Yet, speedup is relatively low, considering the number of cores used on the two systems.
%

\begin{figure}[tbp]
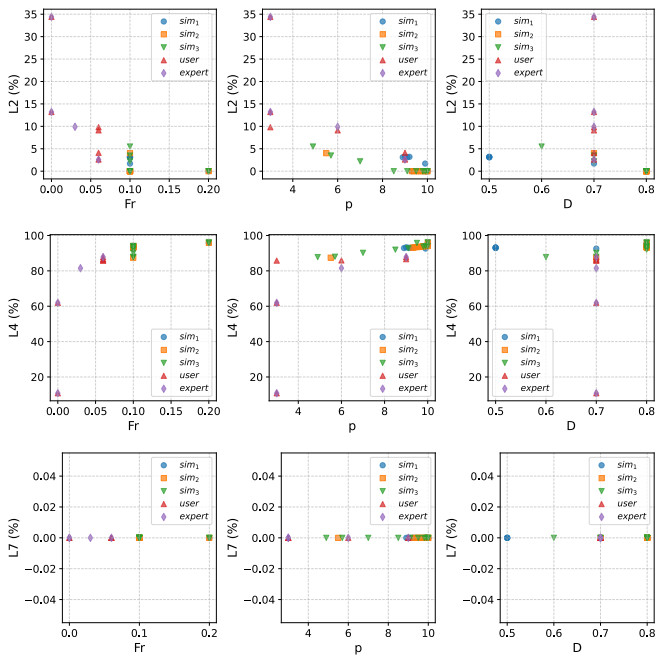

    \centering
    \includesvg[width=\linewidth]{figures/sim/MOE/DACH-VWS/inputs_vs_L2.svg}
    \includesvg[width=\linewidth]{figures/sim/MOE/DACH-VWS/inputs_vs_L4.svg}
    \includesvg[width=\linewidth]{figures/sim/MOE/DACH-VWS/inputs_vs_L7.svg}
    \caption{Input samples plotted against the target parameters L2, L4, and L7 for the roof part. For most target parameters, low values are desired, except for L4, which indicates the safe region and should thus attain high values.}
    \label{fig:dach_input_space}
\end{figure}

The select input parameters are illustrated in \cref{fig:dach_input_space} for the three selected input parameters: friction (Fr), pressure (p), and blank thickness (D). Furthermore, the figure shows the resulting target parameters L2, L4, and L7. 
As observed, Bayesian optimization enables the AI workflow to select promising input configurations in all cases, yielding good simulation results. Most importantly, L7 (cracks) is kept at a low value ($<1\%$) as targeted.

\subsection{Use Case 2: Energy Reduction Through Parallel Samples}
\label{subsec:seat}
%
%

\begin{figure}[bthp]
    \centering
    \includegraphics[width=0.48\linewidth]{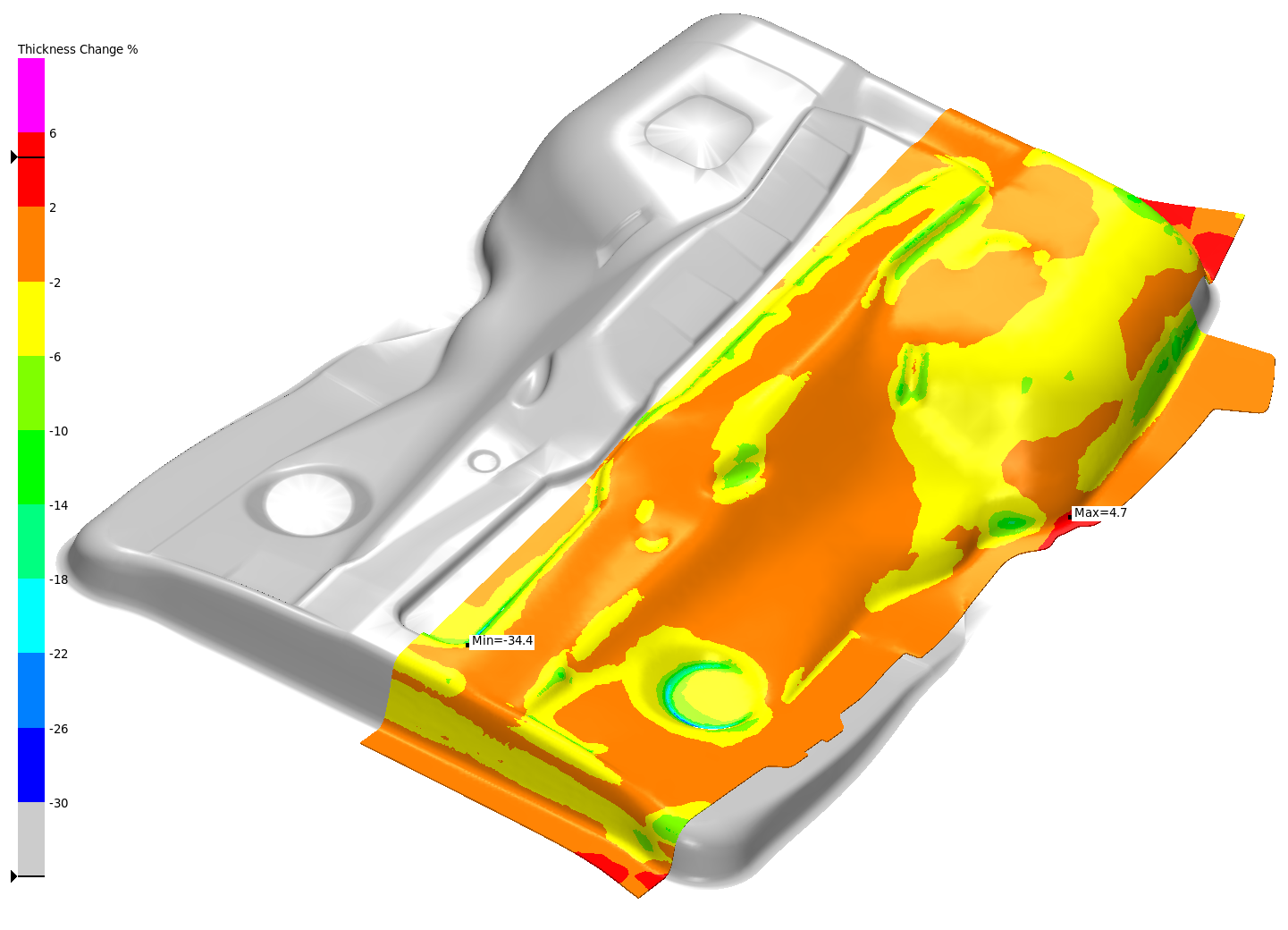}
    \includegraphics[width=0.48\linewidth]{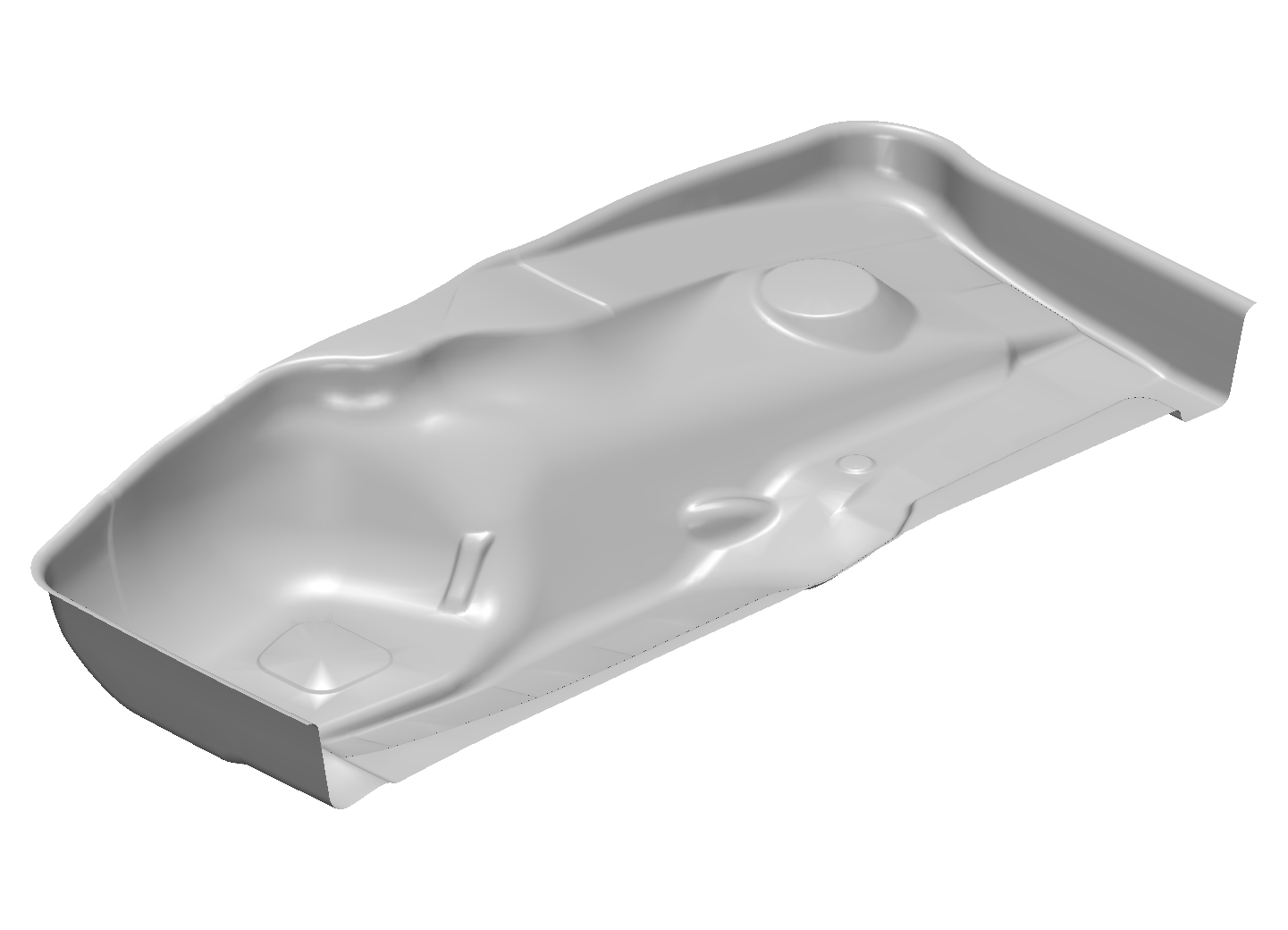}
    \caption{Seat shell used for simulation.}
    \label{fig:seatshell}
\end{figure}

In the next experiment, we focus on reducing the runtime of the optimization while considering the energy consumption. We conducted five simulations using the AI workflow for a seat shell, as illustrated in \cref{fig:seatshell}. For these simulations, we used the GPLVM models directly, building on scratch  (0 iterations) from the result file. Consequently, the first sample was randomly selected in each simulation. Furthermore, all use the  
\emph{combination} method 
%
%
for computing the space of input configurations  $\mathbf{X}_{\ast}$.
 The first setup ($sim_{1}$) is performed using a single sample per iteration, with 48 cores for 50 iterations. The next two setups use parallel sampling with 2 and 5 samples, respectively, and apply the \emph{crowding distance} selection strategy. In contrast, Setups 4 and 5 use the \emph{highest sum} selection strategy with 2 and 5 samples, respectively. The setups with two parallel samples execute 2 simulations in a cycle, each with 24 cores. With 5 samples, the number of cores drops to 9. At the end of each cycle, the model is retrained and the expected improvement $\mathbf{EI}$ is recalculated (see \cref{subsec:parallel_samples}). For comparison, an expert tries to optimize the part manually on system $S2$. 
%
\begin{figure}[bthp]
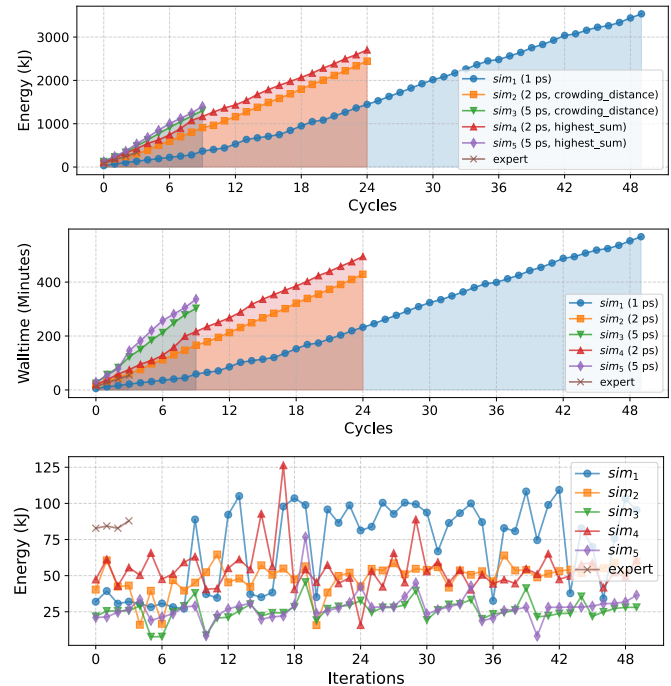

    \centering
    \includesvg[width=\linewidth]{figures/sim/MLVGP/SeatShell_1/cumulative_cycles_energy_vs_iterations.svg}
    \includesvg[width=\linewidth]{figures/sim/MLVGP/SeatShell_1/cumulative_cycles_walltime_vs_iterations.svg}
    \includesvg[width=\linewidth]{figures/sim/MLVGP/SeatShell_1/Energy_vs_iterations.svg}
    \caption{Simulation of a seat shell with 5 different configurations using a varying number of parallel samples (ps) and different selection strategies for these samples.}
    \label{fig:sheetShell_energy}
\end{figure}
As illustrated in the upper part of \cref{fig:sheetShell_energy}, increasing the number of parallel samples leads to lower overall energy consumption after several iterations occur during a single cycle. Compared to $sim_1$, the simulations with 2 parallel samples, namely $sim_2$ and $sim_4$, resulted in an energy reduction of 30.7\% and 23.5\%. The runs with 5 samples ($sim_3$ and $sim_5$) further reduced the energy consumption to 63.1\% and 60\%. 
Interestingly, setups using five samples with the \emph{crowding distance} method consume less energy than those using the \emph{highest sum} method. Typically, \emph{crowding distance} selects spatially diverse points, which can extend the simulation time. However, in this case, the setup using \emph{highest sum} with five parallel samples required more time, as the selected samples were thoroughly examined, whereas the early termination check (see \cref{subsec:approx_computing}) terminated unpromising samples earlier in the \emph{crowding distance} setup. 
\begin{figure}
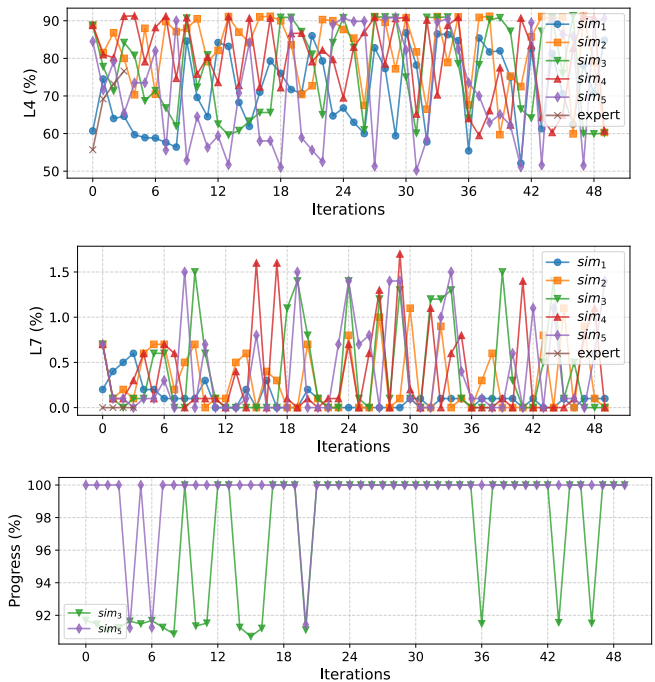

    \centering
    \includesvg[width=0.95\linewidth]{figures/sim/MLVGP/SeatShell_1/L4_vs_iterations.svg}
    \includesvg[width=0.95\linewidth]{figures/sim/MLVGP/SeatShell_1/L7_vs_iterations.svg}
    \includesvg[width=\linewidth]{figures/sim/MLVGP/SeatShell_1/process_vs_iterations.svg}
    \caption{Simulation of a seat shell using a different number of parallel samples and selection strategy. The top and middle plots show the values for L4  and L7  across different iterations. The bottom part shows the process time per iteration.}
    \label{fig:seatshell_sim}
\end{figure}
This effect is shown in the lower part of \cref{fig:seatshell_sim}, where most input configurations in $sim_{5}$ were completed (100\%), in contrast to $sim_{3}$, where the early termination check set at 90\% ended the simulation around this value.  \cref{fig:seatshell_sim} further shows that all simulations return good results (high values for L4 and values below 1\% for L7). 
In particular, already during the first iteration, the simulations with the AI workflow ($sim_1$ and $sim_5$) yielded 8.98\%, 59.43\%, 59.43\%, 59.43\%, and 51.71\% better L4  than the expert. With 4 iterations, the expert managed to catch up, surpassing $sim_1$ by 2.7\%, but still  15.93\%, 15.93\%, 19.06\%, and 10.31\% below simulations $sim_2$ to $sim_5$. 
As \cref{fig:seatshell_sim} shows, it  appears that the expert stopped too early before reaching high values for L4 (safe region). In contrast, aside from the simulation with a single parallel sample, all  simulations using the AI workflow achieved high values for L4 and low values (below 1\%) for L7 (cracks), as desired. 

\begin{figure}[tbp]
    \centering
    \includesvg[width=\linewidth]{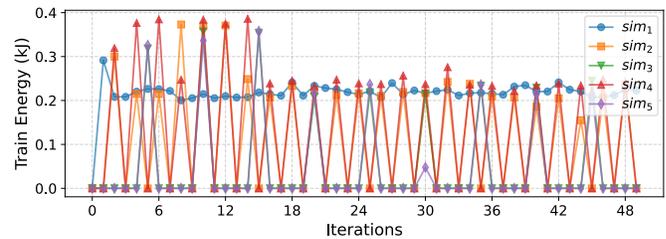}
    \caption{Energy consumption during training for the simulation of the seat shell. }
    \label{fig:sheetShell_train}
\end{figure}

When examining the energy consumption in \cref{fig:sheetShell_energy}, we also see that the energy consumed by the expert simulation is higher than that of the AI workflow. In particular, when comparing the energy consumption of the simulation from the expert for four iterations (not cycles) on the GPU node, the simulations from the workflow (from $sim_1$  to $sim_5$) used 60.4\%, 44.8\%, 70.8\%, 38.6\%, and 72.2\% less energy while already yielding good values for L4 and L7 during these iterations.
Note that the simulation of the expert attained values between 99 and 112 kJ on $S2$, while the runs on the GPU node consumed values between 82 and 88 kJ. Interestingly,
The training energy (\cref{fig:sheetShell_train}) remains negligibly low when compared to the energy consumed by the OpenForm simulation in \cref{fig:sheetShell_energy}. Note that since the simulations with several parallel samples only train the model after all samples have finished (i.e., at the end of the cycle), the training energy is zero except for the first sample. Moreover, for all simulations, the energy consumption is also zero at iteration/cycle zero, as there are not enough points to train the GPLVM model, and the sample is randomly selected. 

\begin{figure}[tbhp]
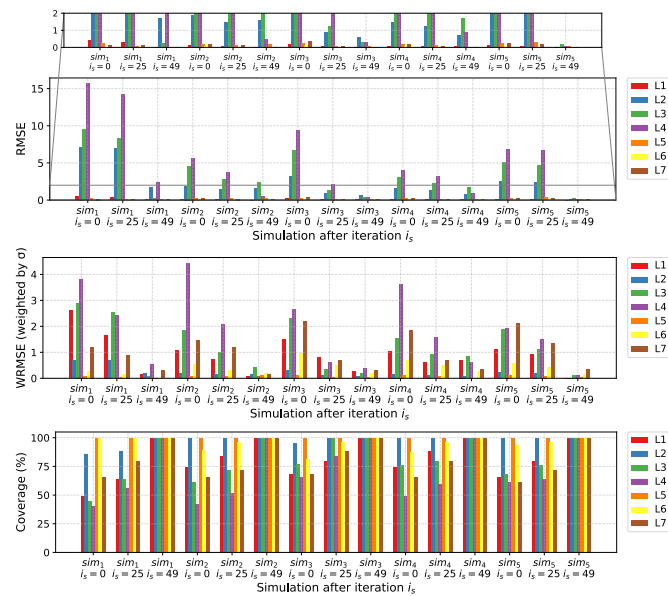

    \centering
    \includesvg[width=\linewidth]{figures/sim/MLVGP/SeatShell_1/RMSE_centered_zommed.svg}
    \includesvg[width=\linewidth]{
    figures/sim/MLVGP/SeatShell_1/WRMSE_centered.svg}
    \includesvg[width=\linewidth]{figures/sim/MLVGP/SeatShell_1/Coverage_centered.svg}
    \caption{RSME, weighted RSME (WRSME), and coverage of the simulations with the seat shell.}
    \label{fig:seatshell_rsme}
\end{figure}

To examine the accuracy of the models, we computed the RSME, weighted RSME (WRSME), and coverage of the simulation with the seat shell. For RSME and WRSME, the difference between the predicted mean from the GPLVM and the simulation results was computed for each target parameter. In contrast, the coverage shows the percentage of target points whose true values lie inside one standard deviation of the predicted mean from the GPLVM model. This is illustrated in \cref{fig:seatshell_rsme}. 
As models are improved after the end of the cycles, we recomputed these metrics at the middle (iteration 25) and the last (iteration 49) iterations. As observed, the error attains low values especially for $sim_5$. 

\begin{figure}[tbp]
    \centering
    \includesvg[width=\linewidth]{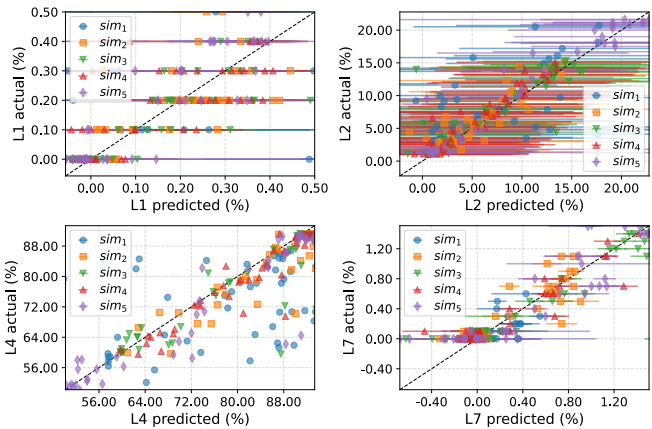}
    \caption{Consistency plots that show the predicted values of the target parameters from the GPLVM models versus the actual values from the OpenForm simulation for the different setups.}
    \label{fig:seatshell_consistency}
\end{figure}
\cref{fig:seatshell_consistency} shows the consistency plots for the target parameters L1, L2, L4, and L7 over all iterations. Aside from plotting the mean, the x values (predicted) show the standard deviation from this value. As the lower right corner of the L4 plot shows, $sim_1$ exhibits several poor predictions, which account for the lower values achieved compared to the other simulations. Considering that this simulation had the highest energy consumption and runtime, using parallel samples not only reduces these values but also improves exploration. 

\subsection{Use Case 3: Decreasing the Time to Solution}
\label{subsec:beam}
\begin{figure}[bp]
    \centering
    \includegraphics[width=0.48\linewidth]{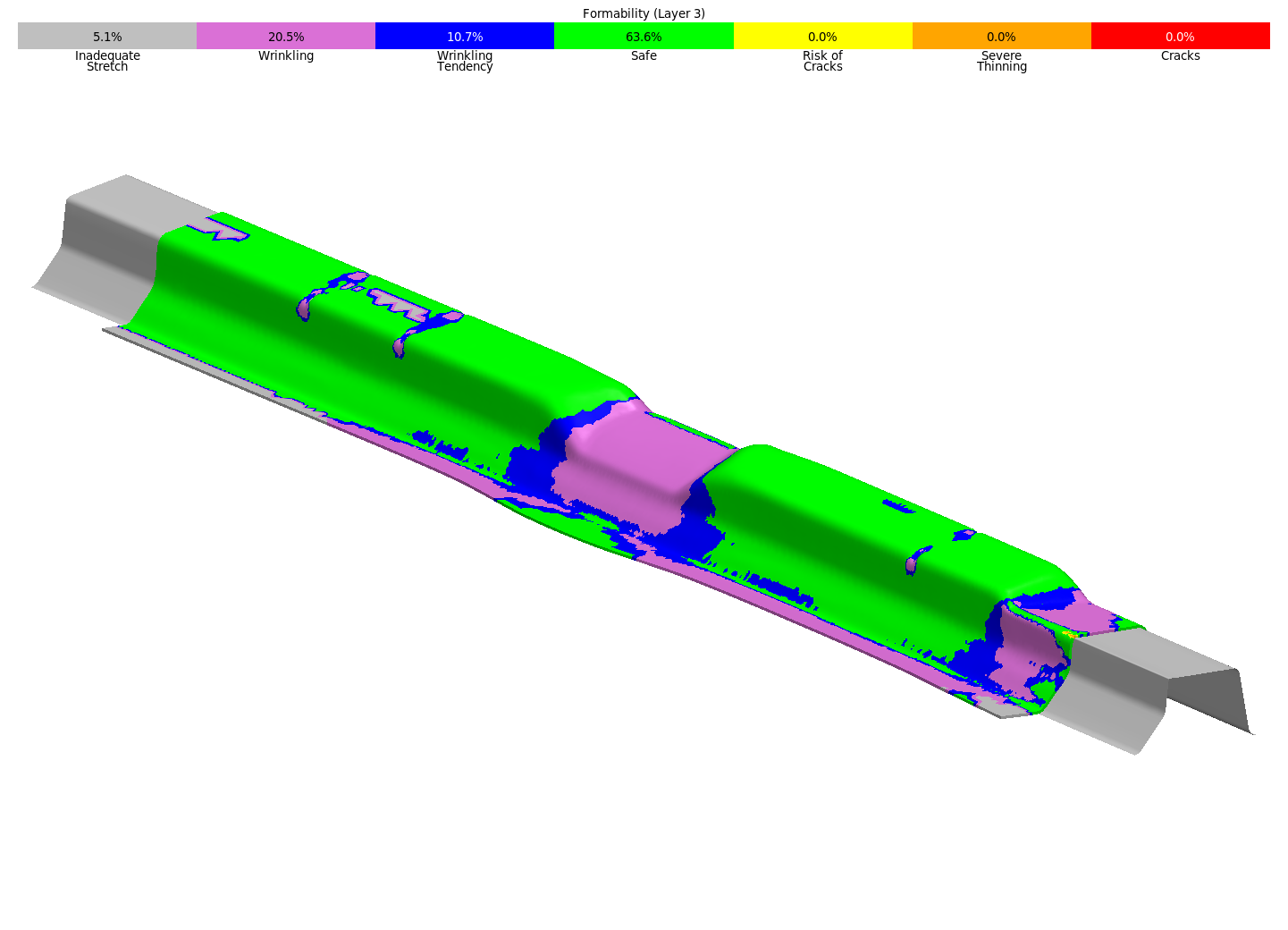}
    \includegraphics[width=0.48\linewidth]{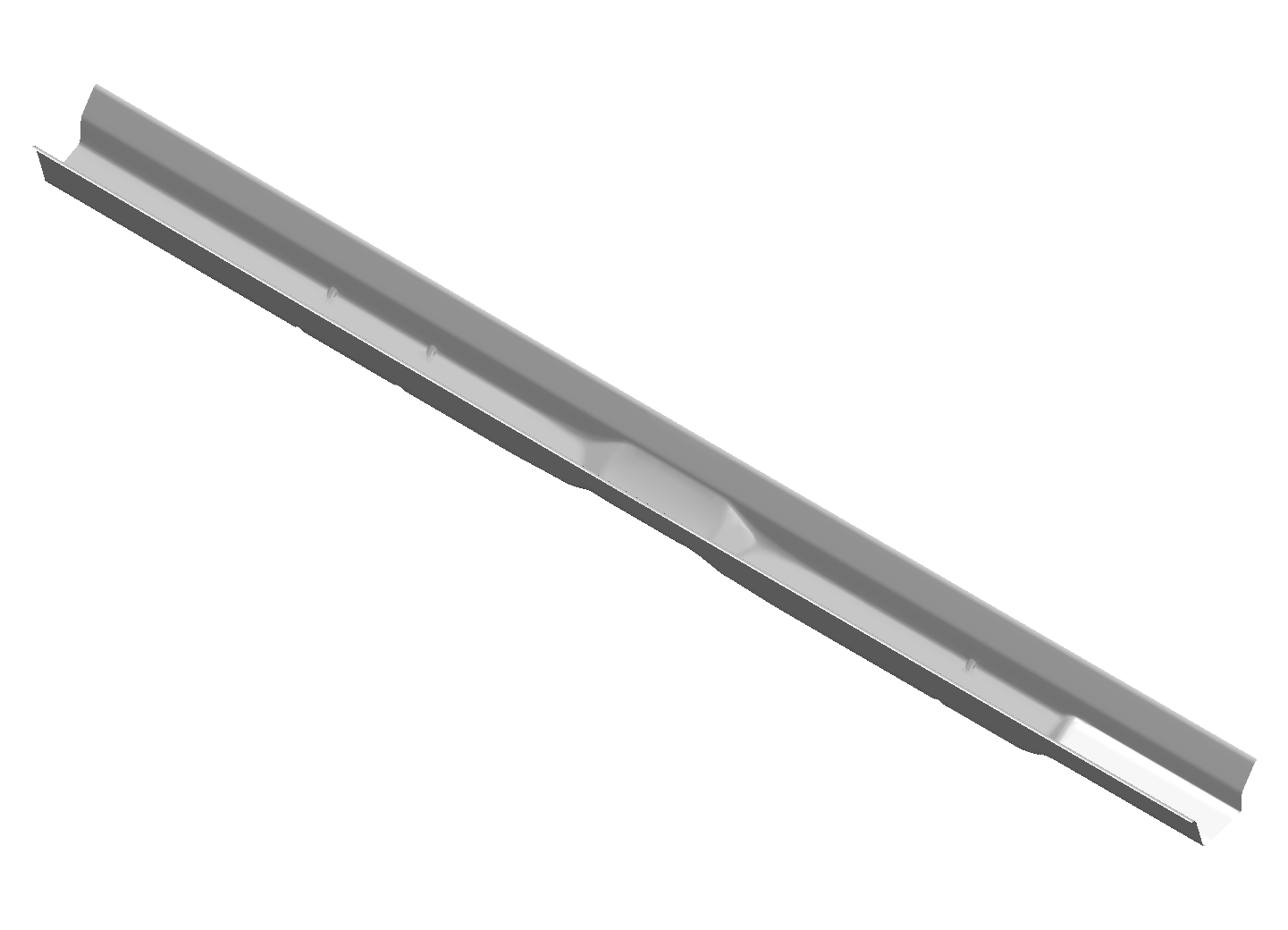}
    \caption{Longitudinal beam used for simulation.}
    \label{fig:Laengstraeger}
\end{figure}

\begin{figure}[bhtp]
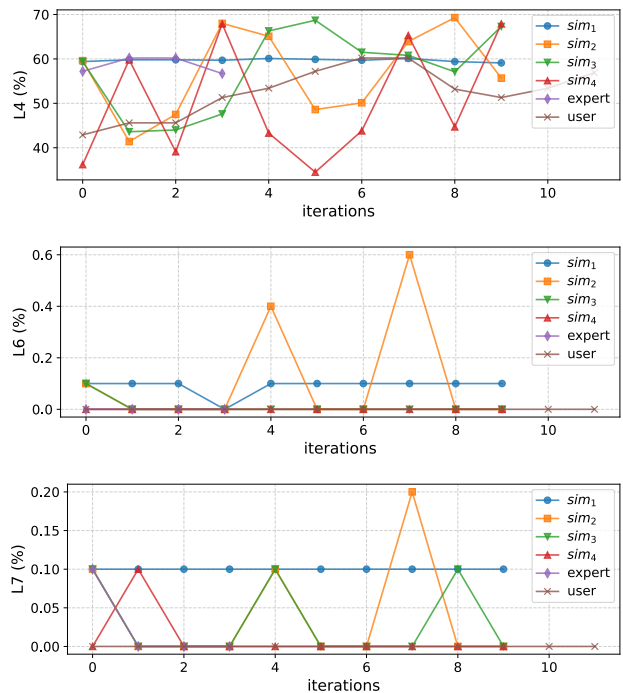

    \centering
    \includesvg[width=0.95\linewidth]{figures/sim/MOE/Laengstraeger_softLimits/L4_vs_iterations.svg}
    \includesvg[width=0.95\linewidth]{figures/sim/MOE/Laengstraeger_softLimits/L6_vs_iterations.svg}
    \includesvg[width=0.95\linewidth]{figures/sim/MOE/Laengstraeger_softLimits/L7_vs_iterations.svg}
    \caption{Simulation  results for the longitudinal beam illustrating three target parameters (L4, L6, and L7) over the iterations.}
    \label{fig:Laengstraeger_sim}
\end{figure}

\begin{figure}[tbp]
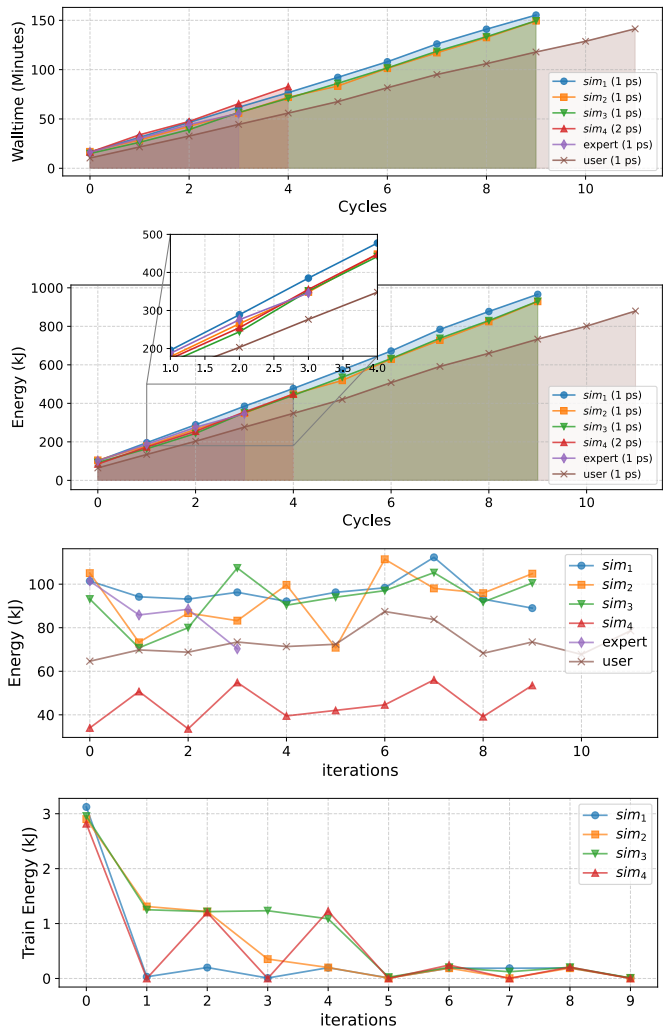

    \centering
    \includesvg[width=\linewidth]{figures/sim/MOE/Laengstraeger_softLimits/cumulative_cycles_walltime_vs_iterations.svg}
    \includesvg[width=\linewidth]{figures/sim/MOE/Laengstraeger_softLimits/cumulative_cycles_energy_vs_iterations_zoomed.svg}
    \includesvg[width=\linewidth]{figures/sim/MOE/Laengstraeger_softLimits/Energy_vs_iterations.svg}
    \includesvg[width=\linewidth]{figures/sim/MOE/Laengstraeger_softLimits/TrainEnergy_vs_iterations.svg}
    \caption{Walltime and energy consumption for the simulation of a longitudinal beam. The bottom part shows the train energy, which is not included in the simulation energy.}
    \label{fig:Laengstraeger_energy}
\end{figure}

%
%
For the third example, we examine the Longitudinal beam illustrated in \cref{fig:Laengstraeger}. 
Unlike the previous examples, we selected four input dimensions, including this time the discrete variable $Rp$, (yield strength). We performed four simulations using our AI workflow with the MOE setup, with a maximum of 10 iterations. In these simulations, we set $i_{moe}$ to 1, 3, 5, and 3 for simulations 1 to 4. Unlike the rest of the simulations, the last simulation ($sim_{4}$) uses two parallel samples with the \emph{highest\_sum} selection strategy.  Furthermore, all simulations use the  \emph{combination} method for computing the space of input configurations  $\mathbf{X}_{\ast}$. For comparison, an expert and an unskilled user manually optimize the same part on system $S2$ using OpenFoam directly. For a fairer comparison, we repeated these simulations on the GPU server with the specified input configurations using all available cores, similar to the AI workflow. 
\begin{figure}[tbhp]
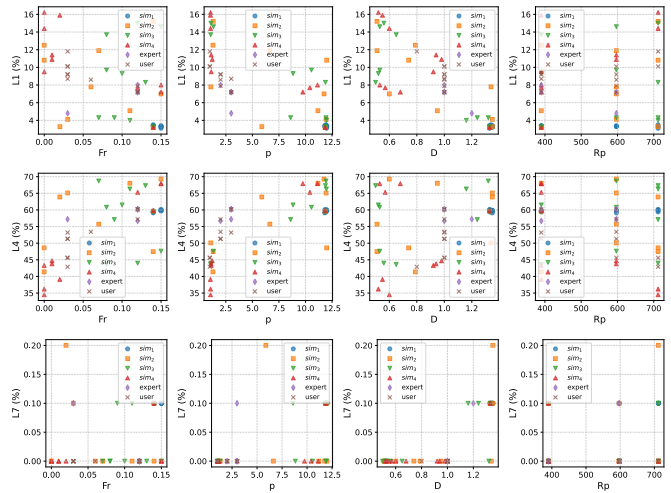

    \centering
    \includesvg[width=\linewidth]{figures/sim/MOE/Laengstraeger_softLimits/inputs_vs_L1.svg}
    \includesvg[width=\linewidth]{figures/sim/MOE/Laengstraeger_softLimits/inputs_vs_L4.svg}
    \includesvg[width=\linewidth]{figures/sim/MOE/Laengstraeger_softLimits/inputs_vs_L7.svg}
    \caption{Input samples plotted against the target parameters L1, L4, and L7 for the longitudinal beam from \cref{fig:Laengstraeger}. }
    \label{fig:Laengstraeger_input_space}
\end{figure}
\cref{fig:Laengstraeger_input_space} shows the samples selected for three out of seven target parameters (L1 to L7). As observed, and in contrast to \cref{fig:dach_input_space}, more diverse samples are selected due to their availability through the \emph{combination} method. 
Furthermore, when comparing $sim_2$ and $sim_4$, the latter option takes more risk (selects two samples before updating $\mathbf{EI}$) and explores regions that might yield suboptimal results. 
Interestingly, this approach yields good results quickly, as evident in iteration 3 of \cref{fig:Laengstraeger_sim}. In contrast, $sim_2$ reaches an even better result at iteration 8 that satisfies the constraints (L7 $<$ 1\%).
When considering the energy consumption shown in \cref{fig:sheetShell_energy}, $sim_4$ attains the lowest values, while yielding the fastest results. Although the results may not be optimal, the fact that two iterations were executed in parallel resulted in almost half the time and energy consumption compared to the other simulations. Additionally, in iterations 0 and 2, $sim_4$ stopped the simulations, as the results were outside the desired ranges (see \cref{subsec:approx_computing}), resulting in even further energy savings. 
For this experiment, the expert stopped too early again (\cref{fig:Laengstraeger_sim}), although the energy consumption is close to what the simulation during the AI workflow required. In contrast, the user attempted more iterations to improve the results, but with little success, as the L4 values were relatively low. In particular, the user managed to reach 60\% after 7 iterations, a value achieved by all simulations with the AI workflow by the fourth iteration at the latest. While all these simulations attained even higher values by the $7^{th}$ iteration, the user was unable to surpass this value, while still consuming a considerable amount of energy. Furthermore, compared to the high values of energy consumed during the simulations, the training of the model (MOE till $i_{moe}$ and GPLVM afterwards) at the end of each cycle required only negligible values, as the bottom part of \cref{fig:Laengstraeger_energy} shows.
Interestingly, the AI workflow with two parallel samples  ($sim_4$) not only resulted in low energy consumption but also identified good input configurations (e.g., iteration 3) with the shortest walltime. 

\subsection{Decreasing the Total Cost of Ownership}\label{subsec:tco}

To evaluate the overall impact of the AI workflow, it is beneficial to examine the Total Cost of Ownership (TCO) associated with fully simulating a specific part.
This includes all costs incurred during the process, including personnel costs and hardware operation costs.
In this section, we compare the TCO of simulating the three previously shown parts for the three user types: an unskilled user, an expert, and the AI model.

For the purposes of this paper, we propose the use case of employing a cloud computing solution to host both the simulation and inference of the AI model.
For the simulation, we referenced the Google Cloud pricing for their \texttt{n1-highcpu-8} machines, which are optimized for high CPU loads and equipped with 8 cores, at a rate of \$0.3651376 per hour.
As the AI inference benefits from the utilization of a GPU, we looked at the \texttt{a2-ultragpu-1g} nodes, which provide a single Nvidia A100 GPU at \$6.526873068 per hour.
The values provided here reference the Google Cloud pricing as of January 2026 at the Frankfurt (europe-west3) location.

The TCO can be broken down into a refinement phase ($r$) and a simulation phase ($s$), as shown in \cref{eq:tco},
\begin{equation}
    \text{TCO} = \sum_i t_rc_r + t_sc_s
    \label{eq:tco}
\end{equation}
Where $i$ refers to the number of iterations needed to determine the correct parameter set, $t_\square$ refers to the time taken in the refinement or simulation phase, and $c_\square$ to the respective costs.

%
%
%

%
%

Since the proposed AI workflow replaces the human-led refinement phase of selecting a new set of parameters, we can utilize the GPU node costs and inference times for the AI user type.
For the human user types, we assumed an hourly wage of \$25 for the user and \$70 for the expert, stemming from discussions with industry insiders.
The times of the refinement phases and the number of iterations were taken from the experiments discussed above, where we always chose the best-performing AI model in each case.
We assumed that the phase times remained constant across different iterations.
Note that the simulation phases are the same for each user type and only differ on a per-part basis, as they are always performed by \emph{OpenForm} on the CPU nodes.

\begin{figure}[bhtp]
    \centering
    \def\mathdefault#1{#1}
    \adjustbox{max width=0.47\textwidth}    {\input{figures/sim/TCO/tco_50iters.pgf}}
    \caption{Total Cost of Ownership Calculations for the three tested parts, based on Google Cloud pricing as of January 2026. Lower is better.}
    \label{fig:TCO}
\end{figure}

A summary of the results is presented in \cref{fig:TCO}.
It is of note that despite the higher salary of the expert, the TCO for that user type is lower than that of the unskilled user, owing to the lower number of iterations needed to reach an acceptable parameter set and the lower time of refinement - the expert needed an average of 12 minutes, and the unskilled user of 30 minutes. Secondly, the AI workflow achieves much lower costs for the roof and beam parts and comparable costs to the expert for the seat part.
The generally lower costs are due to the much lower hourly costs of refinement when compared to those of human users.
For the seat part, the AI workflow stopped after a very high number of iterations (50 compared to 4 expert iterations), but it could have been stopped earlier and reached comparable results to the expert. We opted to include all iterations for the AI models to demonstrate that even with a significantly larger number of iterations, the TCO of the AI workflow remains lower compared to human users. 
Through the use of the AI-supported workflow, the seat shell (4 iterations) achieves a cost saving of about 92\%  compared to the expert, the longitudinal beam about 79\%, and the roof component around 62\%. 
In comparison to the inexperienced user, the savings for the seat shell (4 iterations) are approximately 94\%, for the longitudinal beam around 92\%, and for the roof component roughly 74\%. Even when the seat shell is executed with 50 iterations, the TCO costs remain only slightly higher than those of the experts (around 5\%) but are still about 21\% lower than for the inexperienced users.

In conclusion, the TCO for the AI workflow should be much lower than a human-led parameter refinement in most cases. Yet, this does \emph{not} imply that humans are no longer needed in this design process, as \cref{fig:new_workflow} shows, the intention  of our workflow is to reduce the involvement of the users. This is underscored with the validation step (orange box), which requires the expert to check that the results are actually feasible in practice. 

%% file: figures/sim/TCO/tco_50iters.pgf
\begingroup%
\makeatletter%
\begin{pgfpicture}%
\pgfpathrectangle{\pgfpointorigin}{\pgfqpoint{6.400000in}{4.800000in}}%
\pgfusepath{use as bounding box, clip}%
\begin{pgfscope}%
\pgfsetbuttcap%
\pgfsetmiterjoin%
\definecolor{currentfill}{rgb}{1.000000,1.000000,1.000000}%
\pgfsetfillcolor{currentfill}%
\pgfsetlinewidth{0.000000pt}%
\definecolor{currentstroke}{rgb}{1.000000,1.000000,1.000000}%
\pgfsetstrokecolor{currentstroke}%
\pgfsetdash{}{0pt}%
\pgfpathmoveto{\pgfqpoint{0.000000in}{0.000000in}}%
\pgfpathlineto{\pgfqpoint{6.400000in}{0.000000in}}%
\pgfpathlineto{\pgfqpoint{6.400000in}{4.800000in}}%
\pgfpathlineto{\pgfqpoint{0.000000in}{4.800000in}}%
\pgfpathlineto{\pgfqpoint{0.000000in}{0.000000in}}%
\pgfpathclose%
\pgfusepath{fill}%
\end{pgfscope}%
\begin{pgfscope}%
\pgfsetbuttcap%
\pgfsetmiterjoin%
\definecolor{currentfill}{rgb}{1.000000,1.000000,1.000000}%
\pgfsetfillcolor{currentfill}%
\pgfsetlinewidth{0.000000pt}%
\definecolor{currentstroke}{rgb}{0.000000,0.000000,0.000000}%
\pgfsetstrokecolor{currentstroke}%
\pgfsetstrokeopacity{0.000000}%
\pgfsetdash{}{0pt}%
\pgfpathmoveto{\pgfqpoint{0.836107in}{0.986070in}}%
\pgfpathlineto{\pgfqpoint{6.250000in}{0.986070in}}%
\pgfpathlineto{\pgfqpoint{6.250000in}{4.650000in}}%
\pgfpathlineto{\pgfqpoint{0.836107in}{4.650000in}}%
\pgfpathlineto{\pgfqpoint{0.836107in}{0.986070in}}%
\pgfpathclose%
\pgfusepath{fill}%
\end{pgfscope}%
\begin{pgfscope}%
\pgfpathrectangle{\pgfqpoint{0.836107in}{0.986070in}}{\pgfqpoint{5.413893in}{3.663930in}}%
\pgfusepath{clip}%
\pgfsetbuttcap%
\pgfsetmiterjoin%
\definecolor{currentfill}{rgb}{0.121569,0.466667,0.705882}%
\pgfsetfillcolor{currentfill}%
\pgfsetlinewidth{0.000000pt}%
\definecolor{currentstroke}{rgb}{0.000000,0.000000,0.000000}%
\pgfsetstrokecolor{currentstroke}%
\pgfsetstrokeopacity{0.000000}%
\pgfsetdash{}{0pt}%
\pgfpathmoveto{\pgfqpoint{1.287265in}{0.986070in}}%
\pgfpathlineto{\pgfqpoint{1.588037in}{0.986070in}}%
\pgfpathlineto{\pgfqpoint{1.588037in}{2.834298in}}%
\pgfpathlineto{\pgfqpoint{1.287265in}{2.834298in}}%
\pgfpathlineto{\pgfqpoint{1.287265in}{0.986070in}}%
\pgfpathclose%
\pgfusepath{fill}%
\end{pgfscope}%
\begin{pgfscope}%
\pgfpathrectangle{\pgfqpoint{0.836107in}{0.986070in}}{\pgfqpoint{5.413893in}{3.663930in}}%
\pgfusepath{clip}%
\pgfsetbuttcap%
\pgfsetmiterjoin%
\definecolor{currentfill}{rgb}{0.121569,0.466667,0.705882}%
\pgfsetfillcolor{currentfill}%
\pgfsetlinewidth{0.000000pt}%
\definecolor{currentstroke}{rgb}{0.000000,0.000000,0.000000}%
\pgfsetstrokecolor{currentstroke}%
\pgfsetstrokeopacity{0.000000}%
\pgfsetdash{}{0pt}%
\pgfpathmoveto{\pgfqpoint{3.091896in}{0.986070in}}%
\pgfpathlineto{\pgfqpoint{3.392668in}{0.986070in}}%
\pgfpathlineto{\pgfqpoint{3.392668in}{2.730914in}}%
\pgfpathlineto{\pgfqpoint{3.091896in}{2.730914in}}%
\pgfpathlineto{\pgfqpoint{3.091896in}{0.986070in}}%
\pgfpathclose%
\pgfusepath{fill}%
\end{pgfscope}%
\begin{pgfscope}%
\pgfpathrectangle{\pgfqpoint{0.836107in}{0.986070in}}{\pgfqpoint{5.413893in}{3.663930in}}%
\pgfusepath{clip}%
\pgfsetbuttcap%
\pgfsetmiterjoin%
\definecolor{currentfill}{rgb}{0.121569,0.466667,0.705882}%
\pgfsetfillcolor{currentfill}%
\pgfsetlinewidth{0.000000pt}%
\definecolor{currentstroke}{rgb}{0.000000,0.000000,0.000000}%
\pgfsetstrokecolor{currentstroke}%
\pgfsetstrokeopacity{0.000000}%
\pgfsetdash{}{0pt}%
\pgfpathmoveto{\pgfqpoint{4.896527in}{0.986070in}}%
\pgfpathlineto{\pgfqpoint{5.197299in}{0.986070in}}%
\pgfpathlineto{\pgfqpoint{5.197299in}{4.475527in}}%
\pgfpathlineto{\pgfqpoint{4.896527in}{4.475527in}}%
\pgfpathlineto{\pgfqpoint{4.896527in}{0.986070in}}%
\pgfpathclose%
\pgfusepath{fill}%
\end{pgfscope}%
\begin{pgfscope}%
\pgfpathrectangle{\pgfqpoint{0.836107in}{0.986070in}}{\pgfqpoint{5.413893in}{3.663930in}}%
\pgfusepath{clip}%
\pgfsetbuttcap%
\pgfsetmiterjoin%
\definecolor{currentfill}{rgb}{1.000000,0.498039,0.054902}%
\pgfsetfillcolor{currentfill}%
\pgfsetlinewidth{0.000000pt}%
\definecolor{currentstroke}{rgb}{0.000000,0.000000,0.000000}%
\pgfsetstrokecolor{currentstroke}%
\pgfsetstrokeopacity{0.000000}%
\pgfsetdash{}{0pt}%
\pgfpathmoveto{\pgfqpoint{1.588037in}{0.986070in}}%
\pgfpathlineto{\pgfqpoint{1.888808in}{0.986070in}}%
\pgfpathlineto{\pgfqpoint{1.888808in}{2.267761in}}%
\pgfpathlineto{\pgfqpoint{1.588037in}{2.267761in}}%
\pgfpathlineto{\pgfqpoint{1.588037in}{0.986070in}}%
\pgfpathclose%
\pgfusepath{fill}%
\end{pgfscope}%
\begin{pgfscope}%
\pgfpathrectangle{\pgfqpoint{0.836107in}{0.986070in}}{\pgfqpoint{5.413893in}{3.663930in}}%
\pgfusepath{clip}%
\pgfsetbuttcap%
\pgfsetmiterjoin%
\definecolor{currentfill}{rgb}{1.000000,0.498039,0.054902}%
\pgfsetfillcolor{currentfill}%
\pgfsetlinewidth{0.000000pt}%
\definecolor{currentstroke}{rgb}{0.000000,0.000000,0.000000}%
\pgfsetstrokecolor{currentstroke}%
\pgfsetstrokeopacity{0.000000}%
\pgfsetdash{}{0pt}%
\pgfpathmoveto{\pgfqpoint{3.392668in}{0.986070in}}%
\pgfpathlineto{\pgfqpoint{3.693439in}{0.986070in}}%
\pgfpathlineto{\pgfqpoint{3.693439in}{2.287837in}}%
\pgfpathlineto{\pgfqpoint{3.392668in}{2.287837in}}%
\pgfpathlineto{\pgfqpoint{3.392668in}{0.986070in}}%
\pgfpathclose%
\pgfusepath{fill}%
\end{pgfscope}%
\begin{pgfscope}%
\pgfpathrectangle{\pgfqpoint{0.836107in}{0.986070in}}{\pgfqpoint{5.413893in}{3.663930in}}%
\pgfusepath{clip}%
\pgfsetbuttcap%
\pgfsetmiterjoin%
\definecolor{currentfill}{rgb}{1.000000,0.498039,0.054902}%
\pgfsetfillcolor{currentfill}%
\pgfsetlinewidth{0.000000pt}%
\definecolor{currentstroke}{rgb}{0.000000,0.000000,0.000000}%
\pgfsetstrokecolor{currentstroke}%
\pgfsetstrokeopacity{0.000000}%
\pgfsetdash{}{0pt}%
\pgfpathmoveto{\pgfqpoint{5.197299in}{0.986070in}}%
\pgfpathlineto{\pgfqpoint{5.498070in}{0.986070in}}%
\pgfpathlineto{\pgfqpoint{5.498070in}{2.288068in}}%
\pgfpathlineto{\pgfqpoint{5.197299in}{2.288068in}}%
\pgfpathlineto{\pgfqpoint{5.197299in}{0.986070in}}%
\pgfpathclose%
\pgfusepath{fill}%
\end{pgfscope}%
\begin{pgfscope}%
\pgfpathrectangle{\pgfqpoint{0.836107in}{0.986070in}}{\pgfqpoint{5.413893in}{3.663930in}}%
\pgfusepath{clip}%
\pgfsetbuttcap%
\pgfsetmiterjoin%
\definecolor{currentfill}{rgb}{0.172549,0.627451,0.172549}%
\pgfsetfillcolor{currentfill}%
\pgfsetlinewidth{0.000000pt}%
\definecolor{currentstroke}{rgb}{0.000000,0.000000,0.000000}%
\pgfsetstrokecolor{currentstroke}%
\pgfsetstrokeopacity{0.000000}%
\pgfsetdash{}{0pt}%
\pgfpathmoveto{\pgfqpoint{1.888808in}{0.986070in}}%
\pgfpathlineto{\pgfqpoint{2.189580in}{0.986070in}}%
\pgfpathlineto{\pgfqpoint{2.189580in}{1.471146in}}%
\pgfpathlineto{\pgfqpoint{1.888808in}{1.471146in}}%
\pgfpathlineto{\pgfqpoint{1.888808in}{0.986070in}}%
\pgfpathclose%
\pgfusepath{fill}%
\end{pgfscope}%
\begin{pgfscope}%
\pgfpathrectangle{\pgfqpoint{0.836107in}{0.986070in}}{\pgfqpoint{5.413893in}{3.663930in}}%
\pgfusepath{clip}%
\pgfsetbuttcap%
\pgfsetmiterjoin%
\definecolor{currentfill}{rgb}{0.172549,0.627451,0.172549}%
\pgfsetfillcolor{currentfill}%
\pgfsetlinewidth{0.000000pt}%
\definecolor{currentstroke}{rgb}{0.000000,0.000000,0.000000}%
\pgfsetstrokecolor{currentstroke}%
\pgfsetstrokeopacity{0.000000}%
\pgfsetdash{}{0pt}%
\pgfpathmoveto{\pgfqpoint{3.693439in}{0.986070in}}%
\pgfpathlineto{\pgfqpoint{3.994211in}{0.986070in}}%
\pgfpathlineto{\pgfqpoint{3.994211in}{2.358222in}}%
\pgfpathlineto{\pgfqpoint{3.693439in}{2.358222in}}%
\pgfpathlineto{\pgfqpoint{3.693439in}{0.986070in}}%
\pgfpathclose%
\pgfusepath{fill}%
\end{pgfscope}%
\begin{pgfscope}%
\pgfpathrectangle{\pgfqpoint{0.836107in}{0.986070in}}{\pgfqpoint{5.413893in}{3.663930in}}%
\pgfusepath{clip}%
\pgfsetbuttcap%
\pgfsetmiterjoin%
\definecolor{currentfill}{rgb}{0.172549,0.627451,0.172549}%
\pgfsetfillcolor{currentfill}%
\pgfsetlinewidth{0.000000pt}%
\definecolor{currentstroke}{rgb}{0.000000,0.000000,0.000000}%
\pgfsetstrokecolor{currentstroke}%
\pgfsetstrokeopacity{0.000000}%
\pgfsetdash{}{0pt}%
\pgfpathmoveto{\pgfqpoint{5.498070in}{0.986070in}}%
\pgfpathlineto{\pgfqpoint{5.798842in}{0.986070in}}%
\pgfpathlineto{\pgfqpoint{5.798842in}{1.260454in}}%
\pgfpathlineto{\pgfqpoint{5.498070in}{1.260454in}}%
\pgfpathlineto{\pgfqpoint{5.498070in}{0.986070in}}%
\pgfpathclose%
\pgfusepath{fill}%
\end{pgfscope}%
\begin{pgfscope}%
\pgfsetbuttcap%
\pgfsetroundjoin%
\definecolor{currentfill}{rgb}{0.000000,0.000000,0.000000}%
\pgfsetfillcolor{currentfill}%
\pgfsetlinewidth{0.803000pt}%
\definecolor{currentstroke}{rgb}{0.000000,0.000000,0.000000}%
\pgfsetstrokecolor{currentstroke}%
\pgfsetdash{}{0pt}%
\pgfsys@defobject{currentmarker}{\pgfqpoint{0.000000in}{-0.048611in}}{\pgfqpoint{0.000000in}{0.000000in}}{%
\pgfpathmoveto{\pgfqpoint{0.000000in}{0.000000in}}%
\pgfpathlineto{\pgfqpoint{0.000000in}{-0.048611in}}%
\pgfusepath{stroke,fill}%
}%
\begin{pgfscope}%
\pgfsys@transformshift{1.738422in}{0.986070in}%
\pgfsys@useobject{currentmarker}{}%
\end{pgfscope}%
\end{pgfscope}%
\begin{pgfscope}%
\definecolor{textcolor}{rgb}{0.000000,0.000000,0.000000}%
\pgfsetstrokecolor{textcolor}%
\pgfsetfillcolor{textcolor}%
\pgftext[x=1.630509in, y=0.504100in, left, base,rotate=45.000000]{\color{textcolor}{\rmfamily\fontsize{14.400000}{17.280000}\selectfont\catcode`\^=\active\def^{\ifmmode\sp\else\^{}\fi}\catcode`\%=\active\def
\end{pgfscope}%
\begin{pgfscope}%
\pgfsetbuttcap%
\pgfsetroundjoin%
\definecolor{currentfill}{rgb}{0.000000,0.000000,0.000000}%
\pgfsetfillcolor{currentfill}%
\pgfsetlinewidth{0.803000pt}%
\definecolor{currentstroke}{rgb}{0.000000,0.000000,0.000000}%
\pgfsetstrokecolor{currentstroke}%
\pgfsetdash{}{0pt}%
\pgfsys@defobject{currentmarker}{\pgfqpoint{0.000000in}{-0.048611in}}{\pgfqpoint{0.000000in}{0.000000in}}{%
\pgfpathmoveto{\pgfqpoint{0.000000in}{0.000000in}}%
\pgfpathlineto{\pgfqpoint{0.000000in}{-0.048611in}}%
\pgfusepath{stroke,fill}%
}%
\begin{pgfscope}%
\pgfsys@transformshift{3.543053in}{0.986070in}%
\pgfsys@useobject{currentmarker}{}%
\end{pgfscope}%
\end{pgfscope}%
\begin{pgfscope}%
\definecolor{textcolor}{rgb}{0.000000,0.000000,0.000000}%
\pgfsetstrokecolor{textcolor}%
\pgfsetfillcolor{textcolor}%
\pgftext[x=3.447628in, y=0.529078in, left, base,rotate=45.000000]{\color{textcolor}{\rmfamily\fontsize{14.400000}{17.280000}\selectfont\catcode`\^=\active\def^{\ifmmode\sp\else\^{}\fi}\catcode`\%=\active\def
\end{pgfscope}%
\begin{pgfscope}%
\pgfsetbuttcap%
\pgfsetroundjoin%
\definecolor{currentfill}{rgb}{0.000000,0.000000,0.000000}%
\pgfsetfillcolor{currentfill}%
\pgfsetlinewidth{0.803000pt}%
\definecolor{currentstroke}{rgb}{0.000000,0.000000,0.000000}%
\pgfsetstrokecolor{currentstroke}%
\pgfsetdash{}{0pt}%
\pgfsys@defobject{currentmarker}{\pgfqpoint{0.000000in}{-0.048611in}}{\pgfqpoint{0.000000in}{0.000000in}}{%
\pgfpathmoveto{\pgfqpoint{0.000000in}{0.000000in}}%
\pgfpathlineto{\pgfqpoint{0.000000in}{-0.048611in}}%
\pgfusepath{stroke,fill}%
}%
\begin{pgfscope}%
\pgfsys@transformshift{5.347684in}{0.986070in}%
\pgfsys@useobject{currentmarker}{}%
\end{pgfscope}%
\end{pgfscope}%
\begin{pgfscope}%
\definecolor{textcolor}{rgb}{0.000000,0.000000,0.000000}%
\pgfsetstrokecolor{textcolor}%
\pgfsetfillcolor{textcolor}%
\pgftext[x=5.210922in, y=0.446403in, left, base,rotate=45.000000]{\color{textcolor}{\rmfamily\fontsize{14.400000}{17.280000}\selectfont\catcode`\^=\active\def^{\ifmmode\sp\else\^{}\fi}\catcode`\%=\active\def
\end{pgfscope}%
\begin{pgfscope}%
\definecolor{textcolor}{rgb}{0.000000,0.000000,0.000000}%
\pgfsetstrokecolor{textcolor}%
\pgfsetfillcolor{textcolor}%
\pgftext[x=3.543053in,y=0.363349in,,top]{\color{textcolor}{\rmfamily\fontsize{17.280000}{20.736000}\selectfont\catcode`\^=\active\def^{\ifmmode\sp\else\^{}\fi}\catcode`\%=\active\def
\end{pgfscope}%
\begin{pgfscope}%
\pgfpathrectangle{\pgfqpoint{0.836107in}{0.986070in}}{\pgfqpoint{5.413893in}{3.663930in}}%
\pgfusepath{clip}%
\pgfsetbuttcap%
\pgfsetroundjoin%
\pgfsetlinewidth{1.003750pt}%
\definecolor{currentstroke}{rgb}{0.501961,0.501961,0.501961}%
\pgfsetstrokecolor{currentstroke}%
\pgfsetstrokeopacity{0.200000}%
\pgfsetdash{{3.700000pt}{1.600000pt}}{0.000000pt}%
\pgfpathmoveto{\pgfqpoint{0.836107in}{0.986070in}}%
\pgfpathlineto{\pgfqpoint{6.250000in}{0.986070in}}%
\pgfusepath{stroke}%
\end{pgfscope}%
\begin{pgfscope}%
\pgfsetbuttcap%
\pgfsetroundjoin%
\definecolor{currentfill}{rgb}{0.000000,0.000000,0.000000}%
\pgfsetfillcolor{currentfill}%
\pgfsetlinewidth{0.803000pt}%
\definecolor{currentstroke}{rgb}{0.000000,0.000000,0.000000}%
\pgfsetstrokecolor{currentstroke}%
\pgfsetdash{}{0pt}%
\pgfsys@defobject{currentmarker}{\pgfqpoint{-0.048611in}{0.000000in}}{\pgfqpoint{-0.000000in}{0.000000in}}{%
\pgfpathmoveto{\pgfqpoint{-0.000000in}{0.000000in}}%
\pgfpathlineto{\pgfqpoint{-0.048611in}{0.000000in}}%
\pgfusepath{stroke,fill}%
}%
\begin{pgfscope}%
\pgfsys@transformshift{0.836107in}{0.986070in}%
\pgfsys@useobject{currentmarker}{}%
\end{pgfscope}%
\end{pgfscope}%
\begin{pgfscope}%
\definecolor{textcolor}{rgb}{0.000000,0.000000,0.000000}%
\pgfsetstrokecolor{textcolor}%
\pgfsetfillcolor{textcolor}%
\pgftext[x=0.640969in, y=0.916625in, left, base]{\color{textcolor}{\rmfamily\fontsize{14.400000}{17.280000}\selectfont\catcode`\^=\active\def^{\ifmmode\sp\else\^{}\fi}\catcode`\%=\active\def
\end{pgfscope}%
\begin{pgfscope}%
\pgfpathrectangle{\pgfqpoint{0.836107in}{0.986070in}}{\pgfqpoint{5.413893in}{3.663930in}}%
\pgfusepath{clip}%
\pgfsetbuttcap%
\pgfsetroundjoin%
\pgfsetlinewidth{1.003750pt}%
\definecolor{currentstroke}{rgb}{0.501961,0.501961,0.501961}%
\pgfsetstrokecolor{currentstroke}%
\pgfsetstrokeopacity{0.200000}%
\pgfsetdash{{3.700000pt}{1.600000pt}}{0.000000pt}%
\pgfpathmoveto{\pgfqpoint{0.836107in}{1.447608in}}%
\pgfpathlineto{\pgfqpoint{6.250000in}{1.447608in}}%
\pgfusepath{stroke}%
\end{pgfscope}%
\begin{pgfscope}%
\pgfsetbuttcap%
\pgfsetroundjoin%
\definecolor{currentfill}{rgb}{0.000000,0.000000,0.000000}%
\pgfsetfillcolor{currentfill}%
\pgfsetlinewidth{0.803000pt}%
\definecolor{currentstroke}{rgb}{0.000000,0.000000,0.000000}%
\pgfsetstrokecolor{currentstroke}%
\pgfsetdash{}{0pt}%
\pgfsys@defobject{currentmarker}{\pgfqpoint{-0.048611in}{0.000000in}}{\pgfqpoint{-0.000000in}{0.000000in}}{%
\pgfpathmoveto{\pgfqpoint{-0.000000in}{0.000000in}}%
\pgfpathlineto{\pgfqpoint{-0.048611in}{0.000000in}}%
\pgfusepath{stroke,fill}%
}%
\begin{pgfscope}%
\pgfsys@transformshift{0.836107in}{1.447608in}%
\pgfsys@useobject{currentmarker}{}%
\end{pgfscope}%
\end{pgfscope}%
\begin{pgfscope}%
\definecolor{textcolor}{rgb}{0.000000,0.000000,0.000000}%
\pgfsetstrokecolor{textcolor}%
\pgfsetfillcolor{textcolor}%
\pgftext[x=0.543054in, y=1.378163in, left, base]{\color{textcolor}{\rmfamily\fontsize{14.400000}{17.280000}\selectfont\catcode`\^=\active\def^{\ifmmode\sp\else\^{}\fi}\catcode`\%=\active\def
\end{pgfscope}%
\begin{pgfscope}%
\pgfpathrectangle{\pgfqpoint{0.836107in}{0.986070in}}{\pgfqpoint{5.413893in}{3.663930in}}%
\pgfusepath{clip}%
\pgfsetbuttcap%
\pgfsetroundjoin%
\pgfsetlinewidth{1.003750pt}%
\definecolor{currentstroke}{rgb}{0.501961,0.501961,0.501961}%
\pgfsetstrokecolor{currentstroke}%
\pgfsetstrokeopacity{0.200000}%
\pgfsetdash{{3.700000pt}{1.600000pt}}{0.000000pt}%
\pgfpathmoveto{\pgfqpoint{0.836107in}{1.909146in}}%
\pgfpathlineto{\pgfqpoint{6.250000in}{1.909146in}}%
\pgfusepath{stroke}%
\end{pgfscope}%
\begin{pgfscope}%
\pgfsetbuttcap%
\pgfsetroundjoin%
\definecolor{currentfill}{rgb}{0.000000,0.000000,0.000000}%
\pgfsetfillcolor{currentfill}%
\pgfsetlinewidth{0.803000pt}%
\definecolor{currentstroke}{rgb}{0.000000,0.000000,0.000000}%
\pgfsetstrokecolor{currentstroke}%
\pgfsetdash{}{0pt}%
\pgfsys@defobject{currentmarker}{\pgfqpoint{-0.048611in}{0.000000in}}{\pgfqpoint{-0.000000in}{0.000000in}}{%
\pgfpathmoveto{\pgfqpoint{-0.000000in}{0.000000in}}%
\pgfpathlineto{\pgfqpoint{-0.048611in}{0.000000in}}%
\pgfusepath{stroke,fill}%
}%
\begin{pgfscope}%
\pgfsys@transformshift{0.836107in}{1.909146in}%
\pgfsys@useobject{currentmarker}{}%
\end{pgfscope}%
\end{pgfscope}%
\begin{pgfscope}%
\definecolor{textcolor}{rgb}{0.000000,0.000000,0.000000}%
\pgfsetstrokecolor{textcolor}%
\pgfsetfillcolor{textcolor}%
\pgftext[x=0.543054in, y=1.839701in, left, base]{\color{textcolor}{\rmfamily\fontsize{14.400000}{17.280000}\selectfont\catcode`\^=\active\def^{\ifmmode\sp\else\^{}\fi}\catcode`\%=\active\def
\end{pgfscope}%
\begin{pgfscope}%
\pgfpathrectangle{\pgfqpoint{0.836107in}{0.986070in}}{\pgfqpoint{5.413893in}{3.663930in}}%
\pgfusepath{clip}%
\pgfsetbuttcap%
\pgfsetroundjoin%
\pgfsetlinewidth{1.003750pt}%
\definecolor{currentstroke}{rgb}{0.501961,0.501961,0.501961}%
\pgfsetstrokecolor{currentstroke}%
\pgfsetstrokeopacity{0.200000}%
\pgfsetdash{{3.700000pt}{1.600000pt}}{0.000000pt}%
\pgfpathmoveto{\pgfqpoint{0.836107in}{2.370683in}}%
\pgfpathlineto{\pgfqpoint{6.250000in}{2.370683in}}%
\pgfusepath{stroke}%
\end{pgfscope}%
\begin{pgfscope}%
\pgfsetbuttcap%
\pgfsetroundjoin%
\definecolor{currentfill}{rgb}{0.000000,0.000000,0.000000}%
\pgfsetfillcolor{currentfill}%
\pgfsetlinewidth{0.803000pt}%
\definecolor{currentstroke}{rgb}{0.000000,0.000000,0.000000}%
\pgfsetstrokecolor{currentstroke}%
\pgfsetdash{}{0pt}%
\pgfsys@defobject{currentmarker}{\pgfqpoint{-0.048611in}{0.000000in}}{\pgfqpoint{-0.000000in}{0.000000in}}{%
\pgfpathmoveto{\pgfqpoint{-0.000000in}{0.000000in}}%
\pgfpathlineto{\pgfqpoint{-0.048611in}{0.000000in}}%
\pgfusepath{stroke,fill}%
}%
\begin{pgfscope}%
\pgfsys@transformshift{0.836107in}{2.370683in}%
\pgfsys@useobject{currentmarker}{}%
\end{pgfscope}%
\end{pgfscope}%
\begin{pgfscope}%
\definecolor{textcolor}{rgb}{0.000000,0.000000,0.000000}%
\pgfsetstrokecolor{textcolor}%
\pgfsetfillcolor{textcolor}%
\pgftext[x=0.543054in, y=2.301239in, left, base]{\color{textcolor}{\rmfamily\fontsize{14.400000}{17.280000}\selectfont\catcode`\^=\active\def^{\ifmmode\sp\else\^{}\fi}\catcode`\%=\active\def
\end{pgfscope}%
\begin{pgfscope}%
\pgfpathrectangle{\pgfqpoint{0.836107in}{0.986070in}}{\pgfqpoint{5.413893in}{3.663930in}}%
\pgfusepath{clip}%
\pgfsetbuttcap%
\pgfsetroundjoin%
\pgfsetlinewidth{1.003750pt}%
\definecolor{currentstroke}{rgb}{0.501961,0.501961,0.501961}%
\pgfsetstrokecolor{currentstroke}%
\pgfsetstrokeopacity{0.200000}%
\pgfsetdash{{3.700000pt}{1.600000pt}}{0.000000pt}%
\pgfpathmoveto{\pgfqpoint{0.836107in}{2.832221in}}%
\pgfpathlineto{\pgfqpoint{6.250000in}{2.832221in}}%
\pgfusepath{stroke}%
\end{pgfscope}%
\begin{pgfscope}%
\pgfsetbuttcap%
\pgfsetroundjoin%
\definecolor{currentfill}{rgb}{0.000000,0.000000,0.000000}%
\pgfsetfillcolor{currentfill}%
\pgfsetlinewidth{0.803000pt}%
\definecolor{currentstroke}{rgb}{0.000000,0.000000,0.000000}%
\pgfsetstrokecolor{currentstroke}%
\pgfsetdash{}{0pt}%
\pgfsys@defobject{currentmarker}{\pgfqpoint{-0.048611in}{0.000000in}}{\pgfqpoint{-0.000000in}{0.000000in}}{%
\pgfpathmoveto{\pgfqpoint{-0.000000in}{0.000000in}}%
\pgfpathlineto{\pgfqpoint{-0.048611in}{0.000000in}}%
\pgfusepath{stroke,fill}%
}%
\begin{pgfscope}%
\pgfsys@transformshift{0.836107in}{2.832221in}%
\pgfsys@useobject{currentmarker}{}%
\end{pgfscope}%
\end{pgfscope}%
\begin{pgfscope}%
\definecolor{textcolor}{rgb}{0.000000,0.000000,0.000000}%
\pgfsetstrokecolor{textcolor}%
\pgfsetfillcolor{textcolor}%
\pgftext[x=0.543054in, y=2.762777in, left, base]{\color{textcolor}{\rmfamily\fontsize{14.400000}{17.280000}\selectfont\catcode`\^=\active\def^{\ifmmode\sp\else\^{}\fi}\catcode`\%=\active\def
\end{pgfscope}%
\begin{pgfscope}%
\pgfpathrectangle{\pgfqpoint{0.836107in}{0.986070in}}{\pgfqpoint{5.413893in}{3.663930in}}%
\pgfusepath{clip}%
\pgfsetbuttcap%
\pgfsetroundjoin%
\pgfsetlinewidth{1.003750pt}%
\definecolor{currentstroke}{rgb}{0.501961,0.501961,0.501961}%
\pgfsetstrokecolor{currentstroke}%
\pgfsetstrokeopacity{0.200000}%
\pgfsetdash{{3.700000pt}{1.600000pt}}{0.000000pt}%
\pgfpathmoveto{\pgfqpoint{0.836107in}{3.293759in}}%
\pgfpathlineto{\pgfqpoint{6.250000in}{3.293759in}}%
\pgfusepath{stroke}%
\end{pgfscope}%
\begin{pgfscope}%
\pgfsetbuttcap%
\pgfsetroundjoin%
\definecolor{currentfill}{rgb}{0.000000,0.000000,0.000000}%
\pgfsetfillcolor{currentfill}%
\pgfsetlinewidth{0.803000pt}%
\definecolor{currentstroke}{rgb}{0.000000,0.000000,0.000000}%
\pgfsetstrokecolor{currentstroke}%
\pgfsetdash{}{0pt}%
\pgfsys@defobject{currentmarker}{\pgfqpoint{-0.048611in}{0.000000in}}{\pgfqpoint{-0.000000in}{0.000000in}}{%
\pgfpathmoveto{\pgfqpoint{-0.000000in}{0.000000in}}%
\pgfpathlineto{\pgfqpoint{-0.048611in}{0.000000in}}%
\pgfusepath{stroke,fill}%
}%
\begin{pgfscope}%
\pgfsys@transformshift{0.836107in}{3.293759in}%
\pgfsys@useobject{currentmarker}{}%
\end{pgfscope}%
\end{pgfscope}%
\begin{pgfscope}%
\definecolor{textcolor}{rgb}{0.000000,0.000000,0.000000}%
\pgfsetstrokecolor{textcolor}%
\pgfsetfillcolor{textcolor}%
\pgftext[x=0.445138in, y=3.224315in, left, base]{\color{textcolor}{\rmfamily\fontsize{14.400000}{17.280000}\selectfont\catcode`\^=\active\def^{\ifmmode\sp\else\^{}\fi}\catcode`\%=\active\def
\end{pgfscope}%
\begin{pgfscope}%
\pgfpathrectangle{\pgfqpoint{0.836107in}{0.986070in}}{\pgfqpoint{5.413893in}{3.663930in}}%
\pgfusepath{clip}%
\pgfsetbuttcap%
\pgfsetroundjoin%
\pgfsetlinewidth{1.003750pt}%
\definecolor{currentstroke}{rgb}{0.501961,0.501961,0.501961}%
\pgfsetstrokecolor{currentstroke}%
\pgfsetstrokeopacity{0.200000}%
\pgfsetdash{{3.700000pt}{1.600000pt}}{0.000000pt}%
\pgfpathmoveto{\pgfqpoint{0.836107in}{3.755297in}}%
\pgfpathlineto{\pgfqpoint{6.250000in}{3.755297in}}%
\pgfusepath{stroke}%
\end{pgfscope}%
\begin{pgfscope}%
\pgfsetbuttcap%
\pgfsetroundjoin%
\definecolor{currentfill}{rgb}{0.000000,0.000000,0.000000}%
\pgfsetfillcolor{currentfill}%
\pgfsetlinewidth{0.803000pt}%
\definecolor{currentstroke}{rgb}{0.000000,0.000000,0.000000}%
\pgfsetstrokecolor{currentstroke}%
\pgfsetdash{}{0pt}%
\pgfsys@defobject{currentmarker}{\pgfqpoint{-0.048611in}{0.000000in}}{\pgfqpoint{-0.000000in}{0.000000in}}{%
\pgfpathmoveto{\pgfqpoint{-0.000000in}{0.000000in}}%
\pgfpathlineto{\pgfqpoint{-0.048611in}{0.000000in}}%
\pgfusepath{stroke,fill}%
}%
\begin{pgfscope}%
\pgfsys@transformshift{0.836107in}{3.755297in}%
\pgfsys@useobject{currentmarker}{}%
\end{pgfscope}%
\end{pgfscope}%
\begin{pgfscope}%
\definecolor{textcolor}{rgb}{0.000000,0.000000,0.000000}%
\pgfsetstrokecolor{textcolor}%
\pgfsetfillcolor{textcolor}%
\pgftext[x=0.445138in, y=3.685853in, left, base]{\color{textcolor}{\rmfamily\fontsize{14.400000}{17.280000}\selectfont\catcode`\^=\active\def^{\ifmmode\sp\else\^{}\fi}\catcode`\%=\active\def
\end{pgfscope}%
\begin{pgfscope}%
\pgfpathrectangle{\pgfqpoint{0.836107in}{0.986070in}}{\pgfqpoint{5.413893in}{3.663930in}}%
\pgfusepath{clip}%
\pgfsetbuttcap%
\pgfsetroundjoin%
\pgfsetlinewidth{1.003750pt}%
\definecolor{currentstroke}{rgb}{0.501961,0.501961,0.501961}%
\pgfsetstrokecolor{currentstroke}%
\pgfsetstrokeopacity{0.200000}%
\pgfsetdash{{3.700000pt}{1.600000pt}}{0.000000pt}%
\pgfpathmoveto{\pgfqpoint{0.836107in}{4.216835in}}%
\pgfpathlineto{\pgfqpoint{6.250000in}{4.216835in}}%
\pgfusepath{stroke}%
\end{pgfscope}%
\begin{pgfscope}%
\pgfsetbuttcap%
\pgfsetroundjoin%
\definecolor{currentfill}{rgb}{0.000000,0.000000,0.000000}%
\pgfsetfillcolor{currentfill}%
\pgfsetlinewidth{0.803000pt}%
\definecolor{currentstroke}{rgb}{0.000000,0.000000,0.000000}%
\pgfsetstrokecolor{currentstroke}%
\pgfsetdash{}{0pt}%
\pgfsys@defobject{currentmarker}{\pgfqpoint{-0.048611in}{0.000000in}}{\pgfqpoint{-0.000000in}{0.000000in}}{%
\pgfpathmoveto{\pgfqpoint{-0.000000in}{0.000000in}}%
\pgfpathlineto{\pgfqpoint{-0.048611in}{0.000000in}}%
\pgfusepath{stroke,fill}%
}%
\begin{pgfscope}%
\pgfsys@transformshift{0.836107in}{4.216835in}%
\pgfsys@useobject{currentmarker}{}%
\end{pgfscope}%
\end{pgfscope}%
\begin{pgfscope}%
\definecolor{textcolor}{rgb}{0.000000,0.000000,0.000000}%
\pgfsetstrokecolor{textcolor}%
\pgfsetfillcolor{textcolor}%
\pgftext[x=0.445138in, y=4.147391in, left, base]{\color{textcolor}{\rmfamily\fontsize{14.400000}{17.280000}\selectfont\catcode`\^=\active\def^{\ifmmode\sp\else\^{}\fi}\catcode`\%=\active\def
\end{pgfscope}%
\begin{pgfscope}%
\definecolor{textcolor}{rgb}{0.000000,0.000000,0.000000}%
\pgfsetstrokecolor{textcolor}%
\pgfsetfillcolor{textcolor}%
\pgftext[x=0.389583in,y=2.818035in,,bottom,rotate=90.000000]{\color{textcolor}{\rmfamily\fontsize{17.280000}{20.736000}\selectfont\catcode`\^=\active\def^{\ifmmode\sp\else\^{}\fi}\catcode`\%=\active\def
\end{pgfscope}%
\begin{pgfscope}%
\pgfsetrectcap%
\pgfsetmiterjoin%
\pgfsetlinewidth{0.803000pt}%
\definecolor{currentstroke}{rgb}{0.000000,0.000000,0.000000}%
\pgfsetstrokecolor{currentstroke}%
\pgfsetdash{}{0pt}%
\pgfpathmoveto{\pgfqpoint{0.836107in}{0.986070in}}%
\pgfpathlineto{\pgfqpoint{0.836107in}{4.650000in}}%
\pgfusepath{stroke}%
\end{pgfscope}%
\begin{pgfscope}%
\pgfsetrectcap%
\pgfsetmiterjoin%
\pgfsetlinewidth{0.803000pt}%
\definecolor{currentstroke}{rgb}{0.000000,0.000000,0.000000}%
\pgfsetstrokecolor{currentstroke}%
\pgfsetdash{}{0pt}%
\pgfpathmoveto{\pgfqpoint{6.250000in}{0.986070in}}%
\pgfpathlineto{\pgfqpoint{6.250000in}{4.650000in}}%
\pgfusepath{stroke}%
\end{pgfscope}%
\begin{pgfscope}%
\pgfsetrectcap%
\pgfsetmiterjoin%
\pgfsetlinewidth{0.803000pt}%
\definecolor{currentstroke}{rgb}{0.000000,0.000000,0.000000}%
\pgfsetstrokecolor{currentstroke}%
\pgfsetdash{}{0pt}%
\pgfpathmoveto{\pgfqpoint{0.836107in}{0.986070in}}%
\pgfpathlineto{\pgfqpoint{6.250000in}{0.986070in}}%
\pgfusepath{stroke}%
\end{pgfscope}%
\begin{pgfscope}%
\pgfsetrectcap%
\pgfsetmiterjoin%
\pgfsetlinewidth{0.803000pt}%
\definecolor{currentstroke}{rgb}{0.000000,0.000000,0.000000}%
\pgfsetstrokecolor{currentstroke}%
\pgfsetdash{}{0pt}%
\pgfpathmoveto{\pgfqpoint{0.836107in}{4.650000in}}%
\pgfpathlineto{\pgfqpoint{6.250000in}{4.650000in}}%
\pgfusepath{stroke}%
\end{pgfscope}%
\begin{pgfscope}%
\pgfsetbuttcap%
\pgfsetmiterjoin%
\definecolor{currentfill}{rgb}{1.000000,1.000000,1.000000}%
\pgfsetfillcolor{currentfill}%
\pgfsetfillopacity{0.800000}%
\pgfsetlinewidth{1.003750pt}%
\definecolor{currentstroke}{rgb}{0.800000,0.800000,0.800000}%
\pgfsetstrokecolor{currentstroke}%
\pgfsetstrokeopacity{0.800000}%
\pgfsetdash{}{0pt}%
\pgfpathmoveto{\pgfqpoint{5.289351in}{3.957871in}}%
\pgfpathlineto{\pgfqpoint{6.152778in}{3.957871in}}%
\pgfpathquadraticcurveto{\pgfqpoint{6.180556in}{3.957871in}}{\pgfqpoint{6.180556in}{3.985648in}}%
\pgfpathlineto{\pgfqpoint{6.180556in}{4.552778in}}%
\pgfpathquadraticcurveto{\pgfqpoint{6.180556in}{4.580556in}}{\pgfqpoint{6.152778in}{4.580556in}}%
\pgfpathlineto{\pgfqpoint{5.289351in}{4.580556in}}%
\pgfpathquadraticcurveto{\pgfqpoint{5.261573in}{4.580556in}}{\pgfqpoint{5.261573in}{4.552778in}}%
\pgfpathlineto{\pgfqpoint{5.261573in}{3.985648in}}%
\pgfpathquadraticcurveto{\pgfqpoint{5.261573in}{3.957871in}}{\pgfqpoint{5.289351in}{3.957871in}}%
\pgfpathlineto{\pgfqpoint{5.289351in}{3.957871in}}%
\pgfpathclose%
\pgfusepath{stroke,fill}%
\end{pgfscope}%
\begin{pgfscope}%
\pgfsetbuttcap%
\pgfsetmiterjoin%
\definecolor{currentfill}{rgb}{0.121569,0.466667,0.705882}%
\pgfsetfillcolor{currentfill}%
\pgfsetlinewidth{0.000000pt}%
\definecolor{currentstroke}{rgb}{0.000000,0.000000,0.000000}%
\pgfsetstrokecolor{currentstroke}%
\pgfsetstrokeopacity{0.000000}%
\pgfsetdash{}{0pt}%
\pgfpathmoveto{\pgfqpoint{5.317128in}{4.427778in}}%
\pgfpathlineto{\pgfqpoint{5.594906in}{4.427778in}}%
\pgfpathlineto{\pgfqpoint{5.594906in}{4.525000in}}%
\pgfpathlineto{\pgfqpoint{5.317128in}{4.525000in}}%
\pgfpathlineto{\pgfqpoint{5.317128in}{4.427778in}}%
\pgfpathclose%
\pgfusepath{fill}%
\end{pgfscope}%
\begin{pgfscope}%
\definecolor{textcolor}{rgb}{0.000000,0.000000,0.000000}%
\pgfsetstrokecolor{textcolor}%
\pgfsetfillcolor{textcolor}%
\pgftext[x=5.706017in,y=4.427778in,left,base]{\color{textcolor}{\rmfamily\fontsize{10.000000}{12.000000}\selectfont\catcode`\^=\active\def^{\ifmmode\sp\else\^{}\fi}\catcode`\%=\active\def
\end{pgfscope}%
\begin{pgfscope}%
\pgfsetbuttcap%
\pgfsetmiterjoin%
\definecolor{currentfill}{rgb}{1.000000,0.498039,0.054902}%
\pgfsetfillcolor{currentfill}%
\pgfsetlinewidth{0.000000pt}%
\definecolor{currentstroke}{rgb}{0.000000,0.000000,0.000000}%
\pgfsetstrokecolor{currentstroke}%
\pgfsetstrokeopacity{0.000000}%
\pgfsetdash{}{0pt}%
\pgfpathmoveto{\pgfqpoint{5.317128in}{4.234105in}}%
\pgfpathlineto{\pgfqpoint{5.594906in}{4.234105in}}%
\pgfpathlineto{\pgfqpoint{5.594906in}{4.331327in}}%
\pgfpathlineto{\pgfqpoint{5.317128in}{4.331327in}}%
\pgfpathlineto{\pgfqpoint{5.317128in}{4.234105in}}%
\pgfpathclose%
\pgfusepath{fill}%
\end{pgfscope}%
\begin{pgfscope}%
\definecolor{textcolor}{rgb}{0.000000,0.000000,0.000000}%
\pgfsetstrokecolor{textcolor}%
\pgfsetfillcolor{textcolor}%
\pgftext[x=5.706017in,y=4.234105in,left,base]{\color{textcolor}{\rmfamily\fontsize{10.000000}{12.000000}\selectfont\catcode`\^=\active\def^{\ifmmode\sp\else\^{}\fi}\catcode`\%=\active\def
\end{pgfscope}%
\begin{pgfscope}%
\pgfsetbuttcap%
\pgfsetmiterjoin%
\definecolor{currentfill}{rgb}{0.172549,0.627451,0.172549}%
\pgfsetfillcolor{currentfill}%
\pgfsetlinewidth{0.000000pt}%
\definecolor{currentstroke}{rgb}{0.000000,0.000000,0.000000}%
\pgfsetstrokecolor{currentstroke}%
\pgfsetstrokeopacity{0.000000}%
\pgfsetdash{}{0pt}%
\pgfpathmoveto{\pgfqpoint{5.317128in}{4.040432in}}%
\pgfpathlineto{\pgfqpoint{5.594906in}{4.040432in}}%
\pgfpathlineto{\pgfqpoint{5.594906in}{4.137654in}}%
\pgfpathlineto{\pgfqpoint{5.317128in}{4.137654in}}%
\pgfpathlineto{\pgfqpoint{5.317128in}{4.040432in}}%
\pgfpathclose%
\pgfusepath{fill}%
\end{pgfscope}%
\begin{pgfscope}%
\definecolor{textcolor}{rgb}{0.000000,0.000000,0.000000}%
\pgfsetstrokecolor{textcolor}%
\pgfsetfillcolor{textcolor}%
\pgftext[x=5.706017in,y=4.040432in,left,base]{\color{textcolor}{\rmfamily\fontsize{10.000000}{12.000000}\selectfont\catcode`\^=\active\def^{\ifmmode\sp\else\^{}\fi}\catcode`\%=\active\def
\end{pgfscope}%
\end{pgfpicture}%
\makeatother%
\endgroup%

%% file: content/06_discussion.tex

%% file: content/07_related_work.tex
\section{Related Work}
\label{sec:related}

Digital prototyping has become an indispensable tool in product design,
allowing it to avoid development risks and shorten the time-to-market. It
enables industrial designers and manufacturers to virtually explore the
complete product development process before it is physically realized.
Digital prototyping can be used early in product design and validation, as well
as to optimize the functionality, energy efficiency, and cost of existing
products entirely virtually, using suitable simulation software.
Indeed, since numerical simulations have been used in metal forming processes,
the development cycle from design to production has been almost halved from the
early 1990s (48 months) to the 2020s (18-21 months)~\cite{JC24}.

One of the challenges in digital prototyping is to find optimal design
parameters that yield the desired target parameters. In some cases, it is
possible to find the optimal design parameters directly. Since there is no
information available about the function (simulation) that we wish to optimize,
we are left with black-box optimization methods. Black-box optimization refers
to scenarios where the objective function and the constraints are unknown,
unexploitable, or non-existent~\cite{alarie2021two}. Such methods have been
applied where the function is expensive, and only a limited number of function
evaluations are possible, such as in computational fluid dynamics (CFD)
simulations~\cite{ansotegui2021learning}. Black-box optimization is also
referred to as derivative-free optimization, gradient-free optimization, or
optimization without derivatives, simulation-based optimization, and
zeroth-order optimization~\cite{larson2019derivative}, meaning the derivative
information is not available, unreliable, or impractical to obtain; alongside
the function that is expensive to evaluate or somewhat noisy, makes most
algorithms based on finite differences
unsuitable~\cite{rios2013derivative,kolda2003optimization,wild2008orbit}.
Black-box optimization methods have been used for a long time across various
fields, including computer science, mathematics, and mechanical engineering,
among others~\cite{SA21}. One of the most widely spread optimization approaches
is Bayesian Optimization. Various works have employed Bayesian optimization for
diverse scientific problems, including chemical synthesis~\cite{BJS21},
materials science~\cite{DP17a}, numerical physics simulations~\cite{SGM23}, and
numerous other fields~\cite{SA21}. Often, Gaussian Processes are used to model
the objective function, providing predictions coupled with uncertainty
estimations. Recent approaches have attempted to eliminate the need for a
separate Gaussian Process model per constraint by utilizing prior-data fitted
networks and have demonstrated their applicability on a variety of
problems~\cite{RTYY25}. Unlike such approaches, we employ multiple latent
Gaussian Processes with a distinct Marten kernel and model each output as a
linear combination of these latent functions. As our target parameters are
strongly correlated, our GPLVM enables us to accurately model the dependencies
between them, allowing us to efficiently explore the multi-objective design
space.

Sheet metal forming simulation plays an indispensable role in integrating
manufacturing necessities into the product design process at an early
stage~\cite{Ranganath2012Finite}. In \cite{BS21}, the authors focus on
optimizing numerical methods for simultaneously optimizing the blank shape and
forming tool geometry. In contrast, we do not focus on the numerical method but
rather target the selection of the input samples using AI methods. Other
studies employ multi-objective optimization using a Pareto-based genetic
algorithm~\cite{LW08}, revealing a conflicting relationship between the
objective functions in sheet forming, such that solutions that simultaneously
minimize each objective are nearly impossible to obtain. In a recent
work~\cite{JC24}, the authors surveyed the state-of-the-art of AI algorithms
used for metal forming processes. Many of these works focus on data-driven
material modeling (e.g., \cite{MM19,LW20}), prediction of
fracture~\cite{XS22,DY22,KSP20}, or the forming limit
diagram~\cite{CJ19,AMC19}. Still, the adoption of ML-based or hybrid approaches
in industrial forming simulation has been slow, often due to data
availability~\cite{JC24}.
Our approach integrates Bayesian optimization with the design of sheet metal
forming parts. It allows specifying target and input parameters with their
constraints and optimizes toward a global target. This reduces reliance on
expert knowledge, accelerates design space exploration, shortens time to
solution, and improves the quality of the results.

%% file: content/08_conclusion.tex
\section{Conclusion} 
\label{sec:conclusion}
In this paper, we presented an approach to reduce the expert involvement in the sheet metal design process. Through the use of different AI models (DNN and GPLVM), we designed a workflow that builds upon previous simulations to determine the optimal input configuration. For new parts, we provide a model of experts that guides the selection of input parameter configurations built on similar structured parts. Overall, our approach not only reduced expert involvement but also improved the design by identifying better configurations compared to those suggested by the expert. While we have already incorporated the geometry of the parts into the AI workflow (i.e., geometric encoding, as described in \cref{subsec:embedding}), we reserve aspects such as modifying the geometry for future work. Furthermore, future work also targets running the parallel samples on multiple nodes using MPI, utilizing approximated computing to adjust the accuracy of the solver, and replacing the GPLVM model with a DNN once sufficient samples are available for improved scalability.

%% file: content/10_acknowledgement.tex
\section*{Acknowledgment}
\label{sec:acknowledgements}
The Federal Ministry of Research, Technology and Space (BMFTR) funded the project ENSIMA on which this report is based
under the funding code 16ME0633. The authors are responsible for the content of this publication.

The authors gratefully acknowledge the computing time provided to them on the high-performance computer Lichtenberg II at TU Darmstadt, funded by BMFTR and the State of Hesse.

%% file: preamble/gloss_print.tex
\ifgloss
\printglossary[style=dottedlocations, type=\acronymtype, title=List of Acronyms and Abbreviations, toctitle=List of Acronyms and Abbreviations]
%
\printglossary[type=main]
\fi